\documentclass{article}

\usepackage{hyperref}

\usepackage[preprint]{icml2026}

\usepackage{graphicx}
\usepackage{subcaption}
\usepackage{booktabs} % for professional tables
\usepackage{tabularx}
\usepackage{tabularray}
\usepackage{amsmath,amssymb,amsfonts,amsthm,mathrsfs}
\usepackage{thmtools, thm-restate}
\usepackage[dvipsnames,svgnames]{xcolor}  
\usepackage{titletoc}

\usepackage{trimclip} % Required for the \clipbox command
\usepackage{stmaryrd} % Often required for \shortrightarrow

\makeatletter
\DeclareRobustCommand{\shortto}{%
  \mathrel{\mathpalette\short@to\relax}%
}

\newcommand{\short@to}[2]{%
  \mkern2mu
  \clipbox{{.5\width} 0 0 0}{$\m@th#1\vphantom{+}{\shortrightarrow}$}%
}
\makeatother

\usepackage{xurl}            % simple URL typesetting
\usepackage{booktabs}       % professional-quality tables
\usepackage{amsfonts}       % blackboard math symbols
\usepackage{nicefrac}       % compact symbols for 1/2, etc.
\usepackage{microtype}      % microtypography
\usepackage{xcolor}         % colors
\usepackage{multirow}
\usepackage{quoting}
\usepackage{marvosym}
\usepackage{algorithm}
\usepackage{algorithmic}
\usepackage{bbm}
\usepackage{verbatim}

\usepackage{xspace}
\usepackage{mathtools}
\usepackage[nameinlink,capitalize]{cleveref}
\usepackage{tcolorbox}   % Required for the tcolorbox environments
\tcbuselibrary{breakable} % Required for the 'breakable' option inside tcolorbox
\usepackage{pifont}      % Required for the finger pointer symbol \ding{43}

\definecolor{takeawayframe}{RGB}{80, 100, 135} 
\definecolor{takeawayback}{RGB}{251, 252, 255} 
\definecolor{intuitionframe}{RGB}{90, 75, 145} 
\definecolor{intuitionback}{RGB}{251, 250, 255}

\newcounter{takeaway}
\crefname{takeaway}{Takeaway}{Takeaways}
\newenvironment{takeaway}[1]{%
  \refstepcounter{takeaway}%
  \begin{tcolorbox}[
    colframe=takeawayframe, 
    colback=takeawayback,   
    boxrule=0.5pt,           
    arc=2pt,                 
    left=2pt, right=2pt, top=2pt, bottom=2pt,
    breakable
  ]
  \textbf{\ding{43} Takeaway \thetakeaway} (#1)\textbf{.}
}{%
  \end{tcolorbox}%
}

\newenvironment{proofintuition}{%
  \begin{tcolorbox}[
    colframe=intuitionframe, 
    colback=intuitionback,   
    boxrule=0.5pt,           
    arc=2pt,                 
    left=2pt, right=2pt, top=2pt, bottom=2pt,
    breakable,
    parbox=false, 
    before upper={\setlength{\parskip}{\smallskipamount}} 
  ]
  \textbf{\ding{43} Proof intuition.} 
}{%
  \end{tcolorbox}%
}

\theoremstyle{plain}

\newtheorem{lemma}{Lemma}[section]
\newtheorem{corollary}{Corollary}[section]

\theoremstyle{definition}
\newtheorem{definition}{Definition}[section]
\newtheorem{assumption}{Assumption}[section]
\theoremstyle{remark}
\newtheorem{remark}{Remark}[section]

\crefname{section}{\S}{\S\S}
\crefname{appendix}{App.}{Apps.}
\crefname{subappendix}{App.}{Apps.}
\crefname{subsubappendix}{App.}{Apps.}
\crefname{figure}{Fig.}{Figs.}
\crefname{table}{Tab.}{Tabs.}
\crefname{equation}{Eq.}{Eqs.}
\crefname{appendix}{App.}{Apps.}
\crefname{theorem}{Thm.}{Thms.}
\crefname{proposition}{Prop.}{Props.}
\crefname{lemma}{Lem.}{Lems.}
\crefname{corollary}{Cor.}{Cors.}
\crefname{definition}{Def.}{Defs.}
\crefname{assumption}{Asm.}{Asms.} 
\crefname{example}{Ex.}{Exs.}
\crefname{remark}{Rem.}{Rems.}

\usepackage{enumitem}

\icmltitlerunning{The Geometric Mechanics of Contrastive Representation Learning}

\begin{document}

\twocolumn[
  \icmltitle{The Geometric Mechanics of Contrastive Representation Learning:\\ Alignment Potentials, Entropic Dispersion, and Cross-Modal Divergence}



  \icmlsetsymbol{equal}{*}

  \begin{icmlauthorlist}
    \icmlauthor{Yichao Cai}{sch}
    \icmlauthor{Zhen Zhang}{sch,com}
    \icmlauthor{Yuhang Liu}{sch,com}
    \icmlauthor{Javen Qinfeng Shi}{sch,com}
  \end{icmlauthorlist}
  
  \icmlaffiliation{sch}{Australian Institute for Machine Learning, Adelaide University.}
  \icmlaffiliation{com}{Responsible AI Research Centre, Australia}
  \icmlcorrespondingauthor{Yichao Cai}{yichao.cai@adelaide.edu.au}

  \icmlkeywords{Contrastive Learning, InfoNCE, Representation Learning Theory, Modality Gap}

  \vskip 0.3in
]



\printAffiliationsAndNotice{}  % no special notice (required even if empty)

\begin{abstract}%
While InfoNCE underlies modern contrastive learning, its geometric mechanisms remain under-characterized beyond the canonical alignment--uniformity decomposition. We develop a measure-theoretic framework in which representation measures evolve on a fixed embedding manifold. In the large-batch limit, we prove value and gradient consistency, linking the stochastic objective to explicit deterministic energy landscapes and revealing a geometric bifurcation between unimodal and symmetric multimodal regimes. In the unimodal case, the intrinsic energy is strictly convex and admits a unique Gibbs equilibrium, showing that entropy acts as a tie-breaker within the aligned basin. In the multimodal case, the intrinsic geometry becomes cross-coupled and contains a persistent negative symmetric divergence term: each modality's marginal reshapes the effective landscape of the other, allowing strong pairwise alignment to coexist with a persistent modality gap. Controlled synthetic experiments and analyses of pretrained CLIP representations support these predictions. Overall, our results shift the analytical lens from pointwise discrimination to population geometry, showing that pairwise alignment alone is insufficient to control cross-modal marginal structure. Code available at: \url{https://github.com/YichaoCai1/InfoNCE_Geometry}.
\end{abstract}

\section{Introduction}
\label{sec:intro}
InfoNCE-based contrastive learning is a central objective in self-supervised and multimodal representation learning \citep{oord2018representation,radford2021learning,jia2021scaling}, yet its representational geometry remains poorly understood. In multimodal systems such as CLIP, training can achieve strong \emph{pairwise} alignment while the modality \emph{marginals} remain separated, an empirically observed phenomenon known as the modality gap \citep{liang2022mind,levi2025the}, for which a mechanistic explanation is still lacking. Even in the unimodal setting, it remains unclear what population-level energy InfoNCE induces and how its geometry should be understood beyond the alignment--uniformity view \citep{wang2020understanding}. This raises a key geometric question for representation learning: why does matching positive pairs fail to match the induced population distributions? More broadly, what population energies do unimodal and symmetric multimodal InfoNCE induce, and what equilibria does their geometry favor?

Existing theory still lacks a first-principles account of these questions. Density-ratio analyses characterize the optimal critic as pointwise mutual information \citep{gutmann2010nce,oord2018representation}, but do not directly describe the deterministic descent directions induced by the softmax denominator or how these directions reshape the induced representations. The alignment--uniformity view \citep{wang2020understanding} clarifies important asymptotic trade-offs, but leaves open what population structure the InfoNCE objective itself favors. Identifiability results offer recovery guarantees under explicit generative assumptions \citep{zimmermann2021contrastive,liu2026beyond}, addressing what is learnable in principle rather than the objective-induced geometry of the learned marginals. This lack of a geometric account is particularly consequential in multimodal settings, where the symmetric InfoNCE objective is formally an average of two directional terms, yet its induced geometry is not merely the simple average of their individual geometries.

\begin{figure*}[t]
    \centering
    \includegraphics[width=0.95\linewidth]{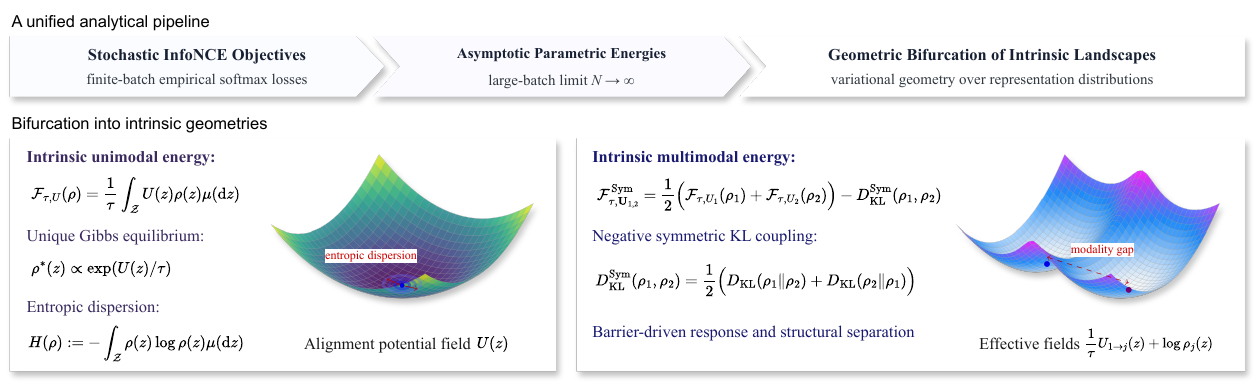}
\caption{A unified analytical pipeline for contrastive learning geometry. Starting from stochastic InfoNCE losses, the large-batch limit yields deterministic parametric energies, which in turn lift to intrinsic landscapes over representation densities. At this intrinsic level, the geometry bifurcates: in the unimodal case, the landscape is strictly convex, yielding a Gibbs equilibrium; multimodally, a negative symmetric KL coupling induces barrier-driven coordinate responses and can sustain structural separation.  The landscape visualizations are illustrative schematics.}
\label{fig:roadmap}
\end{figure*}

We therefore adopt a measure-theoretic viewpoint that makes the population geometry of InfoNCE explicit (see \Cref{fig:roadmap} for an illustration). In the canonical feature normalization setting, we regard the embedding space $\mathcal{Z}$ as a fixed compact manifold equipped with a volume measure $\mu$. Encoders push data distributions forward onto $\mathcal{Z}$, thereby inducing both representation laws and positive-pair laws directly in the embedding space. Under the exponential similarity kernel, the InfoNCE denominator converges, in the large-batch limit, to a population \emph{partition field}: a kernel-averaged functional of the current representation law that determines the softmax gradient. We show that stochastic InfoNCE is consistent with a closed-form deterministic energy over representation measures, both in value and in gradient. This passage from finite-batch objectives to population energies allows us to separate intrinsic geometric effects from parametrization or training artifacts.

From this intrinsic view, unimodal and symmetric multimodal contrastive learning exhibit a fundamental \emph{geometric bifurcation}. In the unimodal regime, the intrinsic population energy is strictly convex and admits a unique Gibbs equilibrium: at low temperature, mass concentrates in highly aligned regions, while entropy controls dispersion \emph{within} those regions. This reframes ``uniformity'' \citep{wang2020understanding} not as a global force opposing alignment, but as entropy-driven spreading inside the alignment basin. In contrast, symmetric multimodal InfoNCE induces a cross-coupled geometry: heterogeneous modalities can define different conditional laws and hence different alignment fields, producing a negative symmetric divergence coupling between their marginals. Rather than reducing to a shared unimodal-style equilibrium, each modality enters the effective landscape of the other through a density-dependent barrier, yielding barrier-driven coordinate geometry. Exact marginal coincidence thus requires additional compatibility between the two directional structures; absent such compatibility, the cross-coupled geometry can sustain a persistent modality gap even when pairwise alignment is strong. 

These intrinsic results are not merely formal: under suitable encoder expressivity, the corresponding parametric problems inherit these intrinsic concentration and barrier structures up to controllable KDE mismatch, linking the population-level theory back to encoder parametrizations. We validate the resulting predictions through controlled synthetic experiments and measurements on pretrained CLIP representations. The results show that strong retrieval performance can coexist with a measurable modality gap, and that weakening cross-modal compatibility systematically enlarges this gap even when category-level semantics are preserved. Together, these findings suggest that the modality gap is not merely a failure of pairwise alignment, but reflects a population-level geometric effect; reducing it may therefore require controlling cross-modal divergence beyond positive-pair matching alone.

\paragraph{Contributions.} Our main contributions are: (i) large-batch value- and gradient-consistency results connecting stochastic InfoNCE to deterministic population energies; (ii) a characterization of the intrinsic unimodal InfoNCE geometry as a strictly convex population energy with a unique Gibbs equilibrium, together with a reinterpretation of ``uniformity'' as entropy-driven dispersion within the alignment basin; (iii) a derivation of the multimodal bifurcation, in which a negative symmetric divergence coupling induces barrier-driven coordinate geometry and can sustain a modality gap under conditional heterogeneity; (iv) synthetic and real-data evidence supporting these predictions through controlled experiments and measurements on pretrained OpenCLIP models across CNN- and ViT-based backbones.

\section{Problem Setup}
\label{sec:setup}

We introduce the population-level objects used throughout the analysis for unimodal and symmetric multimodal InfoNCE. For ease of reference, \Cref{tab:notation_summary} summarizes the principal notations used throughout the paper.

\paragraph{Data, encoders, and induced laws.}
Let $\mathcal P(\mathcal S)$ denote the set of probability measures on a measurable space $\mathcal S$. In the unimodal setting, $r_{\mathrm{um}}\in\mathcal P(\mathcal X\times\mathcal X)$ governs positive observational pairs $(\mathbf x,\tilde{\mathbf x})$, typically two views of the same underlying example. We assume that its two $\mathcal X$-marginals coincide, and denote the common marginal by $p_{\mathbf x}$. In the multimodal setting, $r_{\mathrm{mm}}\in\mathcal P(\mathcal X\times\mathcal Y)$ governs positive cross-modal pairs $(\mathbf x,\mathbf y)$, where $\mathcal X$ and $\mathcal Y$ are measurable sample spaces; we write $p_{\mathbf x}$ and $p_{\mathbf y}$ for its $\mathcal X$- and $\mathcal Y$-marginals.

We consider two families of measurable encoders $\{f_\theta:\mathcal X\to\mathcal Z\}_{\theta\in\Theta}$, and $\{g_\phi:\mathcal Y\to\mathcal Z\}_{\phi\in\Phi}$, with compact parameter spaces $\Theta$ and $\Phi$, where $\mathcal Z\subset\mathbb R^n$ is a common representation space. We model $\mathcal Z$ as a Riemannian manifold (e.g., $\mathbb S^{n-1}$) with a non-atomic Riemannian volume measure $\mu$. These encoders induce the pushforward laws
\begin{align*}
q_\theta &:= (f_\theta)_\# p_{\mathbf x},&
\pi_{\theta\theta} &:= (f_\theta\times f_\theta)_\# r_{\mathrm{um}},\\
q_\phi &:= (g_\phi)_\# p_{\mathbf y},&
\pi_{\theta\phi} &:= (f_\theta\times g_\phi)_\# r_{\mathrm{mm}}.     
\end{align*}
Here, $q_\theta$ and $q_\phi$ are the encoded marginal laws, while $\pi_{\theta\theta}$ and $\pi_{\theta\phi}$ are the encoded positive-pair laws.

\paragraph{InfoNCE objectives in contrastive learning.}
We consider InfoNCE objectives based on a symmetric bounded measurable similarity function $\mathrm{sim}:\mathcal Z\times\mathcal Z\to\mathbb R$ satisfying $\mathrm{sim}(z,w)=\mathrm{sim}(w,z)$, and define the exponential kernel
\begin{equation}
\label{eq:exp_sim_kernel}
\kappa_\tau(z,w) := \exp\!\big(\mathrm{sim}(z,w)/\tau\big),
\end{equation}
where $\tau>0$ is the temperature. The unimodal and multimodal objectives are defined as follows.

\emph{(i) Unimodal objective.}
For a training batch $\mathcal{B}$ composed of a positive pair $(\mathbf{x},\tilde{\mathbf{x}})\sim r_{\mathrm{um}}$ and $N$ negatives $\{\mathbf{x}_j'\}_{j=1}^N\overset{\mathrm{iid}}{\sim}p_{\mathbf x}$, the expected finite-$N$ InfoNCE objective is\footnote{For analytical clarity, we use an iid negative-sampling formulation. Extending the formal large-batch argument to dependent in-batch negatives requires additional sampling assumptions and is not considered here.}
\begin{equation}
\label{eq:asym_loss_def}
\begin{aligned}
&\mathcal L_{\mathrm{NCE},N}(\theta) :=\\ 
&\mathbb{E}_{\mathcal B}\left[
-\log
\frac{\kappa_\tau(f_\theta(\mathbf x),f_\theta(\tilde{\mathbf x}))}
{\kappa_\tau(f_\theta(\mathbf x),f_\theta(\tilde{\mathbf x}))+\sum_{j=1}^N \kappa_\tau(f_\theta(\mathbf x),f_\theta(\mathbf x_j'))}
\right]. 
\end{aligned}
\end{equation}

\emph{(ii) Symmetric multimodal objective.}
Given a cross-modal positive pair $(\mathbf x,\mathbf y)\sim r_{\mathrm{mm}}$ and negatives $\{\mathbf y_j'\}_{j=1}^N\overset{\mathrm{iid}}{\sim}p_{\mathbf y}$, define the expected directional objective
\begin{equation*}
\begin{aligned}
&\mathcal{L}_{\mathrm{NCE},N}^{x\shortto y}(\theta,\phi)
:=
\mathbb{E}_{\mathcal B_{x\shortto y}}\\
&\Bigg[
-\log
\frac{\kappa_\tau\big(f_\theta(\mathbf x),g_\phi(\mathbf y)\big)}
{\kappa_\tau\big(f_\theta(\mathbf x),g_\phi(\mathbf y)\big)
+\sum_{j=1}^N \kappa_\tau\big(f_\theta(\mathbf x),g_\phi(\mathbf y_j')\big)}
\Bigg],
\end{aligned}
\end{equation*}
where $\mathcal B_{x\shortto y}:=\{\mathbf x,\mathbf y,\mathbf y_1',\dots,\mathbf y_N'\}$. Define $\mathcal L_{\mathrm{NCE},N}^{y\shortto x}(\phi,\theta)$ analogously by swapping modalities. The symmetric multimodal InfoNCE objective is
\begin{equation}
\label{eq:sym_loss_def}
\mathcal{L}_{\mathrm{Sym},N}(\theta,\phi)
:=
\tfrac{1}{2}\big(
\mathcal{L}_{\mathrm{NCE},N}^{x\shortto y}(\theta,\phi)
+
\mathcal{L}_{\mathrm{NCE},N}^{y\shortto x}(\phi,\theta)
\big).
\end{equation}

\paragraph{Population partition fields and smoothed densities.} 
At the population level, the InfoNCE denominator induces kernel-averaged fields on $\mathcal Z$. We use these fields to define the corresponding smoothed representation densities. To normalize them, we impose the following condition.

\begin{assumption}[Constant kernel volume]
\label{ass:kernel_volume}
The representation space $\mathcal{Z}$ is compact with $\mu(\mathcal{Z})<\infty$. For each $\tau>0$, there exists $V_\kappa(\tau)\in(0,\infty)$ such that
\begin{equation}
\label{eq:constant_volume}
\int_\mathcal{Z}\kappa_\tau(z,w)\,\mu(\mathrm{d}w)
= V_\kappa(\tau)
\qquad
\text{for all }z\in\mathcal{Z}.
\end{equation}
\end{assumption}

\begin{remark}[When \Cref{ass:kernel_volume} holds]
\label{rem:kernel_volume_generality}
A sufficient condition for \Cref{ass:kernel_volume} is that $\mathcal{Z}$ is a compact homogeneous Riemannian manifold, $\mu$ is its invariant volume measure, and $\kappa_\tau$ is isotropic, i.e., depends only on geodesic distance. In that case, $V_\kappa(\tau,z):=\int_\mathcal{Z}\kappa_\tau(z,w)\,\mu(\mathrm{d}w)$ is independent of $z$, so \Cref{ass:kernel_volume} holds exactly.
\end{remark}

\begin{definition}[Population partition fields]
\label{def:partition_function}
Given pushforward laws $q_\theta$ and $q_\phi$, define the population \emph{partition fields}
\[
\begin{aligned}
\Gamma_{\theta,\tau}(z)
:=
\int_\mathcal{Z}
\kappa_\tau(z,w)\,q_\theta(\mathrm{d}w),
\\
\Gamma_{\phi,\tau}(z)
:=
\int_\mathcal{Z}
\kappa_\tau(z,w)\,q_\phi(\mathrm{d}w).
\end{aligned}
\]
\end{definition}

\begin{definition}[Kernel-smoothed representation density]
\label{def:smoothed_density}
Under \Cref{ass:kernel_volume}, define the kernel-smoothed representation densities
$\tilde{\rho}_{\theta,\tau}, \tilde{\rho}_{\phi,\tau}: \mathcal{Z} \to (0,\infty)$ by
\[
\tilde{\rho}_{\theta,\tau}(z)
:=
\frac{\Gamma_{\theta,\tau}(z)}{V_\kappa(\tau)},
\qquad
\tilde{\rho}_{\phi,\tau}(z)
:=
\frac{\Gamma_{\phi,\tau}(z)}{V_\kappa(\tau)}.
\]
By symmetry of $\kappa_\tau$ and \Cref{ass:kernel_volume}, both densities integrate to one with respect to $\mu$. Equivalently, $\tilde{\rho}_{\theta,\tau}(z)\,\mu(\mathrm dz)$ and $\tilde{\rho}_{\phi,\tau}(z)\,\mu(\mathrm dz)$ are the kernel-smoothed probability measures on $\mathcal Z$.
\end{definition}

\begin{remark}[Finite-$\tau$ smoothing and positivity]
\label{rem:smoothing_positivity}
For any $\tau>0$, $\tilde{\rho}_{\theta,\tau}$ is a kernel smoothing of $q_\theta$ via $\Gamma_{\theta,\tau}$. In particular, it defines a $\mu$-density even when $q_\theta$ is singular with respect to $\mu$. If $\kappa_\tau(z,w)$ is continuous and strictly positive on $\mathcal Z\times\mathcal Z$, then $\tilde{\rho}_{\theta,\tau}$ is continuous and strictly positive on $\mathcal Z$; the same holds for $\tilde{\rho}_{\phi,\tau}$.
\end{remark}

Here, $\Gamma_{\theta,\tau}(z)$ is the population negative-partition field associated with a query at $z$, while $\tilde{\rho}_{\theta,\tau}$ is the corresponding kernel-smoothed density of $q_\theta$ at scale $\tau$. For i.i.d.\ negative representation variables
$\mathbf w_j\sim q_\theta$,
\(
\frac{1}{N}\sum_{j=1}^{N}\kappa_\tau(z,\mathbf w_j)
\)
is a Monte Carlo estimator of $\Gamma_{\theta,\tau}(z)$; analogously, $\Gamma_{\phi,\tau}$ arises when negatives are drawn from $q_\phi$. These population fields, or equivalently their normalized densities under \Cref{ass:kernel_volume}, are the quantities that enter the large-batch parametric energies and pointwise gradient limits developed in the following sections.

\section{Unimodal InfoNCE: Convex Geometry and Gibbs Equilibrium}
\label{sec:unimodal}

This section develops the unimodal baseline that underlies the later multimodal bifurcation. We first show that, in the large-batch regime, unimodal InfoNCE admits an explicit deterministic energy limit depending on $\theta$ only through the induced laws $(q_\theta,\pi_{\theta\theta})$ on $\mathcal Z$ and $\mathcal Z\times\mathcal Z$. Proofs for all formal statements in this section are deferred to \Cref{app:proofs_unimodal}.

\subsection{Large-Batch Deterministic Energy}
We begin by identifying the population-level energy arising as the large-batch limit of unimodal InfoNCE. The resulting closed-form parametric energy $\mathcal J_\tau(\theta)$ has a gradient matching the InfoNCE gradient up to vanishing error as negative sample size $N\to\infty$.

\begin{assumption}[Optimization regularity]
\label{assump:regularity}
Assume the following conditions hold:
\begin{enumerate}[label=\textup{(\roman*)}, wide=0pt, topsep=0pt]
\item \label{itm:enc_reg} {(Encoder regularity).}
For every $\mathbf x\in\mathcal X$, the map $\theta\mapsto f_\theta(\mathbf x)$ is $C^1$ on $\Theta$, and its Jacobian is uniformly bounded:\footnote{Throughout, $\|\cdot\|$ denotes the $\ell_2$ norm for vectors unless specified; $\|\cdot\|_{\mathrm{op}}$ is the induced operator norm for linear maps.}
\[
\sup\nolimits_{\theta\in\Theta}\sup\nolimits_{\mathbf x\in\mathcal X}
\|J_\theta f_\theta(\mathbf x)\|_{\mathrm{op}}<\infty.
\]
\item \label{itm:crit_reg} {(Similarity regularity).}
The similarity function $\mathrm{sim}$ is $C^1$ on an open neighborhood of $\mathcal Z\times\mathcal Z$, and its input gradient is uniformly bounded on $\mathcal Z\times\mathcal Z$:
\[
\sup\nolimits_{(z,w)\in\mathcal Z\times\mathcal Z}
\|\nabla \mathrm{sim}(z,w)\|<\infty.
\]
\end{enumerate}
\end{assumption}

\begin{remark}[Optimization stability]
\Cref{assump:regularity} ensures that InfoNCE gradients remain uniformly controlled over $\Theta$, which is needed to justify exchanging the large-batch limit with differentiation. Further discussion and the formal uniform bounds are given in \Cref{app:critic_justification} and \Cref{lem:asym_grad_bound}.
\end{remark}

For fixed $\tau$, the normalized finite-$N$ negative partition sum concentrates around the population partition field $\Gamma_{\theta,\tau}$ from \Cref{def:partition_function}. This yields a closed-form population energy with value and gradient consistency in the large-batch limit.

\begin{definition}[Alignment potential field]
\label{def:align_potential}
Let $\pi_{\theta\theta}\in\mathcal P(\mathcal Z\times\mathcal Z)$ be the encoded positive-pair law with first marginal $q_\theta\in\mathcal P(\mathcal Z)$. By disintegration, there exists a measurable family of probability measures $\{\nu_{\theta,z}\}_{z\in\mathcal Z}\subset\mathcal P(\mathcal Z)$ such that
\[
\pi_{\theta\theta}(\mathrm dz,\mathrm dw)
=
q_\theta(\mathrm dz)\,\nu_{\theta,z}(\mathrm dw)
\]
as measures on $\mathcal Z\times\mathcal Z$. We define the alignment potential field by
\begin{equation}
\label{eq:alignment_potential}
U_\theta(z)
:=
-\int_{\mathcal Z}
\mathrm{sim}(z,w)\,\nu_{\theta,z}(\mathrm dw),
\qquad
q_\theta\text{-a.e. }z.
\end{equation}
\end{definition}

\begin{definition}[Unimodal parametric energy]
\label{def:uni_gef}
Fix $\tau>0$. Let $\tilde\rho_{\theta,\tau}$ be the kernel-smoothed $\mu$-density from \Cref{def:smoothed_density}, and let $U_\theta$ be the alignment potential field from \Cref{def:align_potential}. Define the unimodal parametric energy $\mathcal J_\tau:\Theta\to\mathbb R$ by
\begin{equation}
\label{eq:uni_gef_def}
\mathcal J_\tau(\theta)
:=
\frac{1}{\tau}\int_{\mathcal Z}
U_\theta(z)\,q_\theta(\mathrm dz)
-\;
H_\times(q_\theta,\tilde\rho_{\theta,\tau}),
\end{equation}
where $H_\times(q,\tilde\rho):=-\int_{\mathcal Z} \log\tilde\rho(z)\,q(\mathrm dz)$ denotes the cross-entropy of the probability measure $q\in\mathcal P(\mathcal Z)$ against a strictly positive density $\tilde\rho$.
\end{definition}

\begin{restatable}[Large-batch unimodal consistency]{theorem}{thmnceconsistency}
\label{thm:nce_consistency}
Consider the unimodal InfoNCE objective $\mathcal L_{\mathrm{NCE},N}(\theta)$ in \Cref{eq:asym_loss_def} and the parametric energy $\mathcal J_\tau(\theta)$ in \Cref{def:uni_gef}. Assume \Cref{ass:kernel_volume,assump:regularity}. Then for any fixed $\tau>0$ and $\theta\in\Theta$, as $N\to\infty$,
\begin{align*}
\big|
\mathcal L_{\mathrm{NCE},N}(\theta)
-\mathcal J_\tau(\theta)
-\log\!\big(NV_\kappa(\tau)\big)
\big|
\longrightarrow 0,
\\
\big\|
\nabla_\theta\mathcal L_{\mathrm{NCE},N}(\theta)
-\nabla_\theta\mathcal J_\tau(\theta)
\big\|
\longrightarrow 0.
\end{align*}
\end{restatable}

\Cref{thm:nce_consistency} shows that, at each fixed parameter, the expected finite-$N$ InfoNCE objective and its gradient converge to the deterministic energy $\mathcal J_\tau$ and its gradient, up to the additive constant $\log\!\big(NV_\kappa(\tau)\big)$ in value. Thus, the partition field $\Gamma_{\theta,\tau}$, or equivalently the smoothed density $\tilde\rho_{\theta,\tau}$, determines the pointwise large-batch energy and gradient limit.

\subsection{Intrinsic Free Energy and Gibbs Equilibrium}
\label{sec:unimodal_intrinsic}

The parametric energy $\mathcal J_\tau(\theta)$ is generally nonconvex in $\theta$, since the map $\theta\mapsto(q_\theta,\pi_{\theta\theta})$ is implicit. To isolate the intrinsic geometry, we fix a potential field $U$ on $\mathcal Z$ and optimize directly over distributions on $\mathcal Z$. This yields a strictly convex free-energy functional with a unique Gibbs equilibrium. In the low-temperature regime, $\mathcal J_\tau$ approaches this intrinsic landscape up to a controllable KDE mismatch.

\begin{definition}[Intrinsic unimodal energy]
\label{def:uni_free_energy}
Fix $\tau>0$ and a Borel measurable potential $U:\mathcal Z\to\mathbb R$ that is bounded below
$\mu$-a.e. Let
\[
\mathcal D_\mu(\mathcal Z)
:=
\Big\{
\rho\ge0:
\int_{\mathcal Z}\rho(z)\,\mu(\mathrm dz)=1
\Big\}.
\] 
We define the intrinsic unimodal energy $\mathcal F_{\tau,U}:\mathcal D_\mu(\mathcal Z)\to(-\infty,+\infty]$ by
\begin{equation}
\label{eq:free_energy_def}
\mathcal F_{\tau,U}(\rho)
:=
\frac{1}{\tau}
\int_{\mathcal Z}U(z)\rho(z)\,\mu(\mathrm dz)
-
H(\rho),
\end{equation}
where $H(\rho):=-\int_{\mathcal Z}\rho(z)\log\rho(z)\,\mu(\mathrm dz)$ is the differential entropy of $\mu$-density $\rho$. We adopt $0\log 0:=0$, and set $\mathcal F_{\tau,U}(\rho)=+\infty$ if either $\int_{\mathcal Z}U(z)\rho(z)\,\mu(\mathrm dz)$ or $\int_{\mathcal Z}\rho(z)\log\rho(z)\,\mu(\mathrm dz)$ equals $+\infty$.
\end{definition}

\begin{restatable}[Unique Gibbs equilibrium]{proposition}{propgibbsequilibrium}
\label{prop:gibbs_equilibrium}
Assume $\mathcal Z$ is compact with $\mu(\mathcal Z)<\infty$. Fix $\tau>0$ and let $U:\mathcal{Z}\to\mathbb R$ be Borel measurable and bounded below $\mu$-a.e. Then, $\mathcal F_{\tau,U}$ is strictly convex on its effective domain
and has the unique Gibbs-form minimizer $\rho^*(z)
=
\frac{1}{Z_\tau}\exp\!\left(-U(z)/\tau\right)$, with $Z_\tau
:=
\int_{\mathcal Z}
\exp\!\left(-\frac{U(z)}{\tau}\right)\,\mu(\mathrm dz)$.
\end{restatable}

\begin{restatable}[Low-temperature concentration]{proposition}{propgroundstate}
\label{prop:ground_state}
In the setting of \Cref{prop:gibbs_equilibrium}, fix $\sigma>0$ and define
\[
\mathcal W^\sigma
:=
\big\{
z\in\mathcal Z:
U(z)
\le
\operatorname*{ess\,inf}\nolimits_{\mu}U+\sigma
\big\}.
\]
Then, as $\tau\to0^+$, the Gibbs measure $\rho^*(z)\,\mu(\mathrm dz)$ concentrates on $\mathcal W^\sigma$ in the sense that, for every open $\mathcal O\supset\mathcal W^\sigma$,
\(
\int_{\mathcal Z\setminus\mathcal O}
\rho^*(z)\,\mu(\mathrm dz)
\longrightarrow0.
\)
\end{restatable}

In the unimodal regime, the intrinsic landscape admits a unique Gibbs equilibrium for each $\tau$, and as $\tau\!\to\! 0^+$, the equilibrium concentrates on near-minimizers of the alignment potential $U$. 
\Cref{prop:ground_state} formalizes this low-temperature concentration, while \Cref{fig:ground_state} illustrates it by tracking the mass inside a fixed near-minimizer region as $\tau$ decreases.

\paragraph{Connecting back to $\mathcal J_\tau$: sharp kernels control KDE mismatch.}
To connect the intrinsic geometry back to the parametric energy, we consider a sharp-kernel regime in which the kernel-smoothed density $\tilde\rho_{\theta,\tau}$ approximates the induced representation density $\rho_\theta$.

\begin{assumption}[Sharp diagonal peak]
\label{ass:sharp_peak}
Assume \Cref{ass:kernel_volume}.
Let $d_\mathcal Z(\cdot,\cdot)$ denote the geodesic distance on
$\mathcal Z$. There exist constants $r>0$ and
$0<m_1\le m_2<\infty$ such that for all $z\in\mathcal Z$ and
all $w\in\mathcal Z$ satisfying $d_\mathcal Z(z,w)<r$,
\begin{equation}
\label{eq:sharp_peak_quad}
\begin{aligned}
-m_2d_\mathcal Z(z,w)^2
\!\le\!
\mathrm{sim}(z,w)\!-\!\mathrm{sim}(z,z)
\!\le\!
-m_1d_\mathcal Z(z,w)^2,
\\
\text{and}\quad
\operatorname*{arg\,max}_{w\in\mathcal Z}
\mathrm{sim}(z,w)
=
\{z\},
\; z\in\mathcal Z.
\end{aligned}
\end{equation}
\end{assumption}

\begin{remark}
Geometrically, \Cref{ass:sharp_peak} forces the normalized kernel mass to concentrate within an $O(\sqrt{\tau})$ geodesic neighborhood of the diagonal; see \Cref{app:sharp_peak_interpretation} for discussion. As a result, $\tilde\rho_{\theta,\tau}$ acts as a local smoothing of $\rho_\theta$, which is quantified in the low-temperature regime by \Cref{lem:sharp_peak_smoothing}.
\end{remark}

\begin{restatable}[Low-temperature unimodal energy consistency]{theorem}{thmintrinsicunimodal}
\label{thm:intrinsic_unimodal}
Assume \Cref{ass:kernel_volume,assump:regularity,ass:sharp_peak}. Fix $\theta\in\Theta$ and assume $q_\theta\ll\mu$ with a strictly positive, continuous density $\rho_\theta$, i.e., $\inf_{z\in\mathcal Z}\rho_\theta(z)\ge\underline\rho_\theta>0$.\footnote{See \Cref{app:density_floor} for a detailed discussion on the strict positivity assumption and the role of density floors.} Define the KDE mismatch 
\begin{equation}
\varepsilon_{\mathrm{kde}}^{(\theta)}(\tau):=\|\tilde\rho_{\theta,\tau}-\rho_\theta\|_\infty.
\end{equation}
Then there exists $\tau_0(\theta)>0$ such that
\[
\bigl|\mathcal{J}_\tau(\theta)-\mathcal{F}_{\tau,U_\theta}(\rho_\theta)\bigr|
\;\le\;
\frac{2\,\varepsilon_{\mathrm{kde}}^{(\theta)}(\tau)}{\underline\rho_\theta}, 
\qquad \forall\,0<\tau\le\tau_0(\theta),
\]
and in particular, as $\tau\to 0^+$, $\bigl|\mathcal{J}_\tau(\theta)-\mathcal{F}_{\tau,U_\theta}(\rho_\theta)\bigr|\longrightarrow 0$. 
\end{restatable}

\paragraph{Uniformity as a tie-breaker.}
\Cref{thm:intrinsic_unimodal} bridges the intrinsic analysis back to the parametric objective: in the sharp-kernel regime, the parametric energy $\mathcal J_\tau(\theta)$ is close to the intrinsic energy $\mathcal F_{\tau,U_\theta}(\rho_\theta)$ up to the KDE mismatch between $\tilde\rho_{\theta,\tau}$ and $\rho_\theta$. If the encoder family is expressive enough to realize the relevant intrinsic Gibbs equilibrium at the best attainable potential level, then minimizers of $\mathcal J_\tau$ inherit the same ground-state concentration behavior. Within an aligned basin where $U_\theta$ is already near-minimized, the remaining degree of freedom is how probability mass is distributed, and the entropy term in \Cref{eq:free_energy_def} selects the most internally dispersed near-optimal configuration. This is the sense in which ``uniformity'' emerges. \Cref{cor:unimodal_parametric_inheritance} formalizes the expressivity condition and the resulting parametric inheritance; see \Cref{app:unimodal_inheritance} for extended formal statements.

\begin{figure}
    \centering
    \includegraphics[width=0.95\linewidth]{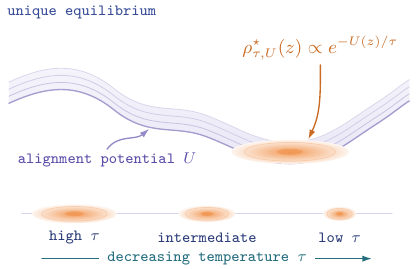}
    \caption{In the unimodal case, alignment and entropy define a free energy with a unique Gibbs equilibrium, $\rho_{\tau,U}^{\star}\propto e^{-U/\tau}$, whose mass increasingly concentrates near low-potential regions as $\tau\downarrow0$.}
    \label{fig:unimodalgeo}
\end{figure}

\begin{takeaway}{Unimodal InfoNCE geometry}\label{take:unimodal}
The defining feature of unimodal contrastive learning is structural cohesion. Because the modality evaluates its cross-entropy against its \emph{own} smoothed density field, the intrinsic energy is strictly convex. Alignment determines the ground-state basin, while entropy selects a unique, internally dispersed equilibrium within it (see \Cref{fig:unimodalgeo}).
\end{takeaway}

\section{Symmetric Multimodal InfoNCE: Divergence-Coupled Geometry}
\label{sec:multimodal}

We analyze the symmetric multimodal InfoNCE objective used in CLIP-style training (\Cref{eq:sym_loss_def}). Unlike the unimodal setting, the geometry is \emph{cross-coupled}: each modality induces the effective field seen by the other. This coupling yields a qualitatively different geometry and can sustain a population-level modality gap. Proofs for all formal statements in this section are deferred to \Cref{app:proofs_multimodal}.

\subsection{Cross-Coupled Multimodal Deterministic Energy}

We now identify the deterministic population energy arising as the large-batch limit of symmetric multimodal InfoNCE. As in the unimodal case, the expected finite-$N$ objective matches a closed-form parametric energy up to an additive large-batch constant in value, with pointwise gradient consistency. The difference is cross-coupling: each modality evaluates its cross-entropy against the other modality's kernel-smoothed density field.

\begin{assumption}[Multimodal optimization regularity]
\label{assump:mm_regularity}
In multimodal contrastive learning, assume \Cref{assump:regularity} holds for both encoder families $\{f_\theta\}_{\theta\in\Theta}$ and $\{g_\phi\}_{\phi\in\Phi}$, uniformly over the joint parameter space $\Theta\times\Phi$.
\end{assumption}

\begin{remark}
Under \Cref{assump:mm_regularity}, the symmetric InfoNCE objective has a uniformly bounded joint gradient over $\Theta\times\Phi$, which is needed to exchange the large-batch limit with differentiation in the multimodal consistency proof. See \Cref{lem:sym_grad_bound} for the formal multimodal gradient bounds.
\end{remark}

\begin{definition}[Directional alignment potential fields]
\label{def:directional_potential}
Let $\pi_{\theta\phi}$ denote the encoded positive-pair law on $\mathcal Z\times\mathcal Z$ with marginals $q_\theta$ and $q_\phi$. By disintegration, there exist measurable families of probability measures
$\{\nu_{\theta\phi,z}\}_{z\in\mathcal Z}$ and $\{\nu_{\phi\theta,w}\}_{w\in\mathcal Z}$ such that
\[
\pi_{\theta\phi}(\mathrm dz,\mathrm dw)
=
q_\theta(\mathrm dz)\,\nu_{\theta\phi,z}(\mathrm dw)
=
q_\phi(\mathrm dw)\,\nu_{\phi\theta,w}(\mathrm dz),
\]
as measures on $\mathcal Z\times\mathcal Z$. Define the directional alignment potential fields
\[
\begin{aligned}
U_{\theta\shortto\phi}(z)
&:=
-\int_{\mathcal Z}
\mathrm{sim}(z,w)\,\nu_{\theta\phi,z}(\mathrm dw),
\; q_\theta\text{-a.e. }z,
\\
U_{\phi\shortto\theta}(w)
&:=
-\int_{\mathcal Z}
\mathrm{sim}(z,w)\,\nu_{\phi\theta,w}(\mathrm dz),
\; q_\phi\text{-a.e. }w.  
\end{aligned}
\]
\end{definition}

\begin{definition}[Symmetric multimodal parametric energy]
\label{def:sym_gef}
Fix $\tau>0$. Recall the kernel-smoothed $\mu$-densities $\tilde\rho_{\theta,\tau},\tilde\rho_{\phi,\tau}$ from \Cref{def:smoothed_density}. Define the directional multimodal energies by
\[
\begin{aligned}
\mathcal J_\tau^{x\shortto y}(\theta,\phi)
&:=
\frac{1}{\tau}
\int_{\mathcal Z}
U_{\theta\shortto\phi}(z)\,q_\theta(\mathrm dz)
-
H_\times\!\big(q_\theta,\tilde\rho_{\phi,\tau}\big),
\\
\mathcal J_\tau^{y\shortto x}(\phi,\theta)
&:=
\frac{1}{\tau}
\int_{\mathcal Z}
U_{\phi\shortto\theta}(w)\,q_\phi(\mathrm dw)
-
H_\times\!\big(q_\phi,\tilde\rho_{\theta,\tau}\big).  
\end{aligned}
\]
and the symmetric multimodal parametric energy by
\begin{equation}
\label{eq:sym_gef_def}
\mathcal J_\tau^{\mathrm{Sym}}(\theta,\phi)
:=
\frac12\left(
\mathcal J_\tau^{x\shortto y}(\theta,\phi)
+
\mathcal J_\tau^{y\shortto x}(\phi,\theta)
\right).
\end{equation}
\end{definition}

\begin{remark}[Directional asymmetry]
\label{rem:directional_asymmetry}
Although both directions are induced by the same pair law $\pi_{\theta\phi}$, the two regular conditional laws
$\nu_{\theta\phi,z}$ and $\nu_{\phi\theta,w}$ need not coincide. Consequently, $U_{\theta\shortto\phi}$ and $U_{\phi\shortto\theta}$ need not coincide: there is no single unimodal-style potential field unless an additional compatibility condition is imposed on the conditional laws.
\end{remark}

\begin{restatable}[Large-batch multimodal consistency]{theorem}{thmsymconsistency}
\label{thm:sym_consistency}
Consider the symmetric multimodal InfoNCE objective $\mathcal L_{\mathrm{Sym},N}(\theta,\phi)$ in \Cref{eq:sym_loss_def} and the energy $\mathcal J^{\mathrm{Sym}}_\tau(\theta,\phi)$ in \Cref{def:sym_gef}. Assume \Cref{ass:kernel_volume,assump:mm_regularity}. Then, for any fixed $\tau>0$ and $(\theta,\phi)\in\Theta\times\Phi$, as $N\to\infty$,
\[
\begin{aligned}
\big|\mathcal L_{\mathrm{Sym},N}(\theta,\phi)-\mathcal J^{\mathrm{Sym}}_\tau(\theta,\phi)-\log\!\big(NV_\kappa(\tau)\big)\big|\longrightarrow 0,
\\
\big\|\nabla_{(\theta,\phi)}\mathcal L_{\mathrm{Sym},N}(\theta,\phi)-\nabla_{(\theta,\phi)}\mathcal J^{\mathrm{Sym}}_\tau(\theta,\phi)\big\|\longrightarrow 0.
\end{aligned}
\]
\end{restatable}

\begin{remark}
Since $\log\!\big(NV_\kappa(\tau)\big)$ is independent of $(\theta,\phi)$, \Cref{thm:sym_consistency} shows that, at each fixed parameter, the expected finite-$N$ symmetric InfoNCE objective converges to $\mathcal J_\tau^{\mathrm{Sym}}$ up to this constant, while its gradient converges directly to $\nabla_{(\theta,\phi)} \mathcal J_\tau^{\mathrm{Sym}}$. Unlike the unimodal case, the limiting energy is cross-coupled: each modality evaluates its cross-entropy against the other's smoothed density field.
\end{remark}

\subsection{Cross-Modal Divergence Coupling and Barrier-Driven Coordinate Geometry}
\label{sec:mm_intrinsic}

We lift it to an intrinsic functional over modality \emph{densities} on $\mathcal Z$. Relative to the unimodal free energy, the key new term is a \emph{negative} symmetric divergence coupling. This coupling persists at fixed temperature and qualitatively changes the coordinate-wise geometry.

\begin{definition}[Intrinsic multimodal energy]
\label{def:mm_free_energy}
Fix $\tau>0$ and let $U_{1\shortto2},U_{2\shortto1}:\mathcal Z\to\mathbb R$ be fixed bounded Borel measurable potential fields. Fix density floors $\underline\rho_1,\underline\rho_2>0$ satisfying $\underline\rho_i\,\mu(\mathcal Z)<1$, and define
\[
\mathcal D_{\mu,\underline\rho_i}(\mathcal Z)
:=
\big\{
\rho\in\mathcal D_\mu(\mathcal Z)\cap L^\infty(\mu):
\rho\ge\underline\rho_i\ \mu\text{-a.e.}
\big\}.
\]
For $(\rho_1,\rho_2)\in\mathcal D_{\mu,\underline\rho_1}(\mathcal Z)\times\mathcal D_{\mu,\underline\rho_2}(\mathcal Z)$, define the intrinsic multimodal energy by
\begin{equation}
\label{eq:mm_free_energy}
\begin{aligned}
\mathcal{F}&^\mathrm{Sym}_{\tau,\mathbf U_{1,2}}(\rho_1,\rho_2):=\\
&\frac{1}{2}\big(\mathcal F_{\tau,U_{1\shortto2}}(\rho_1)+\mathcal F_{\tau,U_{2\shortto1}}(\rho_2)\big)
- D_\mathrm{KL}^\mathrm{Sym}(\rho_1,\rho_2), 
\end{aligned}
\end{equation}
where
$\mathbf U_{1,2}:=(U_{1\shortto2},U_{2\shortto1})$,
$\mathcal F_{\tau,U_{1\shortto2}}$ and
$\mathcal F_{\tau,U_{2\shortto1}}$
are the unimodal free energies from \Cref{eq:free_energy_def}, and
\[
D_{\mathrm{KL}}^{\mathrm{Sym}}(\rho_1,\rho_2)
:=
\frac12
\left(
D_{\mathrm{KL}}(\rho_1\|\rho_2)
+
D_{\mathrm{KL}}(\rho_2\|\rho_1)
\right),
\]
with $D_{\mathrm{KL}}(\rho_1\|\rho_2)
:=
\int_{\mathcal Z}
\rho_1(z)
\log({\rho_1(z)}/{\rho_2(z)})
\,\mu(\mathrm dz)$.
\end{definition}

\begin{remark}
The constraint $\rho_i\ge \underline\rho_i$ is a technical device that rules out degenerate $-\infty$ energies and makes the coordinate-wise geometry explicit. See \Cref{app:density_floor} for a discussion. 
\end{remark}

\begin{restatable}[Barrier coordinate geometry and extremal convergence]{proposition}{propmmcoordinateextremal}
\label{prop:mm_coordinate_extremal}
Assume $\mathcal Z$ is compact with $\mu(\mathcal Z)<\infty$, and let $\mathcal F^{\mathrm{Sym}}_{\tau,\mathbf U_{1,2}}$ be as in \Cref{def:mm_free_energy}. Fix $\rho_2\in\mathcal D_{\mu,\underline\rho_2}(\mathcal Z)$ and define
\[
V_{1\mid2}(z):=\frac{1}{\tau}U_{1\shortto2}(z)+\log\rho_2(z).
\]
Then, the coordinate map $\rho_1\mapsto \mathcal F^{\mathrm{Sym}}_{\tau,\mathbf U_{1,2}}(\rho_1,\rho_2)$ is concave on $\mathcal D_{\mu,\underline\rho_1}(\mathcal Z)$, and its infimum is approached by densities that place the excess mass
\(
M_{\mathrm{ex}}^{(1)}:=1-\underline\rho_1\,\mu(\mathcal Z)
\)
on approximate minimizers of $V_{1\mid2}$ while leaving only the floor $\underline\rho_1$ elsewhere. An analogous statement for \(M_{\mathrm{ex}}^{(2)}:= 1-\underline\rho_2\,\mu(\mathcal Z)\) holds when optimizing in $\rho_2$ with $\rho_1$ fixed, with effective potential field
\(
V_{2\mid1}(z):=\frac{1}{\tau}U_{2\shortto1}(z)+\log\rho_1(z)
\). (The explicit extremal construction is given in \Cref{app:mm_coordinate_extremal}.)
\end{restatable}

\begin{remark}[Barrier-driven coordinate geometry]
Unlike the intrinsic unimodal free energy, the multimodal functional does not admit Gibbs-type coordinate minimization. The term $\log \rho_2$ enters $V_{1|2}=\frac{1}{\tau}U_{1\shortto 2}+\log\rho_2$ as an occupied-mass barrier: regions where modality 2 already assigns high density become more expensive for modality 1, whereas low-density regions of modality 2 are comparatively favored, subject to the alignment potential. Hence the coordinate infimum is approached by extremal sequences: after maintaining the density floor, modality 1 can approach the infimum by concentrating its excess mass near minimizers of $V_{1\mid2}$, and symmetrically for modality 2. This coordinate-wise avoidance of occupied high-density regions, together with alignment-dependent basin selection, yields a winner-take-all-like extremal geometry.
\end{remark}

\begin{restatable}[Low-temperature multimodal energy consistency]{theorem}{thmintrinsicmultimodal}
\label{thm:intrinsic_multimodal}
Assume \Cref{ass:kernel_volume,assump:mm_regularity,ass:sharp_peak}. Fix any $(\theta,\phi)\in\Theta\times\Phi$, and assume $q_\theta\ll\mu$, $q_\phi\ll\mu$ with continuous densities $\rho_\theta,\rho_\phi$ bounded below: \(\inf_{z\in\mathcal Z}\rho_\theta(z)\ge\underline\rho_\theta>0\), \(\inf_{z\in\mathcal Z}\rho_\phi(z)\ge\underline\rho_\phi>0\). Let \(\varepsilon_{\mathrm{kde}}^{(\theta)}(\tau):=\|\tilde\rho_{\theta,\tau}-\rho_\theta\|_\infty\) and \(\varepsilon_{\mathrm{kde}}^{(\phi)}(\tau):=\|\tilde\rho_{\phi,\tau}-\rho_\phi\|_\infty\).
Then, there exists $\tau_0(\theta,\phi)>0$ such that for all $0<\tau\le\tau_0(\theta,\phi)$,
\[
\Big|\mathcal J^{\mathrm{Sym}}_\tau(\theta,\phi)-\mathcal F^{\mathrm{Sym}}_{\tau,\mathbf U_{\theta,\phi}}(\rho_\theta,\rho_\phi)\Big|
\le
\frac{\varepsilon_{\mathrm{kde}}^{(\theta)}(\tau)+\varepsilon_{\mathrm{kde}}^{(\phi)}(\tau)}
{\min\{\underline\rho_\theta,\underline\rho_\phi\}},
\]
where $\mathcal F^{\mathrm{Sym}}_{\tau,\mathbf U_{\theta,\phi}}(\rho_\theta,\rho_\phi)$ is the intrinsic multimodal energy from \Cref{def:mm_free_energy}, evaluated on \(\mathcal D_{\mu,\underline\rho_\theta}(\mathcal Z)\times \mathcal D_{\mu,\underline\rho_\phi}(\mathcal Z)\), with 
\(
\mathbf U_{\theta,\phi}:=(U_{\theta\shortto\phi},U_{\phi\shortto\theta})
\). In particular, 
\[
\left|\mathcal J^{\mathrm{Sym}}_\tau(\theta,\phi)-\mathcal F^{\mathrm{Sym}}_{\tau,\mathbf U_{\theta,\phi}}(\rho_\theta,\rho_\phi)\right|
\rightarrow 0,  \qquad \text{as}\;\tau\to0^+.
\]
\end{restatable}

The limit $\tau\to0^+$ controls only the KDE mismatches, driving $\tilde\rho_{\cdot,\tau}\to\rho_{\cdot}$; it does not remove the divergence term $D_\mathrm{KL}^\mathrm{Sym}(\rho_\theta,\rho_\phi)$. Because this term enters \Cref{def:mm_free_energy} with a negative sign, it remains an explicit separation-inducing contribution as the KDE mismatch vanishes. \Cref{fig:divergence_persist_main} illustrates the associated marginal separation under increasing levels of cross-modal pairing misalignment.

\paragraph{Parametric inheritance under model expressivity.}
Combining \Cref{thm:intrinsic_multimodal} with the barrier coordinate geometry in \Cref{prop:mm_coordinate_extremal} yields a coordinate-wise parametric inheritance statement: under a suitable model-expressivity condition and with one peer encoder fixed, minimizers of the parametric energy concentrate their excess mass near minimizers of the corresponding effective barrier potential. \Cref{cor:multimodal_inheritance} formalizes this condition and establishes the symmetric coordinate-wise inheritance result; see \Cref{app:multimodal_inheritance} for the extended statements.

\begin{figure}[tbp]
    \centering
    \includegraphics[width=0.95\linewidth]{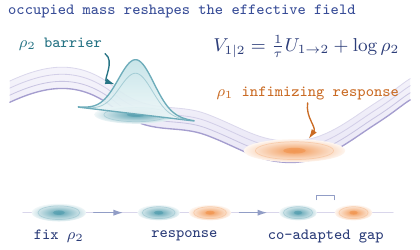}
    \caption{In the multimodal problem, the peer modality's density enters the effective potential through $V_{1\mid2}=U_{1\shortto2}/\tau+\log\rho_2$ (and symmetrically for $V_{2\mid1}$), so regions of high peer density act as barriers. Coordinate-wise minimizing sequences can therefore respond by concentrating mass in low-effective-potential regions away from the peer density, leading to barrier-driven structural separation rather than a single Gibbs equilibrium. }
    \label{fig:schematic-geometry}
\end{figure}

\begin{takeaway}{Multimodal InfoNCE geometry}\label{take:multimodal}
Symmetric multimodal InfoNCE is pairwise attractive yet population-level repulsive. While individual cross-modal pairs align, the cross-coupled geometry can make each modality's marginal act as a repulsive \emph{population barrier} against the other. In the analyzed regime, this provides a natural geometric mechanism by which a modality gap can persist (see \Cref{fig:schematic-geometry}).
\end{takeaway}

\section{Experiments}
\label{sec:empirical}

We test the theory at two levels. First, controlled numerical experiments isolate the core mechanisms predicted by the analysis: large-batch gradient consistency, unimodal Gibbs-type concentration, and structural modality gap under controlled cross-modal misalignment. Second, on MS-COCO \citep{lin2014microsoft}, we test whether the same signatures appear in pretrained CLIP-like models, and whether weakening image--text compatibility systematically enlarges the representational modality gap.

\begin{figure*}[t]
\centering

\begin{subfigure}[t]{0.235\linewidth}
    \centering
    \includegraphics[width=\linewidth]{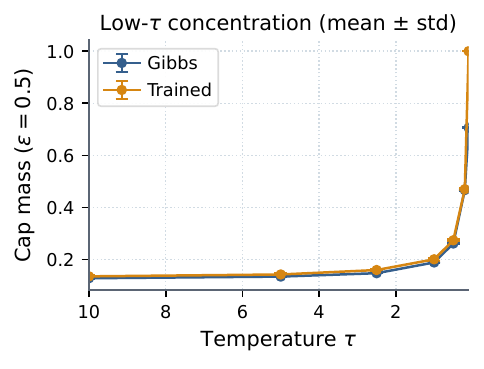}
    \caption{Gibbs concentration.}
    \label{fig:ground_state}
\end{subfigure}
\hfill
\begin{subfigure}[t]{0.246\linewidth}
    \centering
    \includegraphics[width=\linewidth]{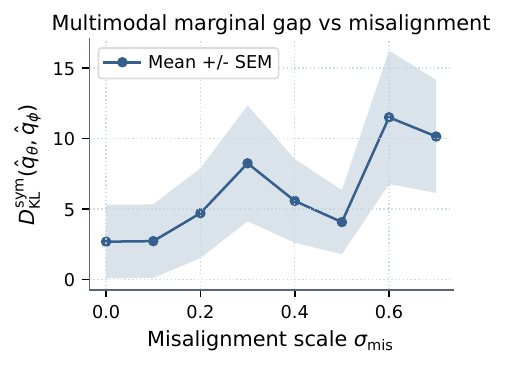}
    \caption{Cross-modal gap.}
    \label{fig:divergence_persist_main}
\end{subfigure}
\hfill
\begin{subfigure}[t]{0.235\linewidth}
    \centering
    \includegraphics[width=\linewidth]{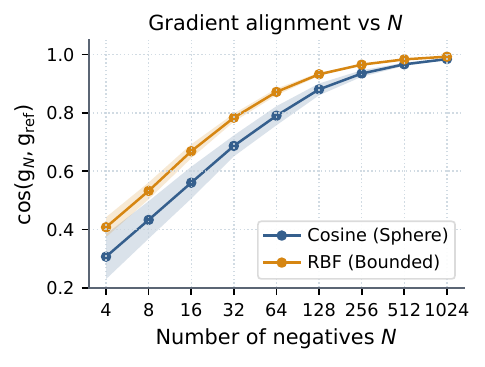}
    \caption{Gradient alignment.}
    \label{fig:grad_alignment}
\end{subfigure}
\hfill
\begin{subfigure}[t]{0.235\linewidth}
    \centering
    \includegraphics[width=\linewidth]{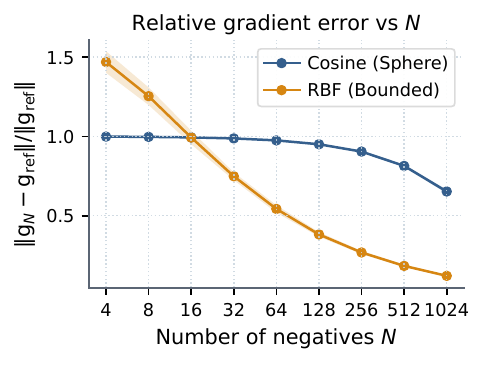}
    \caption{Gradient error.}
    \label{fig:grad_relerr}
\end{subfigure}

\caption{
Numerical evaluation of population predictions and large-batch approximation.
(a) In the unimodal case, the Gibbs equilibrium concentrates increasingly on low-potential regions as \(\tau\downarrow0\).
(b) In the multimodal case, the two representation marginals remain separated, with the gap increasing under stronger cross-modal incompatibility.
(c)--(d) Finite-negative InfoNCE gradients approach a large-batch reference as the number of negatives \(N\) increases: gradient alignment improves while relative gradient error decreases; results for the cosine (spherical regime) and RBF (compact Euclidean regime) similarity functions, showing mean $\pm$ standard deviation over 20 seeds.
}
\label{fig:nce_numerics}
\end{figure*}

\begin{figure*}[ht]
    \centering
    \includegraphics[width=\linewidth]{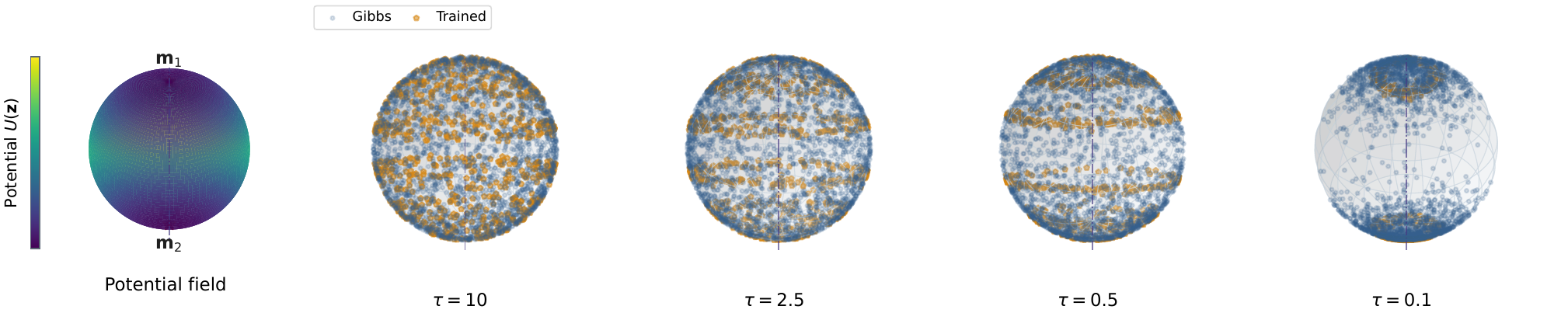}
    \caption{Unimodal potential landscape on $\mathbb S^2$ and distributions across $\tau$. Left: the two-well potential $U$, with minima centers $m_1$ and $m_2$ marked. Right: Gibbs samples (blue) and trained particles (orange) for several temperatures $\tau$. At high temperature, both distributions remain diffuse due to the stronger role of entropy; as $\tau$ decreases, both concentrate around the low-energy wells.}
    \label{fig:uni_s2_overlays_main}
\end{figure*}

\begin{figure*}[ht]
    \centering
    \includegraphics[width=0.245\linewidth]{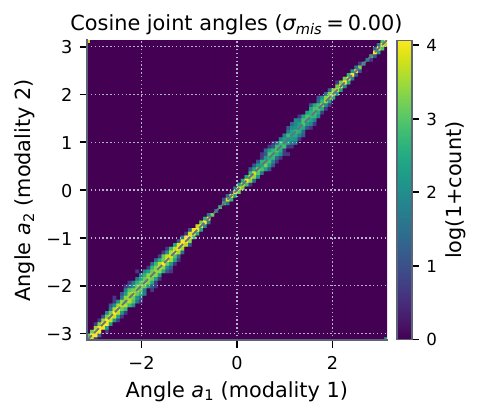}
    \hfill
    \includegraphics[width=0.245\linewidth]{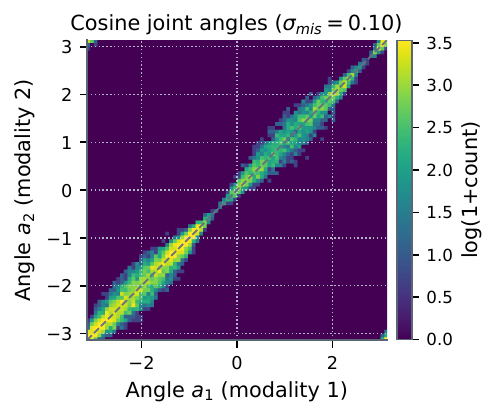}
    \hfill
    \includegraphics[width=0.245\linewidth]{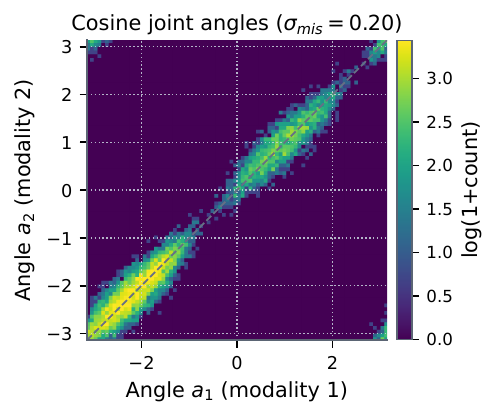}
    \hfill
    \includegraphics[width=0.245\linewidth]{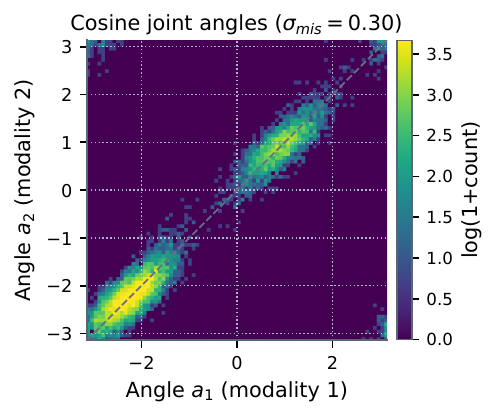}
    \caption{Joint-angle coupling under controlled misalignment. Each plot shows a histogram of joint angles $(a_1,a_2)$ for $\sigma_{\mathrm{mis}} \in \{0,0.1,0.2,0.3\}$. When $\sigma_{\mathrm{mis}}=0$, the coupling is sharply concentrated near the diagonal, indicating near-deterministic pairwise alignment. As $\sigma_{\mathrm{mis}} \uparrow$, the band broadens and deforms, revealing a noisier cross-modal coupling while preserving the coarse latent modes.}
    \label{fig:mm_joint_grid_main}
\end{figure*}
\begin{figure*}[tbp]
    \centering
    \begin{subfigure}{0.46\linewidth}
        \centering
        \includegraphics[width=0.89\linewidth]{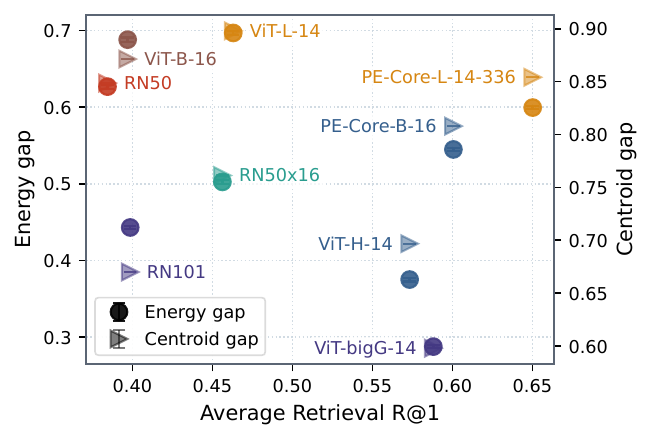}
        \caption{Observational study on OpenCLIP models.}
        \label{subfig:coco_pretrained}
    \end{subfigure}
    \hfill
    \begin{subfigure}{0.46\linewidth}
        \centering
        \includegraphics[width=0.89\linewidth]{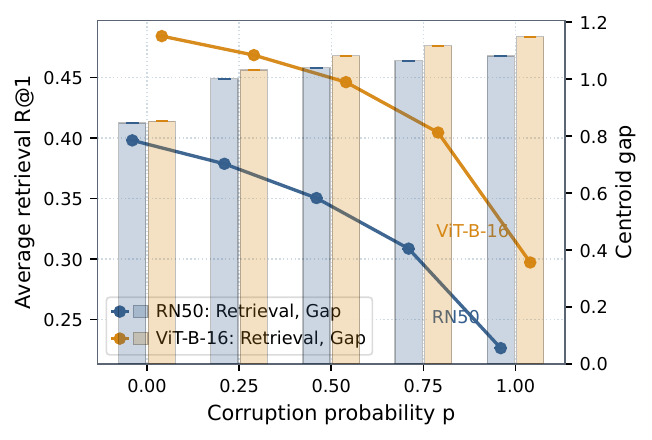}
        \caption{Interventional study with caption corruption.}
        \label{subfig:coco_corruption}
    \end{subfigure}
    \caption{MS-COCO evaluation of the modality-gap mechanism. (\textbf{a}) On frozen OpenCLIP checkpoints, retrieval quality and cross-modal geometric agreement are related but not equivalent. (\textbf{b}) Controlled same-category corruption: Even a semantically plausible weakening of instance-level correspondence systematically enlarges the modality gap while degrading retrieval.}
    \label{fig:coco}
\end{figure*}

\paragraph{Large-batch gradient consistency.}
We first test the large-batch consistency predicted by \Cref{thm:nce_consistency}. On a synthetic Gaussian-mixture dataset, we compare finite-batch InfoNCE gradients with a high-fidelity large-batch reference under both a spherical cosine regime and a compact Euclidean RBF regime. We vary the number of negatives $N$ and report two diagnostics: gradient alignment with the reference and relative gradient error. As shown in \Cref{fig:grad_alignment,fig:grad_relerr}, agreement with the large-batch reference improves steadily as $N$ increases in both regimes: alignment rises, while relative error falls toward zero. This matches the consistency prediction of \Cref{thm:nce_consistency} and suggests that the large-batch gradient approximation can be accurate at moderate $N$ in these controlled settings. Experimental details are in \Cref{app:grad_consist_exp}.

\paragraph{Unimodal equilibrium and low-$\tau$ concentration.}
We next test the intrinsic unimodal picture from \Cref{sec:unimodal}. On a smooth two-well potential on $\mathbb S^2$, we optimize a particle-based surrogate of $\mathcal F_{\tau,U}$ and compare the resulting particle distribution with the Gibbs law $\rho_\tau^* \propto \exp(-U/\tau)$ across temperatures. As shown in \Cref{fig:uni_s2_overlays_main}, the trained particles closely follow the Gibbs trend: at high temperature, both remain diffuse, while at low temperature, both concentrate near the wells. This visual agreement is quantitatively summarized by the cap-mass trend in \Cref{fig:ground_state}, which increases monotonically as $\tau$ decreases for both the Gibbs baseline and the trained particle system. Together, these results support the Gibbs-equilibrium and low-temperature concentration picture of the intrinsic theory. Further experimental details are deferred to \Cref{app:uni_s2_exp}.

\paragraph{Structural modality gap under controlled misalignment.}
We next test the central multimodal prediction from \Cref{sec:multimodal} by examining how controlled conditional heterogeneity affects marginal coincidence. We construct a latent-angle experiment in which both modalities share the same underlying semantic variable, while one modality is perturbed by controlled angular misalignment noise $\sigma_{\mathrm{mis}}$. We train symmetric CLIP-style linear encoders on $\mathbb S^1$ and examine both the learned joint coupling and the induced marginal discrepancy. As shown in \Cref{fig:mm_joint_grid_main}, increasing $\sigma_{\mathrm{mis}}$ progressively broadens and deforms the diagonal coupling structure, indicating weaker cross-modal alignment despite preserved coarse latent structure. This is accompanied by the increasing marginal discrepancy reported in \Cref{fig:divergence_persist_main}. Together, these results are consistent with the theoretical picture that cross-modal heterogeneity can sustain a persistent structural modality gap. Further details are in \Cref{app:mm_gap_exp}.

\paragraph{MS-COCO studies with pretrained OpenCLIP models.}
We finally test whether the multimodal mechanism appears in realistic image--text representations on MS-COCO. In an observational study, we evaluate frozen OpenCLIP checkpoints on \texttt{val2017} using average Recall@1 together with two geometric diagnostics, the centroid gap $\|\mu_I-\mu_T\|$ and the energy distance between image and text marginals. As shown in \Cref{subfig:coco_pretrained}, stronger retrieval does not necessarily imply a smaller modality gap: models with similar AvgR@1 can exhibit substantially different centroid and energy gaps. We then perform a controlled intervention on \texttt{train2017} by replacing each caption, with probability $p$, by a caption from another image sharing at least one COCO category, thereby preserving coarse semantic relatedness while weakening instance-level correspondence. In \Cref{subfig:coco_corruption}, increasing $p$ consistently degrades retrieval and enlarges the centroid gap for both RN50 and ViT-B-16. Together, these results support the prediction that modality gap is not reducible to retrieval quality alone, and that weakening conditional compatibility systematically pushes the image and text marginals further apart. Further details are deferred to \Cref{app:coco_exp}.

\section{Related Work}
\label{app:related_work}

\paragraph{Contrastive objectives and population geometry.} InfoNCE is an instance of noise-contrastive density-ratio estimation, contrasting positives with negatives from a background marginal through softmax normalization \citep{gutmann2010nce,mnih2012fast,oord2018representation}. It is often used as a scalable surrogate for mutual-information criteria \citep{poole2019variational,tschannen2020mutual}, though recent work argues it is better viewed as learning structured density ratios than as an accurate MI estimator \citep{ryu2026contrastive}. A related lens studies population landscapes induced by contrastive objectives, notably through alignment--uniformity on hyperspherical embeddings \citep{wang2020understanding,wang2021understanding,chen2021intriguing,betser2026infonce}; other analyses show such objectives can admit multiple, sometimes trivial, minimizers, with optimization implicitly selecting structured solutions \citep{calder2025understanding}. We instead prove large-batch gradient consistency and derive deterministic energy landscapes for representations on compact manifolds.

\paragraph{Training dynamics and learning theory.}
A distinct line of theory studies contrastive learning under stylized data or generative assumptions. Some analyses track feature or weight dynamics under imperfections such as misalignment, imbalance, or augmentation-induced variation, explaining suppression, robustness, or the effects of filtering and pruning \citep{sun2025contrastive,liao2026theoretical}. Others develop learning-theoretic guarantees under latent-class or multi-view models, and analyze how negative sampling and objective variants affect transfer \citep{saunshi2019theoretical,saunshi2022understanding,chuang2020debiased,tian2020cmc,haochen2023a,ji2023power}; recent CLIP-style analyses further characterize optimal similarity structure and motivate refined objectives \citep{uesaka2025weighted,yoshida2025theoretical}. We do not track weight dynamics under a specific latent model, but characterize InfoNCE's population geometry via large-batch energy limits and variational functionals.

\paragraph{Multimodal alignment and modality gap.}
Empirical work shows that CLIP-style agreement can coexist with a persistent modality gap \citep{liang2022mind,shi2023towards,levi2025the,betser2025whitened}, and recent methods seek to control it via adapters, alternative losses, or explicit regularizers. \citet{yi2025decipher} show that unconstrained or cone-constrained optima close the gap, while dimension collapse into distinct hyperplanes leaves a gap equal to their smallest angle; \citet{yin2026distributional} augment InfoNCE with a kernel-estimated Cauchy--Schwarz divergence to align modality distributions, including unpaired settings; and pairwise sigmoid objectives likewise indicate that cross-modal coupling is sensitive to objective design \citep{zhai2023sigmoid}. We provide a geometric mechanism: symmetric InfoNCE induces cross-modal coupling where each modality's density shapes the other's effective field via a persistent negative symmetric divergence term, providing an objective-level explanation of a modality gap.

\paragraph{Identifiability in contrastive learning.}
A complementary line of work asks when contrastive objectives recover latent generative factors under data-generating assumptions. In nonlinear ICA, auxiliary variables such as temporal nonstationarity enable source recovery up to component-wise transformations \citep{hyvarinen2016unsupervised}. For invertible-mixing models, global optima can invert the generator up to linear ambiguities, often orthogonal transformations on hyperspherical latents \citep{zimmermann2021contrastive,liu2026beyond}. Multi-view models separating invariant content from view-varying style admit recovery of the invariant block up to an invertible mapping \citep{von2021self,Cai2024CLAP}; in multimodal settings with distinct generative mechanisms, contrastive learning further block-identifies shared factors under nontrivial dependencies \citep{daunhawer2023identifiability}. Broader causal analyses with partial observability and data-misalignment formulations also establish identifiability of shared semantics up to smooth bijections \citep{yao2024multiview,cai2025value}. We analyze objective-induced population geometry that standard identifiability analyses do not explain.

\section{Conclusion}
\label{sec:conclusion}
We developed a population-energy view of InfoNCE that exposes a geometric bifurcation in unimodal and symmetric multimodal contrastive learning. In the unimodal regime, intrinsic energy is convex and Gibbs-like, with entropy selecting an internally dispersed equilibrium within the aligned basin. In the multimodal regime, conditional heterogeneity gives rise to a negative symmetric divergence coupling that provides an explicit separation-inducing contribution and can sustain a persistent modality gap. These results show that pairwise alignment alone is not sufficient to control cross-modal marginal structure and motivate objectives that explicitly regulate cross-modal divergence. More broadly, our analysis suggests that multimodal contrastive learning should be understood not only through matched pairs, but through the coupled population geometry of representation marginals. This shift from pointwise matching to population geometry provides a framework for analyzing and designing future representation-learning objectives. See \Cref{app:implications} for further discussion.

\section*{Acknowledgements}
This project was partially supported by the Responsible AI Research (RAIR) Centre, Australia (Z. Zhang, Y. Liu, and J. Q. Shi). We thank the anonymous reviewers for their constructive feedback. 

\section*{Impact Statement}

This paper presents work whose goal is to advance the field of Machine Learning. While there are broad societal consequences associated with the general advancement of AI, our specific theoretical contributions do not present immediate societal impacts that require dedicated discussion.

\bibliography{references}
\bibliographystyle{icml2026}

\newpage
\appendix
\onecolumn  

\crefalias{section}{appendix}      
\crefalias{subsection}{subappendix}   
\crefalias{subsubsection}{subsubappendix} 

% \startcontents[appendix]

% \noindent\textbf{\large Contents of Appendix}
% \vspace{0.5em}
% \hrule
% \begingroup
% \normalsize
% \renewcommand{\addvspace}[1]{\vspace{1em}}
% \setlength{\baselineskip}{1.25em} 
% \printcontents[appendix]{l}{1}{\setcounter{tocdepth}{2}}
% \endgroup

% \vspace{1em}    
% \hrule

\begin{table}[t]
\caption{Core notations used throughout the paper.}
\label{tab:notation_summary}
\centering
\small
{
\setlength{\extrarowheight}{2pt}
\setlength{\tabcolsep}{5pt}
\begin{tabularx}{\textwidth}{lX}
\toprule
\multicolumn{2}{l}{\textbf{Spaces, Measures, Encoders and Laws}} \\
\midrule
\(\mathcal{X}, \mathcal{Y}\) & Observation spaces; in the unimodal case both views lie in \(\mathcal{X}\). \\
\(\mathcal{Z}\subseteq\mathbb{R}^n\) & Common representation space, equipped with a non-atomic Riemannian volume measure $\mu$. \\
\(\mathcal{P}(\mathcal{S})\) & Set of probability measures on a measurable space \(\mathcal{S}\). \\
\(\mathcal{D}_\mu(\mathcal{Z})\) & Set of probability densities on \(\mathcal{Z}\) w.r.t. \(\mu\). \\
\(r_{\mathrm{um}}\in\mathcal{P}(\mathcal{X}\times\mathcal{X})\) & Unimodal positive-pair law governing \((\mathbf{x},\tilde{\mathbf{x}})\). \\
\(r_{\mathrm{mm}}\in\mathcal{P}(\mathcal{X}\times\mathcal{Y})\) & Multimodal positive-pair law governing \((\mathbf{x},\mathbf{y})\). \\
\(p_\mathbf{x}, p_\mathbf{y}\) & Marginal data laws on \(\mathcal{X}\) and \(\mathcal{Y}\), respectively. \\
\(\Theta, \Phi\) & Compact parameter spaces of the encoder families. \\
\(f_\theta:\mathcal{X}\to\mathcal{Z}\) & Encoder for the first modality. \\
\(g_\phi:\mathcal{Y}\to\mathcal{Z}\) & Encoder for the second modality. \\
\(q_\theta:=(f_\theta)_\# p_\mathbf{x}\) & Encoded marginal law induced by \(f_\theta\). \\
\(q_\phi:=(g_\phi)_\# p_\mathbf{y}\) & Encoded marginal law induced by \(g_\phi\). \\
\(\pi_{\theta\theta}:=(f_\theta\times f_\theta)_\# r_{\mathrm{um}}\) & Encoded unimodal positive-pair law. \\
\(\pi_{\theta\phi}:=(f_\theta\times g_\phi)_\# r_{\mathrm{mm}}\) & Encoded multimodal positive-pair law. \\
\midrule
\multicolumn{2}{l}{\textbf{InfoNCE Objectives and Kernel Quantities}} \\
\midrule
\(N \in \mathbb{N}_+\) & Number of negatives in the InfoNCE denominator. \\
\(\mathrm{sim}(z,w)\) & Symmetric similarity function on \(\mathcal Z\times\mathcal Z\).\\
\(\kappa_\tau(z,w)=\exp\!\big(\mathrm{sim}(z,w)/\tau\big)\) & Exponential similarity kernel at temperature $\tau >0$. \\
\(V_\kappa(\tau)\) & Kernel volume constant under \Cref{ass:kernel_volume}. \\
\(\mathcal{L}_{\mathrm{NCE},N}(\theta)\) & Expected finite-$N$ unimodal InfoNCE objective.\\
\(\mathcal{L}_{\mathrm{Sym},N}(\theta,\phi)\) & Expected finite-$N$ symmetric multimodal InfoNCE objective.\\
\(\mathcal{L}_{\mathrm{NCE},N}^{x\shortto y}(\theta,\phi), \mathcal{L}_{\mathrm{NCE},N}^{y\shortto x}(\phi,\theta)\) & Expected finite-$N$ directional multimodal InfoNCE objectives for \(\mathbf{x}\shortto\mathbf{y}\) and \(\mathbf{y}\shortto\mathbf{x}\).\\
\(\Gamma_{\theta,\tau}, \Gamma_{\phi,\tau}\) & Population partition fields at $\tau$, induced by \(q_\theta\) and \(q_\phi\), respectively. \\
\(\tilde{\rho}_{\theta,\tau}, \tilde{\rho}_{\phi,\tau}\) & Kernel-smoothed densities at $\tau$, induced by \(q_\theta\) and \(q_\phi\), respectively. \\
\midrule
\multicolumn{2}{l}{\textbf{Unimodal Geometry}} \\
\midrule
\(\nu_{\theta,z}\) & Conditional law of the positive partner given encoded query $z$ under \(\pi_{\theta\theta}\).\\
\(U_\theta:\mathcal Z\to\mathbb R\) & Unimodal alignment potential field; equivalently, $U_\theta(z)$ is the negative conditional expected similarity at $z$.\\
\(H_\times(q,\tilde{\rho})\) & Cross-entropy of a probability measure \(q\) against a positive density \(\tilde{\rho}\). \\
\(H(\rho)\) & Differential entropy of density \(\rho\). \\
\(\mathcal{J}_\tau(\theta)\) & Unimodal parametric energy arising in the pointwise large-batch limit. \\
\(\mathcal{F}_{\tau,U}(\rho)\) & Intrinsic unimodal free-energy functional over densities \(\rho\). \\
\midrule
\multicolumn{2}{l}{\textbf{Multimodal Geometry}} \\
\midrule
\(\nu_{\theta\phi,z}\) & Conditional law of the $\phi$-encoded partner given $z$ under \(\pi_{\theta\phi}\).\\
\(\nu_{\phi\theta,w}\) & Conditional law of the $\theta$-encoded partner given $w$ under \(\pi_{\theta\phi}\).\\
\(U_{\theta\shortto\phi}(z),\, U_{\phi\shortto\theta}(w)\) & Directional alignment potentials seen by $\theta$ against $\phi$, and conversely. \\
\(\mathcal{J}_\tau^{x\shortto y}(\theta,\phi)\) & Directional multimodal parametric energy for \(x\to y\). \\
\(\mathcal{J}_\tau^{y\shortto x}(\phi,\theta)\) & Directional multimodal parametric energy for $y\shortto x$.\\
\(\mathcal{J}_\tau^{\mathrm{Sym}}(\theta,\phi)\) & Symmetric multimodal parametric energy. \\
\(\mathbf{U}_{1,2}:=(U_{1\shortto2},U_{2\shortto1})\) & Pair of directional potentials in the intrinsic multimodal functional. \\
\(\mathcal{F}_{\tau,\mathbf{U}_{1,2}}^{\mathrm{Sym}}(\rho_1,\rho_2)\) & Intrinsic symmetric multimodal energy. \\
\(D_{\mathrm{KL}}(\rho_1\|\rho_2)\) & Kullback--Leibler divergence. \\
\(D_{\mathrm{KL}}^{\mathrm{Sym}}(\rho_1,\rho_2)\) & Symmetric KL divergence, $\tfrac{1}{2}(D_{\mathrm{KL}}(\rho_1\|\rho_2) + D_{\mathrm{KL}}(\rho_2\|\rho_1))$. \\
\(\underline{\rho}_i>0\) & Density floor used to define the feasible class \(\mathcal D_{\mu,\underline\rho_i}(\mathcal Z)\). \\
\(\mathcal{D}_{\mu,\underline{\rho}}(\mathcal{Z})\) & Floored density class $\{\rho\in\mathcal D_\mu(\mathcal Z)\cap L^\infty(\mu): \rho\ge\underline\rho\ \mu\text{-a.e.}\}$. \\
\(V_{1\mid2}(z)\) & Effective barrier potential for modality 1 given modality 2, $\tfrac{1}{\tau}U_{1\shortto2}(z)+\log\rho_2(z)$. \\
\(V_{2\mid1}(z)\) & Effective barrier potential for modality 2 given modality 1, $\tfrac{1}{\tau}U_{2\shortto1}(z)+\log\rho_1(z)$. \\
\bottomrule
\end{tabularx}
}
\end{table}

\clearpage
\section{Interpretation of Assumptions}
\label{app:justification}

\subsection{On the Optimization Regularity Conditions}
\label{app:critic_justification}

We argue that the regularity conditions in \Cref{assump:regularity} are compatible with standard practice in representation learning. We instantiate these conditions in two dominant geometric regimes: the spherical regime used in modern contrastive learning (e.g., SimCLR \citep{chen2020simple}, CLIP \citep{radford2021learning}) and the Euclidean regime common in classical kernel methods. Both admit the generalized exponential form $\kappa_\tau(z,w):=\exp\!\big(\mathrm{sim}(z,w)/\tau\big)$.

\begin{itemize}[leftmargin=*]
\item \emph{Spherical regime (von Mises--Fisher kernel).}
Representations are constrained to the hypersphere $\mathbb{S}^{n-1}$ via $\ell_2$ normalization, and the similarity function is the cosine similarity $\mathrm{sim}(z,w)=\langle z,w\rangle$:
\begin{equation}
\kappa_\tau^{(\mathrm{vmf})}(z,w)
:=
\exp\!\left(\frac{\langle z,w\rangle}{\tau}\right),
\qquad
z,w\in\mathbb S^{n-1},
\end{equation}

\item \emph{Euclidean regime (RBF kernel).}
The representations lie in a compact subset of $\mathbb{R}^n$, and the similarity function is the negative squared Euclidean distance $\mathrm{sim}(z,w)=-\|z-w\|^2$:
\begin{equation}
\kappa_\tau^{(\mathrm{rbf})}(z,w)
:=
\exp\!\left(-\frac{\|z-w\|^2}{\tau}\right).
\end{equation}
On a generic compact subset of $\mathbb{R}^n$, this example satisfies the boundedness and smoothness requirements used in \Cref{assump:regularity}, but it does not, in general, satisfy the constant kernel-volume condition in \Cref{ass:kernel_volume} because boundary effects make $V_\tau(z):=\int_{\mathcal Z}\kappa_\tau(z,w)\,\mu(\mathrm dw)$ depend on $z$. The condition holds exactly on homogeneous boundaryless domains, such as a flat torus with periodic distance/measure, or after an appropriate boundary-corrected normalization. If \Cref{ass:kernel_volume} is relaxed, the large-batch denominator still converges to the population partition field $\Gamma_{q,\tau}$; however, the clean identity $\Gamma_{q,\tau}=V_\kappa(\tau)\tilde\rho_{q,\tau}$ is replaced by a position-dependent normalization. This introduces additional boundary-volume correction terms, e.g., an effective potential contribution $\tau\log V_\tau(z)$, but does not change the large-batch consistency mechanism or the central population-field interpretation. 
\end{itemize}

\paragraph{Encoder regularity.}
Condition \Cref{assump:regularity}\ref{itm:enc_reg} requires that for each $\mathbf{x}$, the map $\theta\mapsto f_\theta(\mathbf{x})$ is $C^1$ and that its parameter Jacobian is uniformly bounded: $\sup_{\theta\in\Theta}\sup_{\mathbf x\in\mathcal X}\|J_\theta f_\theta(\mathbf x)\|_{\mathrm{op}}<\infty$. For standard neural encoders built from differentiable primitives (e.g., linear maps, convolutions, normalization layers) and smooth activations (e.g., GELU \citep{hendrycks2016gaussian}, SiLU \citep{elfwing2018sigmoid}), the map $\theta\mapsto f_\theta(\mathbf{x})$ is $C^1$ (or at least piecewise $C^1$, which suffices for almost-everywhere analysis). In practice, uniform Jacobian control is promoted by explicit Lipschitz constraints (e.g., spectral normalization, which bounds layer operator norms and hence the network Jacobian via composition \citep{miyato2018spectral}) together with standard norm-regularizing training choices such as weight decay \citep{loshchilov2018decoupled,zhang2018three}. Accordingly, we interpret compactness of $\Theta$ as a mild \emph{localization} assumption: although the ambient parameter space is $\mathbb{R}^p$, successful training typically keeps iterates within a bounded region (often formalized as ``SGD iterates belong to a ball'' in stability analyses \citep{lei2020fine}). One may therefore restrict the analysis to a sufficiently large compact subset $\Theta_R$ containing the training trajectory, or equivalently consider projected updates onto $\Theta_R$ as standard in constrained optimization \citep{boyd2004convex}.

\paragraph{Similarity regularity.}
Condition \Cref{assump:regularity}\ref{itm:crit_reg} requires that $\mathrm{sim}$ is $C^1$ on an open neighborhood of $\mathcal Z\times\mathcal Z$ and that
\(
\sup_{(z,w)\in\mathcal Z\times\mathcal Z}
\|\nabla\mathrm{sim}(z,w)\|<\infty.
\)
In the spherical regime, $\mathrm{sim}(z,w)=\langle z,w\rangle$ is $C^\infty$ on $\mathbb R^n\times\mathbb R^n$, with $\|\nabla_z\mathrm{sim}(z,w)\|=\|w\|$ and $\|\nabla_w\mathrm{sim}(z,w)\|=\|z\|$. Since $\|z\|=\|w\|=1$, these are uniformly bounded by $1$. In the Euclidean regime, $\mathrm{sim}(z,w)=-\|z-w\|^2$ is also $C^\infty$ with gradients scaling linearly with distance. Thus, provided $\mathcal{Z}$ is compact (e.g., via bounded activations like $\tanh$ or explicit clipping), $\|\nabla\mathrm{sim}\|$ remains uniformly bounded on $\mathcal{Z}\times\mathcal{Z}$.

\paragraph{Boundedness of the kernel.}
For any fixed $\tau>0$, the exponential kernel $\kappa_\tau(z,w)=\exp({\mathrm{sim}(z,w)}/{\tau})$ is strictly positive and bounded. For the spherical regime, $\langle z,w\rangle\in[-1,1]$ implies $\kappa_\tau^{(\mathrm{vmf})}(z,w)\in[\exp(-1/\tau),\exp(1/\tau)]$. For the RBF regime on a compact domain with diameter $D = \text{diam}(\mathcal{Z})$, we have $\kappa_\tau^{(\mathrm{rbf})}(z,w)
\in[\exp(-D^2/\tau),1]$. Crucially, in both cases, the kernel is bounded away from zero for each fixed $\tau>0$, ensuring that the logarithmic loss terms are well-defined and Lipschitz continuous.

\subsection{Interpretation of the Sharp Diagonal Peak Regime}
\label{app:sharp_peak_interpretation}

\paragraph{Geometric Interpretation.} \Cref{ass:sharp_peak} is a uniform non-degeneracy condition on the similarity function $\mathrm{sim}(z,w)$ near the diagonal $w=z$. The inequalities in \Cref{eq:sharp_peak_quad} state that for each anchor $z$, the function $w\mapsto\mathrm{sim}(z,w)$ has an isolated maximizer at $w=z$ and admits a locally quadratic upper and lower envelope in geodesic distance. For the exponential kernel $\kappa_\tau(z,w) = \exp(\mathrm{sim}(z,w)/\tau)$, this condition implies that the normalized kernel
\[
\bar\kappa_\tau(z,w)
:=
\frac{\kappa_\tau(z,w)}
{\int_{\mathcal Z}\kappa_\tau(z,u)\,\mu(\mathrm du)}
\]
places asymptotically most of its mass within an $O(\sqrt{\tau})$ geodesic neighborhood of $z$ as $\tau\downarrow0$. Intuitively, $\bar{\kappa}_\tau(z,\cdot)$ behaves like a Gaussian bump of bandwidth $\sqrt{\tau}$ centered at $z$. As a result, the partition field $\Gamma_{\theta,\tau}(z)=\int_{\mathcal Z}\kappa_\tau(z,w)\,q_\theta(\mathrm dw)$ probes only a local neighborhood of $z$ when $\tau$ is small, and the smoothed density $\tilde \rho_{\theta,\tau} = \Gamma_{\theta,\tau}/V_\kappa(\tau)$ becomes a controlled local smoothing of $\rho_\theta$.

\paragraph{Examples.}
For the RBF similarity $\mathrm{sim}(z,w)=-\|z-w\|^2$, \Cref{eq:sharp_peak_quad} holds with equality, so the sharp-diagonal condition in \Cref{ass:sharp_peak} is satisfied locally. This statement concerns the local quadratic peak condition only. The separate constant-volume condition in \Cref{ass:kernel_volume} holds exactly for the RBF kernel on homogeneous boundaryless geometries, such as a flat torus with periodic distance/measure, but not on a generic compact Euclidean subset without boundary correction. On the hypersphere with cosine similarity, $1-\langle z,w\rangle \asymp d_{\mathbb{S}^{n-1}}(z,w)^2$ for nearby points, and the vMF kernel is isotropic on a compact homogeneous manifold, so both the local sharp-peak behavior and the constant-volume condition hold exactly.

\subsection{On Density Floors and Positivity of the Effective Laws}
\label{app:density_floor}

Several parts of our analysis involve logarithms of representation densities (e.g., cross-entropies and KL-type terms).  In the unimodal intrinsic problem, this is benign: no pointwise density floor is required, and the Gibbs minimizer is strictly positive whenever the potential is finite. In the \emph{multimodal} intrinsic problem, however, the objective \emph{subtracts} a symmetric KL-type divergence, creating an explicit ``barrier'' mechanism; without positivity constraints, the variational problem becomes ill-posed. This section explains why (i) the \emph{parametric} large-batch objective is automatically interior due to smoothing, and (ii) the intrinsic density floor is a technically convenient proxy for that interior regime.

\paragraph{Smoothed densities.}
In the large-batch expansion, the quantities that appear are the kernel-smoothed $\mu$-densities
$\tilde\rho_{\theta,\tau},\tilde\rho_{\phi,\tau}$ from \Cref{def:smoothed_density}, not the raw pushforwards $q_\theta,q_\phi$.
Because the exponential kernel satisfies $\kappa_\tau(z,w)=\exp\!\big(\mathrm{sim}(z,w)/\tau\big)>0$ pointwise, we have for every $\tau>0$ and every probability measure $q_\theta$,
\[
\tilde\rho_{\theta,\tau}(z)
=
\frac{1}{V_\kappa(\tau)}
\int_{\mathcal Z}
\kappa_\tau(z,w)\,q_\theta(\mathrm dw)>0,
\qquad z\in\mathcal Z,
\]
and analogously for $\tilde\rho_{\phi,\tau}$. 
Moreover, when $\kappa_\tau$ is continuous (as in our standing setting), $\tilde\rho_{\theta,\tau}$ is continuous on compact $\mathcal Z$ and therefore attains a strictly positive minimum
\[
m_{\theta,\tau}
:=
\min_{z\in\mathcal Z}
\tilde\rho_{\theta,\tau}(z)>0,
\]
and similarly $m_{\phi,\tau}>0$. 
Thus all log-terms appearing in the parametric energy are well-defined at finite temperature.
Importantly, $m_{\theta,\tau}$ depends on $(\theta,\tau)$ and we do \emph{not} assume a uniform lower bound over $\Theta$ (or $\Theta\times\Phi$ in the multimodal case).

\paragraph{From raw parametric densities to smoothed densities.} When we compare $\mathcal J_\tau(\theta)$ to intrinsic functionals expressed in terms of an unsmoothed density $\rho_\theta:=\mathrm dq_\theta/\mathrm d\mu$ (when $q_\theta\ll\mu$), the relevant discrepancy is the KDE bias between $\tilde\rho_{\theta,\tau}$ and $\rho_\theta$. Under the hypothesis used in \Cref{thm:intrinsic_unimodal}, namely
\(
\inf_{z\in\mathcal Z}\rho_\theta(z)
\ge \underline\rho_\theta>0,
\)
whenever
\(
\|\tilde\rho_{\theta,\tau}-\rho_\theta\|_\infty
\le \tfrac12\underline\rho_\theta,
\)
we have
\[
\tilde\rho_{\theta,\tau}(z)
\ge \tfrac12\underline\rho_\theta,
\qquad z\in\mathcal Z.
\]
The sharp-peak smoothing lemma (\Cref{lem:sharp_peak_smoothing}) guarantees this condition for all sufficiently small $\tau$ under its stated hypotheses. Thus, in the low-temperature comparison the assumed floor on the raw density transfers to the corresponding smoothed density.

\paragraph{Density floor in multimodal analysis.}
In the multimodal formulation (\Cref{def:mm_free_energy}), we restrict $(\rho_1,\rho_2)$ to
\[
\mathcal D_{\mu,\underline\rho_i}(\mathcal Z)
:=
\Big\{
\rho\in\mathcal D_\mu(\mathcal Z)\cap L^\infty(\mu):
\rho\ge\underline\rho_i\ \mu\text{-a.e.}
\Big\},
\]
for fixed $\underline\rho_1,\underline\rho_2>0$ satisfying $\underline\rho_i\,\mu(\mathcal Z)<1$, $i\in\{1,2\}$.
This floor serves two roles.

\emph{(i) Well-posedness under negative symmetric divergence.}
Because the multimodal functional contains a \emph{negative} symmetric divergence coupling, allowing densities to vanish can make the objective unbounded below.
Indeed, if $\rho_1=0$ on a set $\mathcal{A}$ with $\mu(\mathcal{A})>0$ while $\rho_2>0$ on $\mathcal{A}$, then $D_{\mathrm{KL}}(\rho_2\|\rho_1)=+\infty$ and hence the negative symmetric divergence term drives $\mathcal F^{\mathrm{Sym}}_{\tau,\mathbf U_{1,2}}(\rho_1,\rho_2)=-\infty$. The floor $\rho_i\ge\underline\rho_i$ prevents this degeneracy and makes the coordinate variational problem well-defined. Equivalently, one may view the constraint as a hard version of the standard ``interior'' condition used to keep log-barrier objectives finite.

\emph{(ii) Transparent coordinate geometry.}
With the floor, we can decompose
\[
\rho_1(z)
=
\underline\rho_1+\eta_1(z),
\qquad
\eta_1(z)\ge0,
\qquad
\int_{\mathcal Z}
\eta_1(z)\,\mu(\mathrm dz)
=
M_{\mathrm{ex}}^{(1)}
=
1-\underline\rho_1\,\mu(\mathcal Z),
\]
and analogously for $\rho_2$. This isolates a fixed ``background'' mass from an ``excess'' budget $M^{(1)}_{\mathrm{ex}}$, making the extremal coordinate behavior in \Cref{prop:mm_coordinate_extremal} quantitatively explicit. In this sense, the density floor is not an arbitrary restriction: it is the simplest device that (a) mirrors the intrinsic interiority of the parametrically smoothed quantities at finite $\tau$, and (b) exposes the barrier-driven extremal coordinate mechanism in closed form.

\section{Lemmas}
\label{app:extended_statement}

\subsection{Uniform Gradient Bounds for InfoNCE Objectives}
\label{app:grad_bound}

In this section, we prove uniform gradient bounds for unimodal and multimodal InfoNCE objectives, which are used in the large-batch gradient-consistency proofs of \Cref{thm:nce_consistency,thm:sym_consistency}.

\subsubsection{Unimodal Optimization Stability}

\begin{restatable}[Uniform unimodal gradient bound]{lemma}{propasymgradbound}
\label{lem:asym_grad_bound}
Assume \Cref{assump:regularity}. Fix any temperature $\tau>0$ and any batch size $N\in\mathbb{N}_+$. Then the unimodal InfoNCE objective defined in \Cref{eq:asym_loss_def} has a uniformly bounded gradient over the parameter space $\Theta$. Formally, there exists a constant $C(\tau)<\infty$ such that
\(
\|\nabla_\theta \mathcal{L}_{\mathrm{NCE},N}(\theta)\|\le C(\tau)
\), uniformly for all $\theta\in\Theta$.
\end{restatable}

\begin{proofintuition}
This proof establishes that the InfoNCE gradients cannot explode, providing a uniform upper bound that is strictly independent of the number of negative samples $N$. The intuition rests on two core mechanisms: \textbf{(i)} Because the representation manifold is compact and both the encoder and similarity functions are sufficiently smooth with bounded Jacobians, the chain rule guarantees that the gradient of any single pairwise similarity score is inherently bounded. \textbf{(ii)} The gradient of the log-partition denominator simplifies to a softmax-weighted sum of the pairwise gradients. Because this is a convex combination, the aggregated gradient norm cannot exceed the maximum norm of the individual bounded gradients. Consequently, increasing the number of negatives does not by itself cause the gradient bound to grow with $N$.
\end{proofintuition}

\begin{proof} 
\textbf{Step 1: Bounding components in the loss.}
Since $\mathcal Z$ is compact, the product $\mathcal Z\times\mathcal Z$ is compact. By \Cref{assump:regularity}\ref{itm:crit_reg}, $\mathrm{sim}$ is $C^1$ on an open neighborhood of $\mathcal Z\times\mathcal Z$, so $\nabla\mathrm{sim}$ is continuous on $\mathcal Z\times\mathcal Z$ and hence attains its maximum norm on this compact set. For a fixed $\tau>0$, define the temperature-dependent similarity sensitivity bound
\begin{equation}
\label{eq:sim_grad_bound}
C_{\mathrm{sim}}(\tau)
:=
\sup_{(z,w)\in\mathcal Z\times\mathcal Z}
\left\|
\nabla_{(z,w)}
\left(
\frac{1}{\tau}\mathrm{sim}(z,w)
\right)
\right\|
=
\frac{1}{\tau}
\sup_{(z,w)\in\mathcal Z\times\mathcal Z}
\|\nabla\mathrm{sim}(z,w)\|
<\infty.
\end{equation}
Next, by \Cref{assump:regularity}\ref{itm:enc_reg}, the encoder Jacobian is uniformly bounded:
\begin{equation}
\label{eq:encod_grad_bound}
C_\Theta
:=
\sup_{\theta\in\Theta}\sup_{x\in\mathcal X}
\|J_\theta f_\theta(x)\|_{\mathrm{op}}
<\infty.
\end{equation}
For any $x,x'\in\mathcal X$, write $z=f_\theta(x)$ and $w=f_\theta(x')$, and denote the corresponding pairwise parameter gradient by
\[
\alpha_\theta(z,w)
:=
\nabla_\theta
\left(
\frac{1}{\tau}\mathrm{sim}(z,w)
\right).
\]
By the chain rule,
\[
\alpha_\theta(z,w)
=
\nabla_z
\left(
\frac{1}{\tau}\mathrm{sim}(z,w)
\right)^\top
J_\theta f_\theta(x)
+
\nabla_w
\left(
\frac{1}{\tau}\mathrm{sim}(z,w)
\right)^\top
J_\theta f_\theta(x').
\]
Therefore, by the triangle inequality and sub-multiplicativity,
\begin{equation}
\label{eq:alpha_bound}
\begin{aligned}
\|\alpha_\theta(z,w)\|
&\le
\Big\|
\nabla_z\frac{1}{\tau}\mathrm{sim}(z,w)
\Big\|
\|J_\theta f_\theta(x)\|_{\mathrm{op}}
+
\Big\|
\nabla_w\frac{1}{\tau}\mathrm{sim}(z,w)
\Big\|
\|J_\theta f_\theta(x')\|_{\mathrm{op}}
\\
&\le
2C_{\mathrm{sim}}(\tau)C_\Theta.
\end{aligned}
\end{equation}
Thus, the parameter gradient of any pairwise similarity term is uniformly bounded.

\textbf{Step 2: Gradient of the finite-batch partition function.}
Fix an anchor $z=f_\theta(x)$, a positive $v=f_\theta(\tilde x)$, and negatives $\{w_j\}_{j=1}^N$. Define the contrastive set
\[
\mathcal C(z):=\{v\}\cup\{w_j\}_{j=1}^N.
\]
The gradient of the log-partition function
\[
\log\mathsf Z_{\mathcal B}(z)
=
\log
\sum_{w\in\mathcal C(z)}
\exp\!\left(\frac{\mathrm{sim}(z,w)}{\tau}\right)
\]
is given by the softmax-weighted sum of the inner gradients \citep[see e.g.][p.~74]{boyd2004convex}. Since $\kappa_\tau(z,w)=\exp(\mathrm{sim}(z,w)/\tau)>0$, we may differentiate $\log\mathsf Z_{\mathcal B}$ and write
\begin{equation}
\label{eq:logZ_grad_softmax}
\nabla_\theta\log\mathsf Z_{\mathcal B}(z)
=
\sum_{w\in\mathcal C(z)}
\sigma_{\mathcal B}(w)
\nabla_\theta\log\kappa_\tau(z,w)
=
\sum_{w\in\mathcal C(z)}
\sigma_{\mathcal B}(w)\alpha_\theta(z,w),
\end{equation}
where
\[
\sigma_{\mathcal B}(w)
:=
\frac{\kappa_\tau(z,w)}
{\mathsf Z_{\mathcal B}(z)}
\quad\text{satisfies}\quad
\sigma_{\mathcal B}(w)\ge0,
\qquad
\sum_{w\in\mathcal C(z)}
\sigma_{\mathcal B}(w)=1.
\]
Therefore, using \Cref{eq:alpha_bound},
\begin{equation}
\label{eq:bound_logsum}
\begin{aligned}
\|\nabla_\theta\log\mathsf Z_{\mathcal B}(z)\|
&\le
\sum_{w\in\mathcal C(z)}
\sigma_{\mathcal B}(w)
\|\alpha_\theta(z,w)\|
\\
&\le
2C_{\mathrm{sim}}(\tau)C_\Theta
\sum_{w\in\mathcal C(z)}
\sigma_{\mathcal B}(w)
\\
&=
2C_{\mathrm{sim}}(\tau)C_\Theta.
\end{aligned}
\end{equation}

\textbf{Step 3: Bounding the loss gradient.}
For a fixed batch $\mathcal B$ and anchor $z=f_\theta(x)$, define the per-batch loss
\[
\ell_{\mathcal B}(\theta)
:=
-\log\kappa_\tau(z,v)
+
\log\mathsf Z_{\mathcal B}(z),
\]
where $v=f_\theta(\tilde x)$ is the positive match. Since $\nabla_\theta\log\kappa_\tau(z,v)=\alpha_\theta(z,v)$,
\[
\nabla_\theta\ell_{\mathcal B}(\theta)
=
-\alpha_\theta(z,v)
+
\nabla_\theta\log\mathsf Z_{\mathcal B}(z).
\]
By the triangle inequality and \Cref{eq:alpha_bound,eq:bound_logsum},
\begin{equation}
\label{eq:batch_grad_bound}
\|\nabla_\theta\ell_{\mathcal B}(\theta)\|
\le
\|\alpha_\theta(z,v)\|
+
\|\nabla_\theta\log\mathsf Z_{\mathcal B}(z)\|
\le
4C_{\mathrm{sim}}(\tau)C_\Theta.
\end{equation}
Finally, the expected finite-$N$ objective is
\[
\mathcal L_{\mathrm{NCE},N}(\theta)
=
\mathbb E_{\mathcal B}
[\ell_{\mathcal B}(\theta)].
\]
Since $\ell_{\mathcal B}$ is differentiable in $\theta$ by \Cref{assump:regularity} for each $\mathcal B$ and $\|\nabla_\theta\ell_{\mathcal B}(\theta)\| \le4C_{\mathrm{sim}}(\tau)C_\Theta$ provides an integrable dominating function, we may differentiate under the expectation, obtaining
\[
\nabla_\theta\mathcal L_{\mathrm{NCE},N}(\theta)
=
\mathbb E_{\mathcal B}
\big[
\nabla_\theta\ell_{\mathcal B}(\theta)
\big].
\]
Taking norms and applying Jensen's inequality yields
\[
\|\nabla_\theta\mathcal L_{\mathrm{NCE},N}(\theta)\|
\le
\mathbb E_{\mathcal B}
\big[
\|\nabla_\theta\ell_{\mathcal B}(\theta)\|
\big]
\le
4C_{\mathrm{sim}}(\tau)C_\Theta.
\]
Thus, for each fixed $\tau>0$ and any $N\in\mathbb N_+$, the gradient of the unimodal InfoNCE objective defined in \Cref{eq:asym_loss_def} is uniformly bounded over $\Theta$ by $C(\tau):=4C_{\mathrm{sim}}(\tau)C_\Theta$.
\end{proof}

\subsubsection{Multimodal Optimization Stability}

\begin{lemma}[Uniform multimodal gradient bound]
\label{lem:sym_grad_bound}
Assume \Cref{assump:mm_regularity}. Fix any $\tau>0$ and any $N\in\mathbb N_+$. Then the symmetric multimodal InfoNCE objective $\mathcal L_{\mathrm{Sym},N}(\theta,\phi)$ defined in \Cref{eq:sym_loss_def} has a uniformly bounded joint gradient over $\Theta\times\Phi$. Formally, there exists a constant $C(\tau)<\infty$ such that
\[
\|\nabla_{(\theta,\phi)}
\mathcal L_{\mathrm{Sym},N}(\theta,\phi)\|
\le C(\tau),
\qquad
\forall(\theta,\phi)\in\Theta\times\Phi.
\]
\end{lemma}

\begin{proofintuition}
The proof extends the unimodal bound to $(\Theta,\Phi)$. For either direction, the softmax-weighted denominator is a convex combination of uniformly bounded pairwise gradients, so the partial gradients with respect to both encoder parameters remain uniformly bounded independently of $N$. The joint gradient is obtained by concatenating the two parameter blocks, and averaging the two directions preserves a uniform bound.
\end{proofintuition}

\begin{proof}
\textbf{Step 1: Uniform bounds on components.}
We retain the similarity sensitivity bound $C_{\mathrm{sim}}(\tau)$ from \Cref{eq:sim_grad_bound}. For the encoders, define
\[
C_\Theta
:=
\sup_{\theta\in\Theta}\sup_{x\in\mathcal X}
\|J_\theta f_\theta(x)\|_{\mathrm{op}}
<\infty,
\qquad
C_\Phi
:=
\sup_{\phi\in\Phi}\sup_{y\in\mathcal Y}
\|J_\phi g_\phi(y)\|_{\mathrm{op}}
<\infty,
\]
which are finite by \Cref{assump:mm_regularity}. Define the global encoder constant
\[
C_{\mathrm{enc}}:=\max\{C_\Theta,C_\Phi\}.
\]
Consider a generic cross-modal representation pair $z=f_\theta(x)$ and $w=g_\phi(y)$. The gradient of $\log\kappa_\tau(z,w)$ with respect to the joint parameter vector $(\theta,\phi)$ splits into block components. By the chain rule and \Cref{eq:sim_grad_bound}, the partial gradients satisfy
\begin{equation}
\label{eq:mm_partial_bounds}
\|\nabla_\theta \log \kappa_\tau(z,w)\|
\le
C_{\mathrm{sim}}(\tau)C_\Theta
\le
C_{\mathrm{sim}}(\tau)C_{\mathrm{enc}},
\qquad
\|\nabla_\phi \log \kappa_\tau(z,w)\|
\le
C_{\mathrm{sim}}(\tau)C_\Phi
\le
C_{\mathrm{sim}}(\tau)C_{\mathrm{enc}}.
\end{equation}

\textbf{Step 2: Bounding directional partial derivatives.}
Consider a fixed batch realization $\mathcal B$ consisting of a positive pair $(x,y)$ and $N$ negative keys $\{y_j'\}_{j=1}^N$, with the analogous construction in the reverse direction. The symmetric per-batch loss is the average of the two directional losses:
\[
\ell_{\mathrm{Sym}}(\mathcal B)
=
\tfrac12
\big(
\ell_{\theta\shortto\phi}(\mathcal B)
+
\ell_{\phi\shortto\theta}(\mathcal B)
\big).
\]
By the triangle inequality,
\[
\|\nabla_{(\theta,\phi)}\ell_{\mathrm{Sym}}\|
\le
\tfrac12
\|\nabla_{(\theta,\phi)}\ell_{\theta\shortto\phi}\|
+
\tfrac12
\|\nabla_{(\theta,\phi)}\ell_{\phi\shortto\theta}\|.
\]

We first analyze the forward component $\ell_{\theta\shortto\phi}$. Here, $x$ serves as the anchor, $y$ as the positive key, and $\{y_j'\}_{j=1}^N$ as the negative keys:
\[
\ell_{\theta\shortto\phi}
:=
-\log\kappa_\tau\big(f_\theta(x),g_\phi(y)\big)
+
\log\!\Big(
\kappa_\tau\big(f_\theta(x),g_\phi(y)\big)
+
\sum_{j=1}^N
\kappa_\tau\big(f_\theta(x),g_\phi(y_j')\big)
\Big).
\]
The joint gradient vector is the concatenation
\[
\nabla_{(\theta,\phi)}\ell_{\theta\shortto\phi}
=
\left(
\nabla_\theta\ell_{\theta\shortto\phi}^{\top},
\nabla_\phi\ell_{\theta\shortto\phi}^{\top}
\right)^{\top}.
\]
We bound the two parameter blocks separately. Let
\[
\mathcal C_\phi
:=
\{g_\phi(y)\}
\cup
\{g_\phi(y_j')\}_{j=1}^N,
\]
and denote the softmax weight assigned to any $v\in\mathcal C_\phi$ by $\sigma(v)$.

\emph{(i) Gradient with respect to $\theta$ (anchor sensitivity).}
In this direction, $\theta$ governs only the anchor $f_\theta(x)$; the keys in $\mathcal C_\phi$ are fixed with respect to $\theta$. Thus,
\[
\nabla_\theta\ell_{\theta\shortto\phi}
=
-\nabla_\theta
\log\kappa_\tau\big(f_\theta(x),g_\phi(y)\big)
+
\sum_{v\in\mathcal C_\phi}
\sigma(v)
\nabla_\theta
\log\kappa_\tau\big(f_\theta(x),v\big).
\]
Applying the triangle inequality and the uniform sensitivity bound in \Cref{eq:mm_partial_bounds},
\begin{align*}
\|\nabla_\theta\ell_{\theta\shortto\phi}\|
&\le
\|\nabla_\theta
\log\kappa_\tau\big(f_\theta(x),g_\phi(y)\big)\|
+
\sum_{v\in\mathcal C_\phi}
\sigma(v)
\|\nabla_\theta
\log\kappa_\tau\big(f_\theta(x),v\big)\|
\\
&\le
C_{\mathrm{sim}}(\tau)C_\Theta
+
C_{\mathrm{sim}}(\tau)C_\Theta
\sum_{v\in\mathcal C_\phi}\sigma(v)
\\
&=
2C_{\mathrm{sim}}(\tau)C_\Theta,
\end{align*}
since $\sum_{v\in\mathcal C_\phi}\sigma(v)=1$.

\emph{(ii) Gradient with respect to $\phi$ (key sensitivity).}
In this direction, $\phi$ parametrizes every key in the batch, while the anchor $f_\theta(x)$ is fixed with respect to $\phi$. Hence,
\[
\nabla_\phi\ell_{\theta\shortto\phi}
=
-\nabla_\phi
\log\kappa_\tau\big(f_\theta(x),g_\phi(y)\big)
+
\sum_{v\in\mathcal C_\phi}
\sigma(v)
\nabla_\phi
\log\kappa_\tau\big(f_\theta(x),v\big).
\]
Similarly,
\begin{align*}
\|\nabla_\phi\ell_{\theta\shortto\phi}\|
&\le
C_{\mathrm{sim}}(\tau)C_\Phi
+
C_{\mathrm{sim}}(\tau)C_\Phi
\sum_{v\in\mathcal C_\phi}\sigma(v)
\\
&=
2C_{\mathrm{sim}}(\tau)C_\Phi.
\end{align*}

\textbf{Step 3: Global uniform bound.}
Combining the two parameter blocks gives
\begin{align*}
\|\nabla_{(\theta,\phi)}\ell_{\theta\shortto\phi}\|
&=
\sqrt{
\|\nabla_\theta\ell_{\theta\shortto\phi}\|^2
+
\|\nabla_\phi\ell_{\theta\shortto\phi}\|^2
}
\\
&\le
\sqrt{
\big(2C_{\mathrm{sim}}(\tau)C_\Theta\big)^2
+
\big(2C_{\mathrm{sim}}(\tau)C_\Phi\big)^2
}
\\
&\le
2\sqrt{2}\,
C_{\mathrm{sim}}(\tau)C_{\mathrm{enc}}.
\end{align*}
By symmetry, the same bound applies to the backward loss $\ell_{\phi\shortto\theta}$. Therefore,
\[
\begin{aligned}
\|\nabla_{(\theta,\phi)}\ell_{\mathrm{Sym}}\|
&=
\tfrac12
\left\|
\nabla_{(\theta,\phi)}\ell_{\theta\shortto\phi}
+
\nabla_{(\theta,\phi)}\ell_{\phi\shortto\theta}
\right\|
\\
&\le
\tfrac12
\left(
\|\nabla_{(\theta,\phi)}\ell_{\theta\shortto\phi}\|
+
\|\nabla_{(\theta,\phi)}\ell_{\phi\shortto\theta}\|
\right)
\\
&\le
2\sqrt{2}\,
C_{\mathrm{sim}}(\tau)C_{\mathrm{enc}}.
\end{aligned}
\]

Finally,
\[
\mathcal L_{\mathrm{Sym},N}(\theta,\phi)
=
\mathbb E_{\mathcal B}
\big[\ell_{\mathrm{Sym}}(\mathcal B)\big].
\]
Since the per-batch gradient is uniformly bounded, we may differentiate under the expectation, obtaining
\[
\nabla_{(\theta,\phi)}
\mathcal L_{\mathrm{Sym},N}(\theta,\phi)
=
\mathbb E_{\mathcal B}
\left[
\nabla_{(\theta,\phi)}
\ell_{\mathrm{Sym}}(\mathcal B)
\right].
\]
Hence, by Jensen's inequality,
\[
\begin{aligned}
\|\nabla_{(\theta,\phi)}
\mathcal L_{\mathrm{Sym},N}(\theta,\phi)\|
&\le
\mathbb E_{\mathcal B}
\left[
\|\nabla_{(\theta,\phi)}
\ell_{\mathrm{Sym}}(\mathcal B)\|
\right]
\\
&\le
2\sqrt{2}\,
C_{\mathrm{sim}}(\tau)C_{\mathrm{enc}}.
\end{aligned}
\]
Thus, with
\[
C(\tau)
:=
2\sqrt{2}\,
C_{\mathrm{sim}}(\tau)C_{\mathrm{enc}}
<\infty,
\]
the gradient is uniformly bounded over $\Theta\times\Phi$.
\end{proof}

\subsection{Sharp-peak Smoothing Implies Vanishing KDE Mismatch}
\label{app:sharp_peak_smoothing_lemma}

The large-batch parametric energy $\mathcal J_\tau(\theta)$ depends on the
\emph{kernel-smoothed} density $\tilde\rho_{\theta,\tau}$ through the
cross-entropy term $H_\times(q_\theta,\tilde\rho_{\theta,\tau})$, whereas the
intrinsic functional $\mathcal F_{\tau,U_\theta}$ is expressed in terms of the
\emph{true} density $\rho_\theta$ and its entropy $H(\rho_\theta)$. To compare
$\mathcal J_\tau(\theta)$ with $\mathcal F_{\tau,U_\theta}(\rho_\theta)$ in
\Cref{thm:intrinsic_unimodal}, we therefore require quantitative control of the
bias introduced by smoothing, i.e., that $\tilde\rho_{\theta,\tau}$ approximates
$\rho_\theta$ when the kernel is sufficiently sharp.

\begin{lemma}[Vanishing KDE error under sharp peak]
\label{lem:sharp_peak_smoothing}
Assume \Cref{ass:kernel_volume,assump:regularity,ass:sharp_peak}. Fix
$\theta\in\Theta$ and assume $q_\theta\ll\mu$ with a continuous density
$\rho_\theta$ bounded away from zero, i.e.,
\[
\inf_{z\in\mathcal Z}\rho_\theta(z)
>
\underline\rho_\theta
>
0.
\]
Define the KDE mismatch
\[
\varepsilon_{\mathrm{kde}}^{(\theta)}(\tau)
:=
\|\tilde\rho_{\theta,\tau}-\rho_\theta\|_\infty,
\]
where $\tilde\rho_{\theta,\tau}$ is the kernel-smoothed density from
\Cref{def:smoothed_density}. Then, as $\tau\to0^+$,
\[
\varepsilon_{\mathrm{kde}}^{(\theta)}(\tau)\to0.
\]
\end{lemma}

\begin{proofintuition}
The goal is to show that as $\tau\to0^+$, the kernel-smoothed density
$\tilde\rho_{\theta,\tau}$ converges uniformly to the continuous density
$\rho_\theta$ over $\mathcal Z$. The intuition relies on two properties.
\textbf{(i)} The similarity function has a sharp, unique peak along the
diagonal. As $\tau\to0^+$, the normalized exponential kernel increasingly
concentrates its mass near $w=z$, while contributions from points separated
from $z$ decay exponentially. \textbf{(ii)} Since $\rho_\theta$ is continuous
on compact $\mathcal Z$, it is uniformly continuous. Thus, within the
shrinking neighborhood on which the kernel concentrates, $\rho_\theta(w)$ is
uniformly close to $\rho_\theta(z)$, while the contribution from the
complement vanishes. Together, these properties imply uniform convergence of
the smoothed density to $\rho_\theta$.
\end{proofintuition}

\begin{proof}
Fix any $\theta\in\Theta$. Let $d_\mathcal Z(\cdot,\cdot)$ denote the
geodesic distance on the compact Riemannian manifold $\mathcal Z$ as in
\Cref{ass:sharp_peak}. Since $\mathcal Z$ is compact and $\rho_\theta$ is
continuous, $\rho_\theta$ is bounded. Define the normalized kernel
\[
\bar\kappa_\tau(z,w)
:=
\frac{\kappa_\tau(z,w)}{V_\kappa(\tau)}.
\]
We first show that $\bar\kappa_\tau$ concentrates near the diagonal
$\{z=w\}$ as $\tau\to0^+$, and then establish the claimed sup-norm
convergence.

\textbf{Step 1: Uniform concentration of $\bar\kappa_\tau$ near the diagonal.}
Fix any $\delta\in(0,r)$, where $r$ is from \Cref{ass:sharp_peak}, and define
\[
\mathcal A_\delta
:=
\left\{
(z,w)\in\mathcal Z\times\mathcal Z:
d_\mathcal Z(z,w)\ge\delta
\right\}.
\]
By compactness of $\mathcal Z$, the set $\mathcal A_\delta$ is compact.
Since $\mathrm{sim}$ is continuous by \Cref{assump:regularity}, consider
\[
g(z,w)
:=
\mathrm{sim}(z,z)-\mathrm{sim}(z,w).
\]
By \Cref{ass:sharp_peak}, for each fixed $z\in\mathcal Z$ the map
$w\mapsto\mathrm{sim}(z,w)$ has a unique maximizer at $w=z$; hence
$g(z,w)>0$ whenever $d_\mathcal Z(z,w)\ge\delta$. Therefore, $g$ is strictly
positive on $\mathcal A_\delta$, and by compactness it attains a strictly
positive minimum:
\[
m_\delta
:=
\min_{(z,w)\in\mathcal A_\delta}
g(z,w)
>
0.
\]
Equivalently, for all $z,w\in\mathcal Z$ satisfying
$d_\mathcal Z(z,w)\ge\delta$,
\begin{equation}
\label{eq:uniform_gap}
\mathrm{sim}(z,w)
\le
\mathrm{sim}(z,z)-m_\delta.
\end{equation}

Fix $\eta\in(0,\delta)$. Let
\[
\mathbb B_l(z)
:=
\left\{
w\in\mathcal Z:
d_\mathcal Z(z,w)<l
\right\}
\]
denote an open geodesic ball of radius $l$. By compactness of $\mathcal Z$,
the open cover
$\{\mathbb B_{\eta/2}(z):z\in\mathcal Z\}$ admits a finite subcover. Thus,
there exist points $\{z_1,\dots,z_M\}\subset\mathcal Z$ such that
\[
\mathcal Z
=
\bigcup_{j=1}^M
\mathbb B_{\eta/2}(z_j).
\]
For any $z\in\mathcal Z$, choose
$i\in\{1,\dots,M\}$ such that
$z\in\mathbb B_{\eta/2}(z_i)$. For any
$w\in\mathbb B_{\eta/2}(z_i)$, the triangle inequality gives
\[
d_\mathcal Z(z,w)
\le
d_\mathcal Z(z_i,w)
+
d_\mathcal Z(z_i,z)
<
\eta.
\]
Hence
\[
\mathbb B_{\eta/2}(z_i)
\subseteq
\mathbb B_\eta(z).
\]
By monotonicity of $\mu$,
\[
\mu\big(\mathbb B_\eta(z)\big)
\ge
\mu\big(\mathbb B_{\eta/2}(z_i)\big)
\ge
\min_{1\le j\le M}
\mu\big(\mathbb B_{\eta/2}(z_j)\big)
=:b_\eta>0,
\]
where positivity follows from the full support of the Riemannian volume
measure $\mu$.

Now choose $\eta\in(0,\delta)$ sufficiently small that
\begin{equation}
\label{eq:rho_choice}
m_2\eta^2<m_\delta,
\end{equation}
where $m_2$ is the constant in \Cref{ass:sharp_peak}. For
$w\in\mathbb B_\eta(z)\subset\mathbb B_r(z)$, the lower quadratic bound in
\Cref{ass:sharp_peak} gives
\[
\mathrm{sim}(z,w)
\ge
\mathrm{sim}(z,z)
-
m_2d_\mathcal Z(z,w)^2
\ge
\mathrm{sim}(z,z)-m_2\eta^2.
\]
Therefore,
\[
\kappa_\tau(z,w)
\ge
\exp\!\big(\mathrm{sim}(z,z)/\tau\big)
\exp\!\big(-m_2\eta^2/\tau\big).
\]
Integrating over $\mathbb B_\eta(z)$ and using
$\mu(\mathbb B_\eta(z))\ge b_\eta$ yields
\begin{equation}
\label{eq:Vk_lower}
V_\kappa(\tau)
=
\int_{\mathcal Z}
\kappa_\tau(z,w)\,\mu(\mathrm dw)
\ge
b_\eta
\exp\!\big(\mathrm{sim}(z,z)/\tau\big)
\exp\!\big(-m_2\eta^2/\tau\big).
\end{equation}
Here, by \Cref{ass:kernel_volume}, $V_\kappa(\tau)$ is independent of $z$.
The factor $\exp(\mathrm{sim}(z,z)/\tau)$ cancels in the ratio below.

Next, for $w\notin\mathbb B_\delta(z)$ we have
$d_\mathcal Z(z,w)\ge\delta$, so by \Cref{eq:uniform_gap},
\[
\kappa_\tau(z,w)
=
\exp\!\big(\mathrm{sim}(z,w)/\tau\big)
\le
\exp\!\big(\mathrm{sim}(z,z)/\tau\big)
\exp\!\big(-m_\delta/\tau\big).
\]
Therefore, with $V_\mu:=\mu(\mathcal Z)<\infty$,
\begin{equation}
\label{eq:tail_num}
\int_{\mathcal Z\setminus\mathbb B_\delta(z)}
\kappa_\tau(z,w)\,\mu(\mathrm dw)
\le
V_\mu
\exp\!\big(\mathrm{sim}(z,z)/\tau\big)
\exp\!\big(-m_\delta/\tau\big).
\end{equation}
Dividing \Cref{eq:tail_num} by \Cref{eq:Vk_lower} and using
\Cref{eq:rho_choice} gives, uniformly in $z$,
\begin{equation}
\label{eq:tail_mass}
\int_{\mathcal Z\setminus\mathbb B_\delta(z)}
\bar\kappa_\tau(z,w)\,\mu(\mathrm dw)
\le
\frac{V_\mu}{b_\eta}
\exp\!\left(
-\frac{m_\delta-m_2\eta^2}{\tau}
\right)
\xrightarrow{\tau\to0^+}
0.
\end{equation}

\textbf{Step 2: Sup-norm convergence of the smoothing.}
Fix $\varepsilon>0$. Since $\rho_\theta$ is continuous on compact
$\mathcal Z$, it is uniformly continuous. Hence, for fixed
$\theta\in\Theta$, choose $\delta\in(0,r)$ such that
\[
|\rho_\theta(w)-\rho_\theta(z)|
\le
\varepsilon
\qquad
\text{whenever }
d_\mathcal Z(z,w)<\delta.
\]
For any $z\in\mathcal Z$,
\[
\tilde\rho_{\theta,\tau}(z)-\rho_\theta(z)
=
\int_{\mathcal Z}
\bar\kappa_\tau(z,w)
\big(
\rho_\theta(w)-\rho_\theta(z)
\big)
\,\mu(\mathrm dw),
\]
and we split the integral over $\mathbb B_\delta(z)$ and its complement.
Using
\[
\int_{\mathcal Z}
\bar\kappa_\tau(z,w)\,\mu(\mathrm dw)
=
1,
\]
the bound on $\mathbb B_\delta(z)$, and
\[
|\rho_\theta(w)-\rho_\theta(z)|
\le
2\|\rho_\theta\|_\infty
\]
everywhere, we obtain
\[
|\tilde\rho_{\theta,\tau}(z)-\rho_\theta(z)|
\le
\varepsilon
+
2\|\rho_\theta\|_\infty
\int_{\mathcal Z\setminus\mathbb B_\delta(z)}
\bar\kappa_\tau(z,w)\,\mu(\mathrm dw).
\]
Taking the supremum over $z\in\mathcal Z$ and invoking
\Cref{eq:tail_mass} gives
\[
\limsup_{\tau\to0^+}
\|\tilde\rho_{\theta,\tau}-\rho_\theta\|_\infty
\le
\varepsilon.
\]
Since $\varepsilon>0$ is arbitrary,
\[
\varepsilon_{\mathrm{kde}}^{(\theta)}(\tau)
:=
\|\tilde\rho_{\theta,\tau}-\rho_\theta\|_\infty
\xrightarrow{\tau\to0^+}
0.
\]
\end{proof}

\section{Proofs of Unimodal Formal Statements}
\label{app:proofs_unimodal}

This section provides rigorous proofs for all formal statements concerning our unimodal analysis. Each statement from the main text is restated for self-containment, accompanied by a proof intuition for providing high-level insights.

\subsection{Proof of \texorpdfstring{\Cref{thm:nce_consistency}}{Thm. 3.1}}
\label{app:gradient_consistency}

\thmnceconsistency*

\begin{proofintuition}
This proof establishes that, for each fixed parameter $\theta$ and temperature $\tau>0$, the expected finite-$N$ InfoNCE objective and its gradient converge, up to the additive value constant $\log\!\big(NV_\kappa(\tau)\big)$, to the deterministic parametric energy $\mathcal J_\tau(\theta)$ and its gradient as $N\to\infty$. The intuition has two parts. \textbf{(i)} In a finite batch, the positive and negative samples compete inside the denominator. As $N\to\infty$, the contribution of the single positive term becomes negligible relative to the negative sum, while the normalized negative sum converges to the population kernel average $\Gamma_{\theta,\tau}$. This yields the alignment and cross-entropy terms appearing in $\mathcal J_\tau$. \textbf{(ii)} The same argument applies to the gradient of the log-partition term: the positive contribution vanishes at rate $O(N^{-1})$, while the ratio formed by the negative-sum gradient and the negative partition sum converges to the gradient of the population log-partition field. Uniform gradient bounds then allow this convergence to pass to the expected finite-$N$ objective.
\end{proofintuition}

\begin{proof}
Consider a random batch $\mathcal B$ consisting of a positive pair and $N$ independent negative samples. Let $(\mathbf x,\tilde{\mathbf x})\sim r_{\mathrm{um}}$, and let the negative samples $\{\mathbf x_j'\}_{j=1}^N$ be drawn i.i.d.\ from the marginal observation distribution $p_{\mathbf x}$, so that
\[
\mathcal B
:=
\{(\mathbf x,\tilde{\mathbf x}),\mathbf x_1',\dots,\mathbf x_N'\}.
\]
Fix any encoder parameter $\theta\in\Theta$, and denote the corresponding random representations by $\mathbf z=f_\theta(\mathbf x)$, $\mathbf v=f_\theta(\tilde{\mathbf x})$, and $\mathbf w_j=f_\theta(\mathbf x_j')$. The corresponding random per-batch unimodal InfoNCE loss is
\begin{equation}
\label{eq:batch_loss_def}
\ell_{\mathcal B}(\theta;\mathbf x)
:=
-\frac{1}{\tau}\mathrm{sim}(\mathbf z,\mathbf v)
+
\log
\underbrace{
\left(
\kappa_\tau(\mathbf z,\mathbf v)
+
\sum_{j=1}^{N}\kappa_\tau(\mathbf z,\mathbf w_j)
\right)
}_{=:\mathsf Z_{\mathcal B}(\mathbf z)},
\end{equation}
where $\kappa_\tau(z,w)=\exp(\mathrm{sim}(z,w)/\tau)$.

\paragraph{Part I. Value consistency.}
We analyze the asymptotic behavior of the loss as $N\to\infty$.

\textbf{Step 1: Isolate the positive bias.}
First, decompose the partition sum $\mathsf Z_{\mathcal B}$ inside the logarithm into the positive term and the sum over negatives:
\[
\mathsf Z_{\mathcal B}
=
\mathsf P+\mathsf S_N,
\qquad
\mathsf P
:=
\kappa_\tau(\mathbf z,\mathbf v),
\qquad
\mathsf S_N
:=
\sum_{j=1}^N
\kappa_\tau(\mathbf z,\mathbf w_j).
\]
The logarithmic term becomes
\[
\log\mathsf Z_{\mathcal B}
=
\log(\mathsf S_N+\mathsf P)
=
\log\mathsf S_N
+
\underbrace{
\log\!\left(
1+\frac{\mathsf P}{\mathsf S_N}
\right)
}_{=:\epsilon_N}.
\]
Fix $\tau>0$. Since $\mathcal Z$ is compact and $\mathrm{sim}$ is continuous on $\mathcal Z\times\mathcal Z$ by \Cref{assump:regularity}\ref{itm:crit_reg}, there exist finite constants $\mathrm{sim}_{\min}$ and $\mathrm{sim}_{\max}$ such that $\mathrm{sim}_{\min}\le\mathrm{sim}(z,w)\le\mathrm{sim}_{\max}$ for all $(z,w)\in\mathcal Z\times\mathcal Z$. Consequently,
\[
0
<
m_\tau
:=
\exp(\mathrm{sim}_{\min}/\tau)
\le
\kappa_\tau(z,w)
\le
\exp(\mathrm{sim}_{\max}/\tau)
=:
M_\tau
<
\infty.
\]
Thus, $m_\tau\le\mathsf P\le M_\tau$ and $Nm_\tau\le\mathsf S_N\le NM_\tau$. Using $\log(1+x)\le x$ for $x\ge0$,
\[
0
<
\epsilon_N
=
\log\!\left(
1+\frac{\mathsf P}{\mathsf S_N}
\right)
\le
\frac{\mathsf P}{\mathsf S_N}
\le
\frac{M_\tau}{Nm_\tau}
=
O_\tau(N^{-1}).
\]
Therefore, for each fixed $\tau>0$,
\[
\log\mathsf Z_{\mathcal B}
=
\log\mathsf S_N
+
O_\tau(N^{-1}).
\]

\textbf{Step 2: Concentration of the negative partition sum.}
Fix $\theta$ and a realization $x$ of the anchor observation, and write $z=f_\theta(x)$. For negatives $\mathbf x_j'\overset{\mathrm{i.i.d.}}{\sim}p_{\mathbf x}$, define $\mathbf w_j:=f_\theta(\mathbf x_j')$ and
\[
\mathsf Y_j(z)
:=
\kappa_\tau(z,\mathbf w_j),
\qquad
\mathsf S_N(z)
:=
\sum_{j=1}^N
\mathsf Y_j(z).
\]
For fixed $z$, the variables $\{\mathsf Y_j(z)\}_{j=1}^N$ are i.i.d.\ with mean
\[
\mu_\kappa(z)
:=
\mathbb E_{\mathbf w\sim q_\theta}
\big[
\kappa_\tau(z,\mathbf w)
\big]
=
\int_{\mathcal Z}
\kappa_\tau(z,w)\,
q_\theta(\mathrm dw)
\equiv
\Gamma_{\theta,\tau}(z)
\ge
m_\tau,
\]
and finite variance. Moreover, for each fixed $\tau>0$, $\kappa_\tau$ is bounded on $\mathcal Z\times\mathcal Z$, so $\mathsf Y_j(z)$ is bounded. By Hoeffding's inequality,
\[
\frac{\mathsf S_N(z)}{N}
-
\mu_\kappa(z)
=
O_p(N^{-1/2}).
\]
Equivalently, writing
\[
\frac{\mathsf S_N(z)}
{N\mu_\kappa(z)}
=
1+\Delta_N(z),
\qquad
\Delta_N(z)
=
O_p(N^{-1/2}),
\]
and noting $\mu_\kappa(z)\ge m_\tau>0$, we have
\[
\log\mathsf S_N(z)
=
\log\!\big(
N\mu_\kappa(z)
\big)
+
\log\!\big(
1+\Delta_N(z)
\big)
=
\log N
+
\log\Gamma_{\theta,\tau}(z)
+
O_p(N^{-1/2}).
\]
Therefore,
\[
\log\mathsf S_N(z)
=
\log N
+
\log\Gamma_{\theta,\tau}(z)
+
O_p(N^{-1/2}).
\]

\textbf{Step 3: Identification of the induced parametric energy.}
Combining Steps~1--2 gives
\[
\log\mathsf Z_{\mathcal B}(\mathbf z)
=
\log N
+
\log\Gamma_{\theta,\tau}(\mathbf z)
+
o_p(1),
\]
where $o_p(1)$ denotes a term vanishing in probability as $N\to\infty$ for fixed $\tau>0$. Substituting into \Cref{eq:batch_loss_def} yields
\begin{equation}
\label{eq:batch_loss_expand}
\ell_{\mathcal B}(\theta;\mathbf x)
=
-\frac{1}{\tau}
\mathrm{sim}(\mathbf z,\mathbf v)
+
\log N
+
\log\Gamma_{\theta,\tau}(\mathbf z)
+
o_p(1).
\end{equation}
Moreover, for fixed $\tau>0$, the remainder in the logarithmic partition term is uniformly bounded in $N$: indeed, both $\mathsf S_N/N$ and $\Gamma_{\theta,\tau}(\mathbf z)$ lie in $[m_\tau,M_\tau]$, while the positive-term correction is $O_\tau(N^{-1})$. Hence the remainder is uniformly integrable. Since it converges to zero in probability, it also converges to zero in $L^1$, and therefore its expectation is $o(1)$.

Taking expectations over the batch construction $(\mathbf x,\tilde{\mathbf x})\sim r_{\mathrm{um}}$ and the i.i.d.\ negatives, and using $\mathcal L_{\mathrm{NCE},N}(\theta)=\mathbb E_{\mathcal B}[\ell_{\mathcal B}(\theta;\mathbf x)]$, we obtain
\begin{equation}
\label{eq:Lnce_pop_expansion}
\mathcal L_{\mathrm{NCE},N}(\theta)
=
\mathbb E_{(\mathbf x,\tilde{\mathbf x})}
\left[
-\frac{1}{\tau}
\mathrm{sim}
\big(
f_\theta(\mathbf x),
f_\theta(\tilde{\mathbf x})
\big)
\right]
+
\mathbb E_{\mathbf x}
\left[
\log\Gamma_{\theta,\tau}
\big(
f_\theta(\mathbf x)
\big)
\right]
+
\log N
+
o(1),
\end{equation}
where $o(1)\to0$ as $N\to\infty$.

By disintegration of $\pi_{\theta\theta}=(f_\theta\times f_\theta)_\#r_{\mathrm{um}}$, define
\[
U_\theta(z)
:=
-
\int_{\mathcal Z}
\mathrm{sim}(z,w)\,
\nu_{\theta,z}(\mathrm dw).
\]
Then the law of iterated expectations gives
\begin{equation}
\label{eq:align_potential}
\mathbb E_{(\mathbf x,\tilde{\mathbf x})}
\left[
-\frac{1}{\tau}
\mathrm{sim}(\mathbf z,\mathbf v)
\right]
=
\frac{1}{\tau}
\mathbb E_{\mathbf z\sim q_\theta}
\big[
U_\theta(\mathbf z)
\big]
=
\frac{1}{\tau}
\int_{\mathcal Z}
U_\theta(z)\,
q_\theta(\mathrm dz).
\end{equation}
Since $\mathbf z=f_\theta(\mathbf x)\sim q_\theta$,
\[
\mathbb E_{\mathbf x}
\left[
\log\Gamma_{\theta,\tau}
\big(
f_\theta(\mathbf x)
\big)
\right]
=
\mathbb E_{\mathbf z\sim q_\theta}
\left[
\log\Gamma_{\theta,\tau}(\mathbf z)
\right].
\]
Under the constant kernel-volume condition \Cref{ass:kernel_volume},
\[
\tilde\rho_{\theta,\tau}(z)
=
\frac{
\Gamma_{\theta,\tau}(z)
}{
V_\kappa(\tau)
}
\]
by \Cref{def:smoothed_density}, hence
\begin{equation}
\label{eq:expect_log_part}
\mathbb E_{\mathbf z\sim q_\theta}
\left[
\log\Gamma_{\theta,\tau}(\mathbf z)
\right]
=
\log V_\kappa(\tau)
+
\mathbb E_{\mathbf z\sim q_\theta}
\left[
\log\tilde\rho_{\theta,\tau}(\mathbf z)
\right]
=
\log V_\kappa(\tau)
-
H_\times
\big(
q_\theta,
\tilde\rho_{\theta,\tau}
\big),
\end{equation}
where
\[
H_\times
\big(
q_\theta,
\tilde\rho_{\theta,\tau}
\big)
:=
-
\mathbb E_{\mathbf z\sim q_\theta}
\left[
\log\tilde\rho_{\theta,\tau}(\mathbf z)
\right].
\]
Substituting \Cref{eq:align_potential,eq:expect_log_part} into \Cref{eq:Lnce_pop_expansion} yields
\[
\begin{aligned}
\mathcal L_{\mathrm{NCE},N}(\theta)
&=
\frac{1}{\tau}
\int_{\mathcal Z}
U_\theta(z)\,
q_\theta(\mathrm dz)
-
H_\times
\big(
q_\theta,
\tilde\rho_{\theta,\tau}
\big)
+
\log\!\big(
NV_\kappa(\tau)
\big)
+
o(1)
\\
&=
\mathcal J_\tau(\theta)
+
\log\!\big(
NV_\kappa(\tau)
\big)
+
o(1).
\end{aligned}
\]
Since $\theta\in\Theta$ was arbitrary, this proves
\[
\left|
\mathcal L_{\mathrm{NCE},N}(\theta)
-
\mathcal J_\tau(\theta)
-
\log\!\big(
NV_\kappa(\tau)
\big)
\right|
\longrightarrow0
\]
for each fixed $\theta\in\Theta$ and $\tau>0$.

\paragraph{Part II. Gradient consistency.}
We now analyze the gradient of the batch loss $\ell_{\mathcal B}(\theta;\mathbf x)$ for fixed $\tau>0$ as $N\to\infty$. Differentiating the finite-batch loss gives
\begin{equation}
\label{eq:batch_grad}
\nabla_\theta
\ell_{\mathcal B}(\theta;\mathbf x)
=
-\frac{1}{\tau}
\nabla_\theta
\mathrm{sim}(\mathbf z,\mathbf v)
+
\nabla_\theta
\log\mathsf Z_{\mathcal B}(\mathbf z),
\qquad
\mathsf Z_{\mathcal B}(\mathbf z)
=
\mathsf P+\mathsf S_N,
\end{equation}
where
\[
\mathsf P
:=
\kappa_\tau(\mathbf z,\mathbf v),
\qquad
\mathsf S_N
:=
\sum_{j=1}^N
\kappa_\tau(\mathbf z,\mathbf w_j).
\]

\textbf{Step 1: Reducing the positive term in the log-partition gradient.}
Let
\[
\mathsf g_N(\mathbf z)
:=
\nabla_\theta
\log(\mathsf P+\mathsf S_N),
\qquad
\mathsf r_N(\mathbf z)
:=
\nabla_\theta
\log\mathsf S_N.
\]
Then
\begin{equation}
\label{eq:pos_removal_identity}
\begin{aligned}
\mathsf g_N-\mathsf r_N
&=
\nabla_\theta
\log(\mathsf P+\mathsf S_N)
-
\nabla_\theta
\log\mathsf S_N
\\
&=
\frac{
\nabla_\theta\mathsf P
+
\nabla_\theta\mathsf S_N
}{
\mathsf P+\mathsf S_N
}
-
\frac{
\nabla_\theta\mathsf S_N
}{
\mathsf S_N
}
\\
&=
\frac{
\nabla_\theta\mathsf P
}{
\mathsf P+\mathsf S_N
}
+
\nabla_\theta\mathsf S_N
\left(
\frac{
\mathsf S_N-(\mathsf P+\mathsf S_N)
}{
\mathsf S_N(\mathsf P+\mathsf S_N)
}
\right)
\\
&=
\frac{
\nabla_\theta\mathsf P
}{
\mathsf P+\mathsf S_N
}
-
\frac{
\mathsf P
}{
\mathsf S_N(\mathsf P+\mathsf S_N)
}
\nabla_\theta\mathsf S_N.
\end{aligned}
\end{equation}
Under \Cref{assump:regularity}, continuity of $\mathrm{sim}$ on the compact domain $\mathcal Z\times\mathcal Z$ ensures that $\kappa_\tau$ is uniformly bounded. Moreover, expanding $\nabla_\theta\kappa_\tau$ by the chain rule yields a product of the kernel, similarity gradients, and encoder Jacobians, all uniformly bounded by assumption. Thus, for each fixed $\tau>0$, there exist finite constants $M_\tau,G_\tau$ such that $0<\kappa_\tau\le M_\tau$ and $\|\nabla_\theta\kappa_\tau\|\le G_\tau$. Since $\kappa_\tau$ is continuous and strictly positive on compact $\mathcal Z\times\mathcal Z$, it also admits the positive lower bound $m_\tau>0$. Hence
\[
\mathsf P
\le
M_\tau,
\qquad
\|\nabla_\theta\mathsf P\|
\le
G_\tau,
\qquad
\mathsf S_N
\ge
Nm_\tau,
\qquad
\|\nabla_\theta\mathsf S_N\|
\le
NG_\tau.
\]
Taking norms in \Cref{eq:pos_removal_identity} and using these bounds gives
\[
\|\mathsf g_N-\mathsf r_N\|
\le
\frac{G_\tau}{Nm_\tau}
+
\frac{M_\tau}{(Nm_\tau)^2}
NG_\tau
=
O_\tau(N^{-1}).
\]
Therefore,
\begin{equation}
\label{eq:pos_removed}
\nabla_\theta
\log\mathsf Z_{\mathcal B}(\mathbf z)
=
\nabla_\theta
\log\mathsf S_N(\mathbf z)
+
O_\tau(N^{-1}).
\end{equation}

\textbf{Step 2: Consistency of the negative-sum ratio.}
Condition on the event that the anchor observation takes the realization $x$, i.e., $\mathbf x=x$, and write $z=f_\theta(x)$. Then
\[
\nabla_\theta
\log\mathsf S_N(z)
=
\frac{
\frac{1}{N}
\sum_{j=1}^N
\nabla_\theta
\kappa_\tau
\big(
f_\theta(x),
f_\theta(\mathbf x_j')
\big)
}{
\frac{1}{N}
\sum_{j=1}^N
\kappa_\tau
\big(
f_\theta(x),
f_\theta(\mathbf x_j')
\big)
}.
\]
Given $x$, the negatives $\mathbf x_j'$ are i.i.d.\ with law $p_{\mathbf x}$, and both $\kappa_\tau(f_\theta(x),f_\theta(\mathbf x_j'))$ and its total parameter gradient are uniformly bounded for fixed $\tau>0$ by \Cref{assump:regularity}. Hence, by the law of large numbers,
\[
\frac{1}{N}
\sum_{j=1}^N
\kappa_\tau
\big(
f_\theta(x),
f_\theta(\mathbf x_j')
\big)
\xrightarrow{p}
\Gamma_{\theta,\tau}
\big(
f_\theta(x)
\big),
\]
and
\[
\frac{1}{N}
\sum_{j=1}^N
\nabla_\theta
\kappa_\tau
\big(
f_\theta(x),
f_\theta(\mathbf x_j')
\big)
\xrightarrow{p}
\mathbb E_{\mathbf x'\sim p_{\mathbf x}}
\left[
\nabla_\theta
\kappa_\tau
\big(
f_\theta(x),
f_\theta(\mathbf x')
\big)
\right].
\]
By the uniform gradient bound, differentiation under the expectation is justified, so
\[
\mathbb E_{\mathbf x'\sim p_{\mathbf x}}
\left[
\nabla_\theta
\kappa_\tau
\big(
f_\theta(x),
f_\theta(\mathbf x')
\big)
\right]
=
\nabla_\theta
\mathbb E_{\mathbf x'\sim p_{\mathbf x}}
\left[
\kappa_\tau
\big(
f_\theta(x),
f_\theta(\mathbf x')
\big)
\right]
=
\nabla_\theta
\Gamma_{\theta,\tau}
\big(
f_\theta(x)
\big),
\]
where the derivative on the right is the total derivative with respect to $\theta$, including the dependence of the anchor representation $f_\theta(x)$ on $\theta$. Since $\Gamma_{\theta,\tau}(f_\theta(x))\ge m_\tau>0$, the continuous mapping theorem yields
\begin{equation}
\label{eq:ratio_consistency}
\nabla_\theta
\log\mathsf S_N
\big(
f_\theta(x)
\big)
\xrightarrow{p}
\frac{
\nabla_\theta
\Gamma_{\theta,\tau}
\big(
f_\theta(x)
\big)
}{
\Gamma_{\theta,\tau}
\big(
f_\theta(x)
\big)
}
=
\nabla_\theta
\log
\Gamma_{\theta,\tau}
\big(
f_\theta(x)
\big).
\end{equation}
Combining \Cref{eq:pos_removed,eq:ratio_consistency} gives
\begin{equation}
\label{eq:logZ_grad_limit}
\nabla_\theta
\log\mathsf Z_{\mathcal B}
\big(
f_\theta(x)
\big)
\xrightarrow{p}
\nabla_\theta
\log
\Gamma_{\theta,\tau}
\big(
f_\theta(x)
\big).
\end{equation}

\textbf{Step 3: Passing to the population objective and identifying $\nabla_\theta\mathcal J_\tau$.}
Recall that
\[
\mathcal L_{\mathrm{NCE},N}(\theta)
=
\mathbb E_{\mathcal B}
\big[
\ell_{\mathcal B}(\theta;\mathbf x)
\big],
\qquad
\nabla_\theta
\mathcal L_{\mathrm{NCE},N}(\theta)
=
\mathbb E_{\mathcal B}
\big[
\nabla_\theta
\ell_{\mathcal B}(\theta;\mathbf x)
\big]
\]
by \Cref{lem:asym_grad_bound}. From \Cref{eq:batch_grad,eq:logZ_grad_limit}, we have convergence in probability
\[
\nabla_\theta
\ell_{\mathcal B}(\theta;\mathbf x)
\xrightarrow{p}
-\frac{1}{\tau}
\nabla_\theta
\mathrm{sim}
\big(
f_\theta(\mathbf x),
f_\theta(\tilde{\mathbf x})
\big)
+
\nabla_\theta
\log
\Gamma_{\theta,\tau}
\big(
f_\theta(\mathbf x)
\big).
\]
Moreover, by \Cref{lem:asym_grad_bound}, the per-batch gradients are uniformly bounded for fixed $\tau>0$ and hence uniformly integrable. Together with the convergence in probability above, this implies convergence in $L^1$. Therefore, we may pass to expectations and obtain
\[
\nabla_\theta
\mathcal L_{\mathrm{NCE},N}(\theta)
\longrightarrow
-\frac{1}{\tau}
\mathbb E_{r_{\mathrm{um}}}
\left[
\nabla_\theta
\mathrm{sim}
\big(
f_\theta(\mathbf x),
f_\theta(\tilde{\mathbf x})
\big)
\right]
+
\mathbb E_{\mathbf x\sim p_{\mathbf x}}
\left[
\nabla_\theta
\log
\Gamma_{\theta,\tau}
\big(
f_\theta(\mathbf x)
\big)
\right].
\tag{$\star$}
\]

We now identify the two terms in $(\star)$ as the gradient of the parametric energy $\mathcal J_\tau(\theta)$. Recall that
\[
\int_{\mathcal Z}
U_\theta(z)\,
q_\theta(\mathrm dz)
:=
\mathbb E_{(\mathbf x,\tilde{\mathbf x})\sim r_{\mathrm{um}}}
\left[
-\mathrm{sim}
\big(
f_\theta(\mathbf x),
f_\theta(\tilde{\mathbf x})
\big)
\right].
\]
Under \Cref{assump:regularity}, the parameter gradient of the similarity term is uniformly bounded, so differentiation under the expectation gives
\[
\nabla_\theta
\int_{\mathcal Z}
U_\theta(z)\,
q_\theta(\mathrm dz)
=
-
\mathbb E_{r_{\mathrm{um}}}
\left[
\nabla_\theta
\mathrm{sim}
\big(
f_\theta(\mathbf x),
f_\theta(\tilde{\mathbf x})
\big)
\right].
\]
Therefore, the first term in $(\star)$ equals
\[
\frac{1}{\tau}
\nabla_\theta
\int_{\mathcal Z}
U_\theta(z)\,
q_\theta(\mathrm dz).
\]
By definition,
\[
H_\times
\big(
q_\theta,
\tilde\rho_{\theta,\tau}
\big)
:=
-
\mathbb E_{\mathbf z\sim q_\theta}
\left[
\log
\tilde\rho_{\theta,\tau}(\mathbf z)
\right]
=
-
\mathbb E_{\mathbf x\sim p_{\mathbf x}}
\left[
\log
\tilde\rho_{\theta,\tau}
\big(
f_\theta(\mathbf x)
\big)
\right].
\]
Under \Cref{ass:kernel_volume},
\[
\tilde\rho_{\theta,\tau}(z)
=
\frac{
\Gamma_{\theta,\tau}(z)
}{
V_\kappa(\tau)
},
\]
with $V_\kappa(\tau)$ independent of $\theta$ by \Cref{def:smoothed_density}. Hence
\[
\nabla_\theta
H_\times
\big(
q_\theta,
\tilde\rho_{\theta,\tau}
\big)
=
-
\mathbb E_{\mathbf x\sim p_{\mathbf x}}
\left[
\nabla_\theta
\log
\tilde\rho_{\theta,\tau}
\big(
f_\theta(\mathbf x)
\big)
\right]
=
-
\mathbb E_{\mathbf x\sim p_{\mathbf x}}
\left[
\nabla_\theta
\log
\Gamma_{\theta,\tau}
\big(
f_\theta(\mathbf x)
\big)
\right],
\]
where differentiation under the expectation is justified by the uniform boundedness established above. Thus, the second term in $(\star)$ equals
\[
-
\nabla_\theta
H_\times
\big(
q_\theta,
\tilde\rho_{\theta,\tau}
\big).
\]
Combining the two terms gives
\[
\nabla_\theta
\mathcal L_{\mathrm{NCE},N}(\theta)
\longrightarrow
\frac{1}{\tau}
\nabla_\theta
\int_{\mathcal Z}
U_\theta(z)\,
q_\theta(\mathrm dz)
-
\nabla_\theta
H_\times
\big(
q_\theta,
\tilde\rho_{\theta,\tau}
\big)
=
\nabla_\theta
\mathcal J_\tau(\theta).
\]
Since $\theta\in\Theta$ was arbitrary, this proves
\[
\left\|
\nabla_\theta
\mathcal L_{\mathrm{NCE},N}(\theta)
-
\nabla_\theta
\mathcal J_\tau(\theta)
\right\|
\longrightarrow0
\]
for each fixed $\theta\in\Theta$ and $\tau>0$.
\end{proof}

\subsection{Proof of \texorpdfstring{\Cref{prop:gibbs_equilibrium}}{Prop. 3.1}}
\label{app:gibbs_equilibrium}

\propgibbsequilibrium*

\begin{proofintuition}
The objective of this proof is to show that the representation energy functional admits a unique Gibbs equilibrium. The argument has three steps. First, the space of probability densities is convex. Second, the potential term is linear in the density, while the negative entropy is strictly convex, making the full functional strictly convex on its effective domain. Finally, a variational calculation identifies the Gibbs density as the natural stationary candidate, and rewriting the free-energy gap from this density as a KL divergence verifies its global optimality and uniqueness.
\end{proofintuition}

\begin{proof}
We analyze the functional $\mathcal F_{\tau,U}$ on the space of probability densities
\[
\mathcal D_\mu(\mathcal Z)
:=
\left\{
\rho\in L^1(\mu):
\rho\ge0\ \mu\text{-a.e.},
\int_{\mathcal Z}\rho(z)\,\mu(\mathrm dz)=1
\right\}.
\]
The proof proceeds in three steps: establishing the geometry of the domain, proving strict convexity on the effective domain, and identifying the unique minimizer.

\textbf{Step 1: Convexity of the domain $\mathcal D_\mu(\mathcal Z)$.}
Consider any two densities $\rho_1,\rho_2\in\mathcal D_\mu(\mathcal Z)$ and a mixing coefficient $\lambda\in(0,1)$. Define $\rho_\lambda:=\lambda\rho_1+(1-\lambda)\rho_2$. Since $\rho_1,\rho_2\ge0$ $\mu$-a.e., we have $\rho_\lambda\ge0$ $\mu$-a.e. Moreover, by linearity of the integral,
\[
\int_{\mathcal Z}\rho_\lambda(z)\,\mu(\mathrm dz)
=
\lambda\int_{\mathcal Z}\rho_1(z)\,\mu(\mathrm dz)
+
(1-\lambda)\int_{\mathcal Z}\rho_2(z)\,\mu(\mathrm dz)
=
1.
\]
Therefore, $\rho_\lambda\in\mathcal D_\mu(\mathcal Z)$ for every $\lambda\in(0,1)$, so $\mathcal D_\mu(\mathcal Z)$ is convex.

\textbf{Step 2: Strict functional convexity.}
Let $\rho_1,\rho_2\in\mathcal D_\mu(\mathcal Z)$ be two distinct densities, i.e., $\rho_1\not\equiv\rho_2$ $\mu$-a.e., and let $\lambda\in(0,1)$. Define $\rho_\lambda:=\lambda\rho_1+(1-\lambda)\rho_2$. We consider densities in the effective domain of $\mathcal F_{\tau,U}$, where the functional is finite.

The potential-energy term is linear in $\rho$:
\begin{align*}
\frac{1}{\tau}\int_{\mathcal Z}U(z)\rho_\lambda(z)\,\mu(\mathrm dz)
&=
\frac{\lambda}{\tau}\int_{\mathcal Z}U(z)\rho_1(z)\,\mu(\mathrm dz)
+
\frac{1-\lambda}{\tau}\int_{\mathcal Z}U(z)\rho_2(z)\,\mu(\mathrm dz).
\end{align*}
The negative entropy is
\[
-H(\rho)
=
\int_{\mathcal Z}\rho(z)\log\rho(z)\,\mu(\mathrm dz),
\]
with the convention $0\log0=0$. Since $\varphi(t)=t\log t$ is strictly convex on $[0,\infty)$, we have
\[
\int_{\mathcal Z}\varphi(\rho_\lambda(z))\,\mu(\mathrm dz)
<
\lambda\int_{\mathcal Z}\varphi(\rho_1(z))\,\mu(\mathrm dz)
+
(1-\lambda)\int_{\mathcal Z}\varphi(\rho_2(z))\,\mu(\mathrm dz)
\]
whenever $\rho_1$ and $\rho_2$ differ on a set of positive $\mu$-measure. Hence $-H$ is strictly convex on its effective domain. Combining the linear potential term with the strictly convex negative entropy gives
\[
\mathcal F_{\tau,U}(\rho_\lambda)
<
\lambda\mathcal F_{\tau,U}(\rho_1)
+
(1-\lambda)\mathcal F_{\tau,U}(\rho_2).
\]
Therefore, $\mathcal F_{\tau,U}$ is strictly convex on its effective domain.

\textbf{Step 3: Existence and uniqueness of the global minimizer.}
Since $\mathcal F_{\tau,U}$ is strictly convex on its effective domain, any minimizer, if it exists, is unique. For fixed $\tau>0$, we first derive the candidate minimizer using the method of Lagrange multipliers. Define the Lagrangian
\[
\Lambda(\rho,\gamma)
:=
\mathcal F_{\tau,U}(\rho)
+
\gamma
\left(
\int_{\mathcal Z}\rho(z)\,\mu(\mathrm dz)-1
\right).
\]
Formally taking the functional derivative with respect to $\rho(z)$ and imposing stationarity gives
\[
\frac{\delta\Lambda}{\delta\rho(z)}
=
\frac{1}{\tau}U(z)
+
\log\rho(z)
+
1+\gamma
=
0.
\]
Hence the stationary density must satisfy
\[
\log\rho^*(z)
=
-\frac{U(z)}{\tau}
-
\gamma
-
1,
\]
and therefore
\[
\rho^*(z)
=
\frac{1}{Z_\tau}
\exp\!\left(
-\frac{U(z)}{\tau}
\right),
\qquad
Z_\tau
:=
\int_{\mathcal Z}
\exp\!\left(
-\frac{U(z)}{\tau}
\right)
\mu(\mathrm dz).
\]
Since $\operatorname*{ess\,inf}_{\mu}U>-\infty$ by assumption, there exists $m\in\mathbb R$ such that $U(z)\ge m$ for $\mu$-a.e.\ $z$. Hence
\[
0
<
Z_\tau
\le
e^{-m/\tau}\mu(\mathcal Z)
<
\infty,
\]
so $\rho^*$ is a well-defined probability density in $\mathcal D_\mu(\mathcal Z)$.

It remains to verify that this stationary density is indeed the global minimizer. Since
\[
\log\rho^*(z)
=
-\frac{U(z)}{\tau}
-
\log Z_\tau,
\]
for any $\rho$ in the effective domain of $\mathcal F_{\tau,U}$,
\begin{align*}
D_{\mathrm{KL}}(\rho\|\rho^*)
&=
\int_{\mathcal Z}
\rho(z)
\log\frac{\rho(z)}{\rho^*(z)}
\,\mu(\mathrm dz)
\\
&=
\int_{\mathcal Z}
\rho(z)\log\rho(z)\,\mu(\mathrm dz)
+
\frac{1}{\tau}
\int_{\mathcal Z}
U(z)\rho(z)\,\mu(\mathrm dz)
+
\log Z_\tau
\\
&=
\mathcal F_{\tau,U}(\rho)
+
\log Z_\tau.
\end{align*}
Equivalently,
\[
\mathcal F_{\tau,U}(\rho)
=
-\log Z_\tau
+
D_{\mathrm{KL}}(\rho\|\rho^*).
\]
Since $D_{\mathrm{KL}}(\rho\|\rho^*)\ge0$, with equality if and only if $\rho=\rho^*$ $\mu$-a.e., we have
\[
\mathcal F_{\tau,U}(\rho)
\ge
-\log Z_\tau
=
\mathcal F_{\tau,U}(\rho^*),
\]
with equality if and only if $\rho=\rho^*$ $\mu$-a.e. Therefore, $\rho^*$ is the unique global minimizer of $\mathcal F_{\tau,U}$.
\end{proof}

\subsection{Proof of \texorpdfstring{\Cref{prop:ground_state}}{Prop. 3.2}}
\label{app:ground_state}

\propgroundstate*

\begin{proofintuition}
This proof establishes that as the temperature $\tau\to0^+$, the Gibbs distribution concentrates near the lowest-potential regions (the $\sigma$-ground state). The argument relies on a direct comparison of exponential decay rates relative to the essential minimum level $m$. We compare the probability mass between two regions: \textbf{(i)} outside the $\sigma$-ground state, the potential is at least $\sigma$ above $m$, so the corresponding Gibbs weight is bounded by a factor of $\exp(-\sigma/\tau)$; \textbf{(ii)} by the definition of the essential infimum, there is a set of positive $\mu$-measure on which the potential lies within $\sigma/2$ of $m$, providing a denominator contribution bounded below by $\exp(-\sigma/(2\tau))$. Taking the ratio of these bounds yields an exponentially vanishing upper bound on the mass outside the ground-state neighborhood as $\tau\to0^+$.
\end{proofintuition}

\begin{proof}
We analyze the Gibbs-equilibrium measure $\rho^*\mu$ as $\tau\to0^+$. Let
\[
m:=\operatorname*{ess\,inf}_{\mu}U
\qquad\text{and}\qquad
\mathcal W^\sigma:=\left\{z\in\mathcal Z:U(z)\le m+\sigma\right\},
\]
for any fixed $\sigma>0$.

Consider any open set $\mathcal O\subset\mathcal Z$ such that $\mathcal O\supset\mathcal W^\sigma$, and denote $\mathcal O^c:=\mathcal Z\setminus\mathcal O$. Since $\mathcal O$ contains $\mathcal W^\sigma$, its complement satisfies
\[
\mathcal O^c
\subset
\left\{
z\in\mathcal Z:
U(z)>m+\sigma
\right\}.
\]
Therefore, using the Gibbs density
\[
\rho^*(z)
=
\frac{1}{Z_\tau}
\exp\!\left(-\frac{U(z)}{\tau}\right),
\]
we have
\[
(\rho^*\mu)(\mathcal O^c)
=
\frac{
\int_{\mathcal O^c}
\exp\!\left(-\frac{U(z)}{\tau}\right)
\mu(\mathrm dz)
}{
\int_{\mathcal Z}
\exp\!\left(-\frac{U(w)}{\tau}\right)
\mu(\mathrm dw)
}.
\]
Multiplying numerator and denominator by $\exp(m/\tau)$ yields
\[
(\rho^*\mu)(\mathcal O^c)
=
\frac{
\int_{\mathcal O^c}
\exp\!\left(-\frac{U(z)-m}{\tau}\right)
\mu(\mathrm dz)
}{
\int_{\mathcal Z}
\exp\!\left(-\frac{U(w)-m}{\tau}\right)
\mu(\mathrm dw)
}.
\]

\textbf{(i) Numerator bound.}
For every $z\in\mathcal O^c$, we have $U(z)-m>\sigma$, and hence
\[
\int_{\mathcal O^c}
\exp\!\left(-\frac{U(z)-m}{\tau}\right)
\mu(\mathrm dz)
\le
\int_{\mathcal O^c}
\exp\!\left(-\frac{\sigma}{\tau}\right)
\mu(\mathrm dz)
=
\mu(\mathcal O^c)
\exp\!\left(-\frac{\sigma}{\tau}\right).
\]

\textbf{(ii) Denominator lower bound.}
By the definition of $m=\operatorname*{ess\,inf}_{\mu}U$, for every $\eta>0$ the set $\{w\in\mathcal Z:U(w)<m+\eta\}$ has strictly positive $\mu$-measure. Fix $\eta:=\sigma/2$ and define
\[
\mathcal A
:=
\left\{
w\in\mathcal Z:
U(w)<m+\frac{\sigma}{2}
\right\},
\qquad
\mu(\mathcal A)>0.
\]
On $\mathcal A$, we have $U(w)-m<\sigma/2$, hence
\[
\int_{\mathcal Z}
\exp\!\left(-\frac{U(w)-m}{\tau}\right)
\mu(\mathrm dw)
\ge
\int_{\mathcal A}
\exp\!\left(-\frac{U(w)-m}{\tau}\right)
\mu(\mathrm dw)
\ge
\mu(\mathcal A)
\exp\!\left(-\frac{\sigma}{2\tau}\right).
\]

Combining the bounds, we obtain
\[
(\rho^*\mu)(\mathcal O^c)
\le
\frac{
\mu(\mathcal O^c)\exp(-\sigma/\tau)
}{
\mu(\mathcal A)\exp(-\sigma/(2\tau))
}
=
\frac{\mu(\mathcal O^c)}{\mu(\mathcal A)}
\exp\!\left(-\frac{\sigma}{2\tau}\right).
\]
Since $\mu(\mathcal O^c)\le\mu(\mathcal Z)<\infty$ and $\mu(\mathcal A)>0$ is fixed, the right-hand side converges to $0$ as $\tau\to0^+$. Hence, for any open neighborhood $\mathcal O\supset\mathcal W^\sigma$, we have $(\rho^*\mu)(\mathcal O^c)\to0$, i.e., the Gibbs measure concentrates in any open neighborhood of $\mathcal W^\sigma$.
\end{proof}

\subsection{Proof of \texorpdfstring{\Cref{thm:intrinsic_unimodal}}{Thm. 3.2}}
\label{app:intrinsic_unimodal}

\thmintrinsicunimodal*

\begin{proofintuition}
The objective of this proof is to establish that at low temperatures, the parametric energy $\mathcal J_\tau(\theta)$ approaches the intrinsic thermodynamic free energy $\mathcal F_{\tau,U_\theta}(\rho_\theta)$ for each fixed $\theta$. The argument hinges on translating their algebraic difference into an information-theoretic distance and then squeezing that distance to zero. \textbf{(i)} We first algebraically decompose the parametric energy and reveal a fundamental identity: the exact gap between the parametric and intrinsic energies is the KL divergence between the induced representation density $\rho_\theta$ and its kernel-smoothed counterpart $\tilde\rho_{\theta,\tau}$. \textbf{(ii)} To prove this KL divergence vanishes, we ensure that the logarithmic ratio inside the integral remains controlled. By assumption, the density $\rho_\theta$ is strictly bounded away from zero. Because \Cref{lem:sharp_peak_smoothing} guarantees that the smoothed density approaches the true density uniformly as $\tau\to0^+$, the smoothed density also remains uniformly bounded away from zero for sufficiently small $\tau$. \textbf{(iii)} The logarithm therefore operates in a Lipschitz regime, allowing the KL divergence to be controlled directly by the KDE mismatch. Since this mismatch vanishes as $\tau\to0^+$, the parametric and intrinsic energy values agree asymptotically for each fixed $\theta$.
\end{proofintuition}

\begin{proof}
\textbf{Step 1: Well-posedness of the alignment potential term.}
Fix any $\theta\in\Theta$. Recall the definition of the alignment potential field (\Cref{def:align_potential}):
\[
U_\theta(z)
:=
-\int_{\mathcal Z}\mathrm{sim}(z,w)\,\nu_{\theta,z}(\mathrm dw)
\qquad\text{for }q_\theta\text{-a.e. }z.
\]
Under \Cref{assump:regularity}, the similarity function $\mathrm{sim}$ is bounded and Borel measurable, hence $U_\theta$ is Borel measurable and bounded, and therefore integrable with respect to $q_\theta$. Moreover, $q_\theta\ll\mu$ with a continuous density $\rho_\theta$ bounded away from zero by assumption. Since $\mathcal Z$ is compact, $\rho_\theta$ is also bounded above, so $\rho_\theta\log\rho_\theta$ is integrable. Thus, the intrinsic representation functional (\Cref{eq:free_energy_def})
\[
\mathcal F_{\tau,U_\theta}(\rho_\theta)
:=
\frac{1}{\tau}\int_{\mathcal Z}U_\theta(z)\rho_\theta(z)\,\mu(\mathrm dz)
-
H(\rho_\theta)
\]
is well-defined for each $\tau>0$.

\textbf{Step 2: Difference reduces to a KL divergence.}
Recall
\[
\mathcal J_\tau(\theta)
=
\frac{1}{\tau}\int_{\mathcal Z}U_\theta(z)\,q_\theta(\mathrm dz)
-
H_\times\!\left(q_\theta,\tilde\rho_{\theta,\tau}\right).
\]
Using
\[
H_\times(q,\tilde\rho)
=
H(\rho)
+
D_{\mathrm{KL}}(\rho\|\tilde\rho),
\qquad
\rho(z):=\frac{q(\mathrm dz)}{\mu(\mathrm dz)},
\]
we obtain
\[
\mathcal J_\tau(\theta)
=
\frac{1}{\tau}\int_{\mathcal Z}U_\theta(z)\rho_\theta(z)\,\mu(\mathrm dz)
-
H(\rho_\theta)
-
D_{\mathrm{KL}}(\rho_\theta\|\tilde\rho_{\theta,\tau}),
\]
and hence
\begin{equation}
\label{eq:geo_difference_revised}
\mathcal F_{\tau,U_\theta}(\rho_\theta)
-
\mathcal J_\tau(\theta)
=
D_{\mathrm{KL}}(\rho_\theta\|\tilde\rho_{\theta,\tau}).
\end{equation}

\textbf{Step 3: Vanishing KL divergence of the smoothing.}
Fix any $\theta\in\Theta$. By assumption, the continuous density is uniformly bounded away from zero:
\[
\inf_{z\in\mathcal Z}\rho_\theta(z)
\ge
\underline\rho_\theta
>
0.
\]
Recall from \Cref{lem:sharp_peak_smoothing} that the KDE mismatch vanishes in the low-temperature limit:
\[
\varepsilon_{\mathrm{kde}}^{(\theta)}(\tau)
:=
\|\tilde\rho_{\theta,\tau}-\rho_\theta\|_\infty
\longrightarrow0
\qquad\text{as }\tau\to0^+.
\]
Hence there exists $\tau_0(\theta)>0$ such that, for all $0<\tau\le\tau_0(\theta)$,
\[
\varepsilon_{\mathrm{kde}}^{(\theta)}(\tau)
\le
\frac{\underline\rho_\theta}{2}.
\]
It follows that, for every $z\in\mathcal Z$,
\begin{equation}
\label{eq:tilde_lower}
\tilde\rho_{\theta,\tau}(z)
\ge
\rho_\theta(z)
-
\frac{\underline\rho_\theta}{2}
\ge
\frac{\underline\rho_\theta}{2}.
\end{equation}
On $[\underline\rho_\theta/2,\infty)$, the derivative of $\log x$ satisfies $1/x\le2/\underline\rho_\theta$. Hence, for all $a,b\ge\underline\rho_\theta/2$,
\[
|\log a-\log b|
\le
\frac{2}{\underline\rho_\theta}|a-b|.
\]
Applying this with $a=\rho_\theta(z)$ and $b=\tilde\rho_{\theta,\tau}(z)$ gives
\begin{equation}
\label{eq:ratio_log_bound}
\left|
\log\frac{\rho_\theta(z)}{\tilde\rho_{\theta,\tau}(z)}
\right|
=
\left|
\log\rho_\theta(z)
-
\log\tilde\rho_{\theta,\tau}(z)
\right|
\le
\frac{2}{\underline\rho_\theta}
\left|
\rho_\theta(z)
-
\tilde\rho_{\theta,\tau}(z)
\right|
\le
\frac{2\,\varepsilon_{\mathrm{kde}}^{(\theta)}(\tau)}
{\underline\rho_\theta}.
\end{equation}
Now
\[
D_{\mathrm{KL}}(\rho_\theta\|\tilde\rho_{\theta,\tau})
=
\int_{\mathcal Z}
\rho_\theta(z)
\log\frac{\rho_\theta(z)}{\tilde\rho_{\theta,\tau}(z)}
\,\mu(\mathrm dz).
\]
Using $\int_{\mathcal Z}\rho_\theta(z)\,\mu(\mathrm dz)=1$ and \Cref{eq:ratio_log_bound},
\begin{equation}
\label{eq:KL_sup_bound}
D_{\mathrm{KL}}(\rho_\theta\|\tilde\rho_{\theta,\tau})
\le
\int_{\mathcal Z}
\rho_\theta(z)
\left|
\log\frac{\rho_\theta(z)}{\tilde\rho_{\theta,\tau}(z)}
\right|
\mu(\mathrm dz)
\le
\frac{2\,\varepsilon_{\mathrm{kde}}^{(\theta)}(\tau)}
{\underline\rho_\theta}.
\end{equation}
Substituting this into \Cref{eq:geo_difference_revised} yields
\[
\left|
\mathcal J_\tau(\theta)
-
\mathcal F_{\tau,U_\theta}(\rho_\theta)
\right|
\le
\frac{2\,\varepsilon_{\mathrm{kde}}^{(\theta)}(\tau)}
{\underline\rho_\theta}.
\]
Finally, \Cref{lem:sharp_peak_smoothing} guarantees that $\varepsilon_{\mathrm{kde}}^{(\theta)}(\tau)\to0$ as $\tau\to0^+$. Consequently,
\[
\left|
\mathcal J_\tau(\theta)
-
\mathcal F_{\tau,U_\theta}(\rho_\theta)
\right|
\longrightarrow0
\qquad\text{as }\tau\to0^+,
\]
which concludes the proof.
\end{proof}

\section{Proofs of Multimodal Formal Statements}
\label{app:proofs_multimodal}

This section provides rigorous proofs for all formal statements concerning our multimodal analysis. Each statement from the main text is restated for self-containment, accompanied by a proof intuition for providing high-level insights.

\subsection{Proof of \texorpdfstring{\Cref{thm:sym_consistency}}{4.1}}
\label{app:sym_consistency}

\thmsymconsistency*

\begin{proofintuition}
This proof extends the large-batch consistency analysis of unimodal InfoNCE to the dual-encoder, cross-modal setting. The logic follows the unimodal proof (\Cref{thm:nce_consistency}) in each direction. \textbf{(i)} The symmetric loss is the average of two directional losses. As $N\to\infty$, the normalized sum of negative keys in each direction converges to the population kernel average of the corresponding target modality. This yields a shared attractive alignment term together with two directional cross-entropy terms. \textbf{(ii)} The same large-batch limit applies to the gradients of the two log-partition terms: the contribution of the single positive pair vanishes, while the negative-sum ratios converge to their population counterparts. Uniform gradient bounds then allow these limits to pass to the expected finite-$N$ objective, yielding pointwise value and gradient consistency with the deterministic symmetric energy.
\end{proofintuition}

\begin{proof}
Consider a random batch $\mathcal B$ consisting of a positive pair $(\mathbf x,\mathbf y)\sim r_{\mathrm{mm}}$ together with $N$ i.i.d.\ negatives $\{\mathbf y_j'\}_{j=1}^N\overset{\mathrm{i.i.d.}}{\sim}p_{\mathbf y}$ and $N$ i.i.d.\ negatives $\{\mathbf x_j'\}_{j=1}^N\overset{\mathrm{i.i.d.}}{\sim}p_{\mathbf x}$, all independent conditional on $(\mathbf x,\mathbf y)$. Write the corresponding random representations as
\[
\mathbf z=f_\theta(\mathbf x),
\qquad
\mathbf w=g_\phi(\mathbf y),
\qquad
\mathbf w_j'=g_\phi(\mathbf y_j'),
\qquad
\mathbf z_j'=f_\theta(\mathbf x_j'),
\]
and define the exponential kernel
\[
\kappa_\tau(z,w)
:=
\exp\!\left(\frac{\mathrm{sim}(z,w)}{\tau}\right).
\]
Denote the directional partition sums by
\[
\mathsf Z_{\mathcal B}^{\theta\shortto\phi}(\mathbf z)
:=
\kappa_\tau(\mathbf z,\mathbf w)
+
\sum_{j=1}^N
\kappa_\tau(\mathbf z,\mathbf w_j'),
\qquad
\mathsf Z_{\mathcal B}^{\phi\shortto\theta}(\mathbf w)
:=
\kappa_\tau(\mathbf z,\mathbf w)
+
\sum_{j=1}^N
\kappa_\tau(\mathbf z_j',\mathbf w).
\]
The symmetric per-batch loss can be written as
\begin{equation}
\label{eq:mm_batch_loss}
\ell_{\mathcal B}^{\mathrm{Sym}}(\theta,\phi)
=
\frac12
\underbrace{
\left[
-\frac{1}{\tau}\mathrm{sim}(\mathbf z,\mathbf w)
+
\log\mathsf Z_{\mathcal B}^{\theta\shortto\phi}(\mathbf z)
\right]
}_{\ell_{\mathcal B}^{x\shortto y}(\theta,\phi)}
+
\frac12
\underbrace{
\left[
-\frac{1}{\tau}\mathrm{sim}(\mathbf z,\mathbf w)
+
\log\mathsf Z_{\mathcal B}^{\phi\shortto\theta}(\mathbf w)
\right]
}_{\ell_{\mathcal B}^{y\shortto x}(\theta,\phi)}.
\end{equation}

\paragraph{Part I. Value consistency.}
Fix any $\tau>0$ and $(\theta,\phi)\in\Theta\times\Phi$. We analyze the directional loss $\mathcal L_{\mathrm{NCE},N}^{x\shortto y}(\theta,\phi)$; the reverse direction follows by symmetry. From \Cref{eq:mm_batch_loss}, the corresponding per-batch directional loss is
\begin{equation}
\label{eq:mm_batch_loss_theta_to_phi}
\ell_{\mathcal B}^{x\shortto y}(\theta,\phi)
:=
-\frac{1}{\tau}\mathrm{sim}(\mathbf z,\mathbf w)
+
\log\mathsf Z_{\mathcal B}^{\theta\shortto\phi}(\mathbf z).
\end{equation}

\textbf{Step 1: Bias--variance control of the log-partition term.}
The structure of $\mathsf Z_{\mathcal B}^{\theta\shortto\phi}$ is identical to the unimodal partition function (\Cref{eq:batch_loss_def}), except that the negatives are drawn from $q_\phi$ rather than $q_\theta$. By assumption, $\mathcal Z$ is compact and $\mathrm{sim}$ is continuous, hence $\kappa_\tau$ is bounded on $\mathcal Z\times\mathcal Z$ for each fixed $\tau>0$. Let
\[
\mathsf P
:=
\kappa_\tau(\mathbf z,\mathbf w),
\qquad
\mathsf S_N^{\theta\shortto\phi}(\mathbf z)
:=
\sum_{j=1}^N
\kappa_\tau(\mathbf z,\mathbf w_j'),
\qquad
\mathsf Z_{\mathcal B}^{\theta\shortto\phi}
=
\mathsf P+\mathsf S_N^{\theta\shortto\phi}.
\]
As in Step~1 of the unimodal proof (Part I of \Cref{thm:nce_consistency}), the positive contribution induces a vanishing bias:
\[
\log\!\left(
\mathsf P+\mathsf S_N^{\theta\shortto\phi}
\right)
=
\log\mathsf S_N^{\theta\shortto\phi}
+
\log\!\left(
1+
\frac{\mathsf P}
{\mathsf S_N^{\theta\shortto\phi}}
\right),
\qquad
\log\!\left(
1+
\frac{\mathsf P}
{\mathsf S_N^{\theta\shortto\phi}}
\right)
=
O_\tau(N^{-1}).
\]
Define the population partition function induced by the target encoder $\phi$:
\[
\Gamma_{\phi,\tau}(z)
:=
\mathbb E_{\mathbf w\sim q_\phi}
\left[
\kappa_\tau(z,\mathbf w)
\right]
=
\int_{\mathcal Z}
\kappa_\tau(z,w)\,
q_\phi(\mathrm dw).
\]
Condition on the anchor observation $\mathbf x=x$ and write $z=f_\theta(x)$. Since the negatives $\{\mathbf y_j'\}_{j=1}^N$ are i.i.d.\ from $p_{\mathbf y}$, the representations $\{\mathbf w_j'\}_{j=1}^N$ are i.i.d.\ with law $q_\phi$. Hence, by the same concentration argument as in Step~2 of Part~I of \Cref{thm:nce_consistency},
\[
\frac{1}{N}
\mathsf S_N^{\theta\shortto\phi}(z)
=
\Gamma_{\phi,\tau}(z)
+
O_p(N^{-1/2}),
\]
which implies, by the same logarithmic expansion used in the unimodal proof,
\[
\log
\mathsf S_N^{\theta\shortto\phi}(z)
=
\log N
+
\log\Gamma_{\phi,\tau}(z)
+
O_p(N^{-1/2}).
\]
Returning to the random anchor $\mathbf z$, combining the above displays yields
\begin{equation}
\label{eq:mm_loss_asymp_theta_to_phi}
\ell_{\mathcal B}^{x\shortto y}(\theta,\phi)
=
-\frac{1}{\tau}
\mathrm{sim}(\mathbf z,\mathbf w)
+
\log N
+
\log\Gamma_{\phi,\tau}(\mathbf z)
+
O_p(N^{-1/2}).
\end{equation}

\textbf{Step 2: Identification of the directional energy.}
Taking expectations of \Cref{eq:mm_loss_asymp_theta_to_phi} over the batch construction gives
\[
\mathcal L_{\mathrm{NCE},N}^{x\shortto y}(\theta,\phi)
:=
\mathbb E_{\mathcal B}
\left[
\ell_{\mathcal B}^{x\shortto y}(\theta,\phi)
\right]
=
\mathbb E
\left[
-\frac{1}{\tau}
\mathrm{sim}(\mathbf z,\mathbf w)
\right]
+
\mathbb E
\left[
\log\Gamma_{\phi,\tau}(\mathbf z)
\right]
+
\log N
+
o(1).
\]
Here, for fixed $\tau>0$, the remainder is uniformly bounded and hence uniformly integrable. Since it converges to zero in probability, it also converges to zero in $L^1$, so its expectation is $o(1)$.

Using the disintegration in \Cref{def:directional_potential}, the law of iterated expectations yields
\[
\mathbb E_{(\mathbf x,\mathbf y)\sim r_{\mathrm{mm}}}
\left[
-\frac{1}{\tau}
\mathrm{sim}
\big(
f_\theta(\mathbf x),
g_\phi(\mathbf y)
\big)
\right]
=
\frac{1}{\tau}
\mathbb E_{\mathbf z\sim q_\theta}
\left[
U_{\theta\shortto\phi}(\mathbf z)
\right]
=
\frac{1}{\tau}
\int_{\mathcal Z}
U_{\theta\shortto\phi}(z)\,
q_\theta(\mathrm dz).
\]
Since $\mathbf z=f_\theta(\mathbf x)\sim q_\theta$,
\[
\mathbb E
\left[
\log\Gamma_{\phi,\tau}(\mathbf z)
\right]
=
\mathbb E_{\mathbf z\sim q_\theta}
\left[
\log\Gamma_{\phi,\tau}(\mathbf z)
\right].
\]
Under the constant kernel-volume condition \Cref{ass:kernel_volume},
\[
\tilde\rho_{\phi,\tau}(z)
=
\frac{
\Gamma_{\phi,\tau}(z)
}{
V_\kappa(\tau)
}
\]
by \Cref{def:smoothed_density}, and hence
\begin{align*}
\mathbb E_{\mathbf z\sim q_\theta}
\left[
\log\Gamma_{\phi,\tau}(\mathbf z)
\right]
&=
\log V_\kappa(\tau)
+
\mathbb E_{\mathbf z\sim q_\theta}
\left[
\log\tilde\rho_{\phi,\tau}(\mathbf z)
\right]
\\
&=
\log V_\kappa(\tau)
-
H_\times
\big(
q_\theta,
\tilde\rho_{\phi,\tau}
\big).
\end{align*}
Substituting into the preceding expansion gives
\begin{align*}
\mathcal L_{\mathrm{NCE},N}^{x\shortto y}(\theta,\phi)
&=
\frac{1}{\tau}
\int_{\mathcal Z}
U_{\theta\shortto\phi}(z)\,
q_\theta(\mathrm dz)
-
H_\times
\big(
q_\theta,
\tilde\rho_{\phi,\tau}
\big)
+
\log\!\big(
NV_\kappa(\tau)
\big)
+
o(1)
\\
&=
\mathcal J_\tau^{x\shortto y}(\theta,\phi)
+
\log\!\big(
NV_\kappa(\tau)
\big)
+
o(1).
\end{align*}
The reverse direction $y\shortto x$ follows analogously, yielding
\[
\mathcal L_{\mathrm{NCE},N}^{y\shortto x}(\phi,\theta)
=
\mathcal J_\tau^{y\shortto x}(\phi,\theta)
+
\log\!\big(
NV_\kappa(\tau)
\big)
+
o(1).
\]
Averaging the two directional expansions and using \Cref{eq:sym_gef_def} gives
\[
\left|
\mathcal L_{\mathrm{Sym},N}(\theta,\phi)
-
\mathcal J_\tau^{\mathrm{Sym}}(\theta,\phi)
-
\log\!\big(
NV_\kappa(\tau)
\big)
\right|
\longrightarrow0
\]
as $N\to\infty$, for each fixed $\tau>0$ and $(\theta,\phi)\in\Theta\times\Phi$.

\paragraph{Part II. Gradient consistency.}
Fix any $\tau>0$ and $(\theta,\phi)\in\Theta\times\Phi$. Differentiating the finite-batch loss in \Cref{eq:mm_batch_loss} gives
\begin{equation}
\label{eq:mm_batch_grad}
\nabla
\ell_{\mathcal B}^{\mathrm{Sym}}(\theta,\phi)
=
-\frac{1}{\tau}
\nabla
\mathrm{sim}(\mathbf z,\mathbf w)
+
\frac12
\nabla
\log
\mathsf Z_{\mathcal B}^{\theta\shortto\phi}(\mathbf z)
+
\frac12
\nabla
\log
\mathsf Z_{\mathcal B}^{\phi\shortto\theta}(\mathbf w),
\end{equation}
where $\nabla:=\nabla_{(\theta,\phi)}$ denotes the joint gradient.

\textbf{Step 1: Removing the positive term in each log-partition gradient.}
Define
\[
\mathsf g_N^{\theta\shortto\phi}
:=
\nabla
\log
\left(
\mathsf P+
\mathsf S_N^{\theta\shortto\phi}
\right),
\qquad
\mathsf r_N^{\theta\shortto\phi}
:=
\nabla
\log
\mathsf S_N^{\theta\shortto\phi},
\]
and analogously define $(\mathsf g_N^{\phi\shortto\theta},\mathsf r_N^{\phi\shortto\theta})$. As in \Cref{eq:pos_removal_identity}, a direct algebraic manipulation yields
\[
\mathsf g_N^{\theta\shortto\phi}
-
\mathsf r_N^{\theta\shortto\phi}
=
\frac{
\nabla\mathsf P
}{
\mathsf P+\mathsf S_N^{\theta\shortto\phi}
}
-
\frac{
\mathsf P
}{
\mathsf S_N^{\theta\shortto\phi}
\big(
\mathsf P+\mathsf S_N^{\theta\shortto\phi}
\big)
}
\nabla
\mathsf S_N^{\theta\shortto\phi},
\]
and the same identity holds with $\theta\shortto\phi$ replaced by $\phi\shortto\theta$.

By \Cref{assump:mm_regularity}, $\mathcal Z$ is compact and $\mathrm{sim}$ is $C^1$ with bounded input gradients. Together with the $C^1$ encoders with uniformly bounded Jacobians, this implies that, for each fixed $\tau>0$, there exist finite constants
\[
0
<
m_\tau
\le
\kappa_\tau(\cdot,\cdot)
\le
M_\tau
<
\infty,
\qquad
\sup_{(\theta,\phi)\in\Theta\times\Phi}
\sup_{\mathcal B}
\|
\nabla\kappa_\tau
\|
\le
G_\tau
<
\infty.
\]
Consequently, uniformly over $(\theta,\phi)$ and batch realizations,
\[
\mathsf P
\le
M_\tau,
\qquad
\|
\nabla\mathsf P
\|
\le
G_\tau,
\qquad
\mathsf S_N^{\theta\shortto\phi}
\ge
Nm_\tau,
\qquad
\|
\nabla
\mathsf S_N^{\theta\shortto\phi}
\|
\le
NG_\tau,
\]
and the same bounds hold for $\mathsf S_N^{\phi\shortto\theta}$. Taking norms in the identity above gives
\begin{equation}
\label{eq:mm_pos_removed}
\left\|
\nabla
\log
\mathsf Z_{\mathcal B}^{\theta\shortto\phi}(\mathbf z)
-
\nabla
\log
\mathsf S_N^{\theta\shortto\phi}(\mathbf z)
\right\|
=
O_\tau(N^{-1}),
\qquad
\left\|
\nabla
\log
\mathsf Z_{\mathcal B}^{\phi\shortto\theta}(\mathbf w)
-
\nabla
\log
\mathsf S_N^{\phi\shortto\theta}(\mathbf w)
\right\|
=
O_\tau(N^{-1}).
\end{equation}

\textbf{Step 2: Consistency of the negative-sum ratios.}
We treat the direction $\theta\shortto\phi$; the reverse direction is analogous. Condition on the anchor observations $\mathbf x=x$ and $\mathbf y=y$, and write
\[
z=f_\theta(x),
\qquad
w=g_\phi(y).
\]
Then
\[
\nabla
\log
\mathsf S_N^{\theta\shortto\phi}(z)
=
\frac{
\frac1N
\sum_{j=1}^N
\nabla
\kappa_\tau
\big(
f_\theta(x),
g_\phi(\mathbf y_j')
\big)
}{
\frac1N
\sum_{j=1}^N
\kappa_\tau
\big(
f_\theta(x),
g_\phi(\mathbf y_j')
\big)
}.
\]
Given $x$, the negatives $\{\mathbf y_j'\}_{j=1}^N$ are i.i.d.\ with law $p_{\mathbf y}$. Define
\[
\Gamma_{\phi,\tau}
\big(
f_\theta(x)
\big)
=
\mathbb E_{\mathbf y'\sim p_{\mathbf y}}
\left[
\kappa_\tau
\big(
f_\theta(x),
g_\phi(\mathbf y')
\big)
\right].
\]
Since $\Gamma_{\phi,\tau}(f_\theta(x))\ge m_\tau>0$, the denominator is bounded away from zero at the population level. By the law of large numbers and boundedness of $\kappa_\tau$,
\[
\frac1N
\sum_{j=1}^N
\kappa_\tau
\big(
f_\theta(x),
g_\phi(\mathbf y_j')
\big)
\xrightarrow{p}
\Gamma_{\phi,\tau}
\big(
f_\theta(x)
\big).
\]
Similarly, since the total parameter gradient of $\kappa_\tau$ is uniformly bounded,
\[
\frac1N
\sum_{j=1}^N
\nabla
\kappa_\tau
\big(
f_\theta(x),
g_\phi(\mathbf y_j')
\big)
\xrightarrow{p}
\mathbb E_{\mathbf y'\sim p_{\mathbf y}}
\left[
\nabla
\kappa_\tau
\big(
f_\theta(x),
g_\phi(\mathbf y')
\big)
\right].
\]
Since the expectation is taken with respect to the parameter-free distribution $p_{\mathbf y}$ and the total parameter gradient is uniformly bounded by \Cref{assump:mm_regularity}, differentiation under the expectation is justified. Hence
\[
\mathbb E_{\mathbf y'\sim p_{\mathbf y}}
\left[
\nabla
\kappa_\tau
\big(
f_\theta(x),
g_\phi(\mathbf y')
\big)
\right]
=
\nabla
\mathbb E_{\mathbf y'\sim p_{\mathbf y}}
\left[
\kappa_\tau
\big(
f_\theta(x),
g_\phi(\mathbf y')
\big)
\right]
=
\nabla
\Gamma_{\phi,\tau}
\big(
f_\theta(x)
\big),
\]
where the derivative on the right is the total derivative with respect to $(\theta,\phi)$, including the dependence of the anchor representation $f_\theta(x)$ on $\theta$. The continuous mapping theorem therefore yields
\begin{equation}
\label{eq:mm_ratio_consistency_1}
\nabla
\log
\mathsf S_N^{\theta\shortto\phi}(z)
\xrightarrow{p}
\frac{
\nabla
\Gamma_{\phi,\tau}
\big(
f_\theta(x)
\big)
}{
\Gamma_{\phi,\tau}
\big(
f_\theta(x)
\big)
}
=
\nabla
\log
\Gamma_{\phi,\tau}
\big(
f_\theta(x)
\big).
\end{equation}
The reverse direction follows identically. Conditioning on $\mathbf x=x$ and $\mathbf y=y$ and writing $w=g_\phi(y)$, define
\[
\Gamma_{\theta,\tau}
\big(
g_\phi(y)
\big)
:=
\mathbb E_{\mathbf x'\sim p_{\mathbf x}}
\left[
\kappa_\tau
\big(
f_\theta(\mathbf x'),
g_\phi(y)
\big)
\right].
\]
Then
\begin{equation}
\label{eq:mm_ratio_consistency_2}
\nabla
\log
\mathsf S_N^{\phi\shortto\theta}(w)
\xrightarrow{p}
\nabla
\log
\Gamma_{\theta,\tau}
\big(
g_\phi(y)
\big).
\end{equation}
Returning to the random anchors and combining \Cref{eq:mm_pos_removed,eq:mm_ratio_consistency_1,eq:mm_ratio_consistency_2}, as $N\to\infty$ we obtain
\begin{equation}
\label{eq:mm_logZ_grad_limit}
\nabla
\log
\mathsf Z_{\mathcal B}^{\theta\shortto\phi}(\mathbf z)
\xrightarrow{p}
\nabla
\log
\Gamma_{\phi,\tau}(\mathbf z),
\qquad
\nabla
\log
\mathsf Z_{\mathcal B}^{\phi\shortto\theta}(\mathbf w)
\xrightarrow{p}
\nabla
\log
\Gamma_{\theta,\tau}(\mathbf w).
\end{equation}

\textbf{Step 3: Passing to the population objective and identifying $\nabla_{(\theta,\phi)}\mathcal J_\tau^{\mathrm{Sym}}$.}
By \Cref{lem:sym_grad_bound}, $\|\nabla\ell_{\mathcal B}^{\mathrm{Sym}}(\theta,\phi)\|$ is uniformly bounded for fixed $\tau>0$ and hence uniformly integrable. Together with the convergence in probability in \Cref{eq:mm_logZ_grad_limit}, this implies convergence in $L^1$, so we may pass to expectations and obtain
\begin{equation}
\label{eq:mm_grad_limit_pop}
\begin{aligned}
\nabla
\mathcal L_{\mathrm{Sym},N}(\theta,\phi)
=
\mathbb E_{\mathcal B}
\left[
\nabla
\ell_{\mathcal B}^{\mathrm{Sym}}(\theta,\phi)
\right]
\longrightarrow\;&
-\frac{1}{\tau}
\mathbb E_{(\mathbf x,\mathbf y)\sim r_{\mathrm{mm}}}
\left[
\nabla
\mathrm{sim}
\big(
f_\theta(\mathbf x),
g_\phi(\mathbf y)
\big)
\right]
\\
&+
\frac12
\mathbb E_{\mathbf x\sim p_{\mathbf x}}
\left[
\nabla
\log
\Gamma_{\phi,\tau}
\big(
f_\theta(\mathbf x)
\big)
\right]
\\
&+
\frac12
\mathbb E_{\mathbf y\sim p_{\mathbf y}}
\left[
\nabla
\log
\Gamma_{\theta,\tau}
\big(
g_\phi(\mathbf y)
\big)
\right].
\end{aligned}
\end{equation}

We now identify the right-hand side as $\nabla\mathcal J_\tau^{\mathrm{Sym}}(\theta,\phi)$. First, by the definition of $\pi_{\theta\phi}=(f_\theta\times g_\phi)_\#r_{\mathrm{mm}}$ and the law of iterated expectations applied to \Cref{def:directional_potential},
\[
\int_{\mathcal Z}
U_{\theta\shortto\phi}(z)\,
q_\theta(\mathrm dz)
=
\mathbb E_{(\mathbf x,\mathbf y)\sim r_{\mathrm{mm}}}
\left[
-\mathrm{sim}
\big(
f_\theta(\mathbf x),
g_\phi(\mathbf y)
\big)
\right]
=
\mathbb E_{(\mathbf z,\mathbf w)\sim\pi_{\theta\phi}}
\left[
-\mathrm{sim}(\mathbf z,\mathbf w)
\right],
\]
and similarly,
\[
\int_{\mathcal Z}
U_{\phi\shortto\theta}(w)\,
q_\phi(\mathrm dw)
=
\mathbb E_{(\mathbf x,\mathbf y)\sim r_{\mathrm{mm}}}
\left[
-\mathrm{sim}
\big(
f_\theta(\mathbf x),
g_\phi(\mathbf y)
\big)
\right].
\]
While these two binding energies coincide after integration, the directional potential fields $U_{\theta\shortto\phi}$ and $U_{\phi\shortto\theta}$ generally differ pointwise. Under \Cref{assump:mm_regularity}, differentiation under the parameter-free expectation is justified, yielding
\[
\nabla
\int_{\mathcal Z}
U_{\theta\shortto\phi}(z)\,
q_\theta(\mathrm dz)
=
\nabla
\int_{\mathcal Z}
U_{\phi\shortto\theta}(w)\,
q_\phi(\mathrm dw)
=
-
\mathbb E_{(\mathbf x,\mathbf y)\sim r_{\mathrm{mm}}}
\left[
\nabla
\mathrm{sim}
\big(
f_\theta(\mathbf x),
g_\phi(\mathbf y)
\big)
\right].
\]

Second, using
\[
\log\tilde\rho_{\phi,\tau}(z)
=
\log\Gamma_{\phi,\tau}(z)
-
\log V_\kappa(\tau),
\qquad
\log\tilde\rho_{\theta,\tau}(w)
=
\log\Gamma_{\theta,\tau}(w)
-
\log V_\kappa(\tau),
\]
with $V_\kappa(\tau)$ independent of $(\theta,\phi)$ by \Cref{ass:kernel_volume}, we may write
\begin{align*}
\nabla
H_\times
\big(
q_\theta,
\tilde\rho_{\phi,\tau}
\big)
&=
-
\nabla
\mathbb E_{\mathbf x\sim p_{\mathbf x}}
\left[
\log
\tilde\rho_{\phi,\tau}
\big(
f_\theta(\mathbf x)
\big)
\right]
=
-
\mathbb E_{\mathbf x\sim p_{\mathbf x}}
\left[
\nabla
\log
\Gamma_{\phi,\tau}
\big(
f_\theta(\mathbf x)
\big)
\right],
\\
\nabla
H_\times
\big(
q_\phi,
\tilde\rho_{\theta,\tau}
\big)
&=
-
\nabla
\mathbb E_{\mathbf y\sim p_{\mathbf y}}
\left[
\log
\tilde\rho_{\theta,\tau}
\big(
g_\phi(\mathbf y)
\big)
\right]
=
-
\mathbb E_{\mathbf y\sim p_{\mathbf y}}
\left[
\nabla
\log
\Gamma_{\theta,\tau}
\big(
g_\phi(\mathbf y)
\big)
\right],
\end{align*}
where differentiation under the expectations is justified by the uniform gradient bounds in \Cref{assump:mm_regularity,lem:sym_grad_bound}.

Combining these identities with the definition of $\mathcal J_\tau^{\mathrm{Sym}}$ in \Cref{def:sym_gef} shows that the limit in \Cref{eq:mm_grad_limit_pop} equals $\nabla\mathcal J_\tau^{\mathrm{Sym}}(\theta,\phi)$. Therefore, as $N\to\infty$, for each fixed $\tau>0$ and $(\theta,\phi)\in\Theta\times\Phi$, the two parameter components satisfy
\[
\left\|
\nabla_\theta
\mathcal L_{\mathrm{Sym},N}(\theta,\phi)
-
\nabla_\theta
\mathcal J_\tau^{\mathrm{Sym}}(\theta,\phi)
\right\|
\longrightarrow0,
\qquad
\left\|
\nabla_\phi
\mathcal L_{\mathrm{Sym},N}(\theta,\phi)
-
\nabla_\phi
\mathcal J_\tau^{\mathrm{Sym}}(\theta,\phi)
\right\|
\longrightarrow0.
\]
Consequently, for the joint gradient $\nabla_{(\theta,\phi)}=(\nabla_\theta,\nabla_\phi)$,
\[
\begin{aligned}
&\Big\|
\nabla_{(\theta,\phi)}
\mathcal L_{\mathrm{Sym},N}(\theta,\phi)
-
\nabla_{(\theta,\phi)}
\mathcal J_\tau^{\mathrm{Sym}}(\theta,\phi)
\Big\|\\
&=
\sqrt{
\left\|
\nabla_\theta
\mathcal L_{\mathrm{Sym},N}(\theta,\phi)
-
\nabla_\theta
\mathcal J_\tau^{\mathrm{Sym}}(\theta,\phi)
\right\|^2
+
\left\|
\nabla_\phi
\mathcal L_{\mathrm{Sym},N}(\theta,\phi)
-
\nabla_\phi
\mathcal J_\tau^{\mathrm{Sym}}(\theta,\phi)
\right\|^2
}
\longrightarrow0.
\end{aligned}
\]
\end{proof}

\subsection{Explicit Extremal Construction and Proof of \texorpdfstring{\Cref{prop:mm_coordinate_extremal}}{Prop. 4.1}}
\label{app:mm_coordinate_extremal}

We now give the extremal-density construction underlying \Cref{prop:mm_coordinate_extremal} and the full proof of the proposition.

\begin{restatable}[Constructive form of the coordinate infimum and extremal convergence]{proposition}{propmmcoordinateextremalextend}
\label{prop:mm_coordinate_extremal_extended}
Assume $\mathcal Z$ is compact with $\mu(\mathcal Z)<\infty$, and let $\mathcal F^{\mathrm{Sym}}_{\tau,\mathbf U_{1,2}}$ be as in \Cref{def:mm_free_energy}. Fix $\rho_2\in\mathcal D_{\mu,\underline\rho_2}(\mathcal Z)$ and define the \emph{effective potential field} by
\[
V_{1\mid2}(z)
:=
\frac{1}{\tau}U_{1\shortto2}(z)+\log\rho_2(z).
\]
Then, the coordinate map $\rho_1\mapsto\mathcal F^{\mathrm{Sym}}_{\tau,\mathbf U_{1,2}}(\rho_1,\rho_2)$ is concave on $\mathcal D_{\mu,\underline\rho_1}(\mathcal Z)$, and its infimum is approached by sequences that concentrate the \emph{excess mass} $M_{\mathrm{ex}}^{(1)}:=1-\underline\rho_1\mu(\mathcal Z)$ onto approximate minimizers of $V_{1\mid2}$, leaving only the floor $\underline\rho_1$ elsewhere.

Formally, for $\sigma>0$, let
\[
\mathcal W_1^\sigma
:=
\left\{
z\in\mathcal Z:
V_{1\mid2}(z)
\le
\operatorname*{ess\,inf}\nolimits_{\mu}V_{1\mid2}
+
\sigma
\right\}.
\]
Choose measurable sets $\mathcal S_1^\sigma\subseteq\mathcal W_1^\sigma$ with $0<\mu(\mathcal S_1^\sigma)<\infty$ and $\mu(\mathcal S_1^\sigma)\to0$ as $\sigma\to0$, and define the \emph{$\sigma$-approximate density}
\begin{equation}
\label{eq:sigma_approx_minimizer}
\rho_1^{(\sigma)}(z)
=
\underline\rho_1
+
\mathbbm{1}_{\mathcal S_1^\sigma}(z)
\frac{M_{\mathrm{ex}}^{(1)}}{\mu(\mathcal S_1^\sigma)}.
\end{equation}
Then, as $\sigma\to0$, the energy of $\rho_1^{(\sigma)}$ converges to the coordinate infimum:
\begin{equation}
\label{eq:extremal_collapse}
\lim_{\sigma\to0}
\mathcal F^{\mathrm{Sym}}_{\tau,\mathbf U_{1,2}}
\big(
\rho_1^{(\sigma)},
\rho_2
\big)
=
\inf_{\rho_1\in\mathcal D_{\mu,\underline\rho_1}(\mathcal Z)}
\mathcal F^{\mathrm{Sym}}_{\tau,\mathbf U_{1,2}}
\big(
\rho_1,
\rho_2
\big).
\end{equation}
An analogous statement holds when optimizing in $\rho_2$ with $\rho_1$ fixed, with respect to the dual effective potential field
\(
V_{2\mid1}(z)
:=
\frac{1}{\tau}U_{2\shortto1}(z)+\log\rho_1(z).
\)
\end{restatable}

\begin{proofintuition}
This proof establishes that when one modality's distribution ($\rho_2$) is held fixed, an infimizing sequence for the other modality ($\rho_1$) approaches an extreme, ``spiky'' configuration. Rather than spreading out, it approaches the minimum allowed density everywhere while concentrating its remaining probability mass near the lowest regions of an effective potential landscape. The logic unfolds in three geometric phases: \textbf{(i)} Unlike the unimodal free energy, the cross-entropy term makes the functional \emph{concave} with respect to $\rho_1$. This motivates extremal configurations near the boundary of the feasible density class. We derive a lower bound corresponding to a background floor $\underline\rho_1$ together with all excess probability mass placed near the essential minimum of the effective potential $V_{1\mid2}$. \textbf{(ii)} We then construct a sequence of feasible densities whose excess mass is concentrated on shrinking subsets of the corresponding low-effective-potential regions. \textbf{(iii)} Although this concentration makes the spike increasingly tall, the peer density is bounded, so the additional logarithmic term scales as $\mu(\mathcal S_1^\sigma)\log(1/\mu(\mathcal S_1^\sigma))$ and therefore vanishes as the support shrinks. The resulting sequence approaches the coordinate infimum while concentrating its excess mass near minimizers of the effective potential.
\end{proofintuition}

\begin{proof}
Fix $\tau>0$ and $\rho_2\in\mathcal D_{\mu,\underline\rho_2}(\mathcal Z)$. Throughout, we treat $\rho_2$ as fixed and analyze $\rho_1$; the argument for $\rho_2$ is symmetric. Write the coordinate map as
\[
\mathcal F_1(\rho_1)
:=
\mathcal F^{\mathrm{Sym}}_{\tau,\mathbf U_{1,2}}(\rho_1,\rho_2),
\qquad
\rho_1\in\mathcal D_{\mu,\underline\rho_1}(\mathcal Z).
\]

\textbf{Step 1: Concavity of $\rho_1\mapsto\mathcal F_1$ and its lower bound.}
Using
\[
H(\rho_1)
=
-\int_{\mathcal Z}
\rho_1(z)\log\rho_1(z)\,
\mu(\mathrm dz),
\]
\[
D_{\mathrm{KL}}(\rho_1\|\rho_2)
=
\int_{\mathcal Z}
\rho_1(z)
\log\frac{\rho_1(z)}{\rho_2(z)}
\,
\mu(\mathrm dz),
\]
and
\[
D_{\mathrm{KL}}(\rho_2\|\rho_1)
=
\int_{\mathcal Z}
\rho_2(z)
\log\frac{\rho_2(z)}{\rho_1(z)}
\,
\mu(\mathrm dz),
\]
we expand the $\rho_1$-dependent terms in $\mathcal F_1(\rho_1)$:
\begin{equation}
\label{eq:f1_form}
\mathcal F_1(\rho_1)
=
\frac{1}{2}
\int_{\mathcal Z}
\rho_1(z)V_{1\mid2}(z)\,
\mu(\mathrm dz)
+
\frac{1}{2}
\int_{\mathcal Z}
\rho_2(z)\log\rho_1(z)\,
\mu(\mathrm dz)
+
C(\rho_2),
\end{equation}
where
\[
V_{1\mid2}(z)
:=
\frac{1}{\tau}U_{1\shortto2}(z)
+
\log\rho_2(z),
\]
and $C(\rho_2)$ does not depend on $\rho_1$.

The map $\rho_1\mapsto\int_{\mathcal Z}\rho_1(z)V_{1\mid2}(z)\,\mu(\mathrm dz)$ is linear, while $\rho_1\mapsto\int_{\mathcal Z}\rho_2(z)\log\rho_1(z)\,\mu(\mathrm dz)$ is concave on the positive cone since $\log$ is concave and $\rho_2\ge\underline\rho_2>0$ $\mu$-a.e. Thus, $\rho_1\mapsto\mathcal F_1(\rho_1)$ is concave on the convex set $\mathcal D_{\mu,\underline\rho_1}(\mathcal Z)$. This concave coordinate geometry motivates the boundary-concentrating construction below.

Write any feasible $\rho_1\in\mathcal D_{\mu,\underline\rho_1}(\mathcal Z)$ as
\[
\rho_1(z)
=
\underline\rho_1+\eta_1(z),
\qquad
\eta_1(z)\ge0.
\]
The excess probability mass is
\[
M_{\mathrm{ex}}^{(1)}
=
\int_{\mathcal Z}
\eta_1(z)\,
\mu(\mathrm dz)
=
1-\underline\rho_1\mu(\mathcal Z)
>
0.
\]
Let
\[
v_*
:=
\operatorname*{ess\,inf}_{\mu}
V_{1\mid2}.
\]
Since $V_{1\mid2}(z)\ge v_*$ and $\log\rho_1(z)\ge\log\underline\rho_1$ $\mu$-a.e., we have
\begin{align*}
\int_{\mathcal Z}
\rho_1(z)V_{1\mid2}(z)\,
\mu(\mathrm dz)
&=
\underline\rho_1
\int_{\mathcal Z}
V_{1\mid2}(z)\,
\mu(\mathrm dz)
+
\int_{\mathcal Z}
\eta_1(z)V_{1\mid2}(z)\,
\mu(\mathrm dz)
\\
&\ge
\underline\rho_1
\int_{\mathcal Z}
V_{1\mid2}(z)\,
\mu(\mathrm dz)
+
v_*M_{\mathrm{ex}}^{(1)},
\end{align*}
and
\[
\int_{\mathcal Z}
\rho_2(z)\log\rho_1(z)\,
\mu(\mathrm dz)
\ge
\int_{\mathcal Z}
\rho_2(z)\log\underline\rho_1\,
\mu(\mathrm dz)
=
\log\underline\rho_1.
\]
Plugging these bounds into \Cref{eq:f1_form} yields
\begin{equation}
\label{eq:f1_lowerbound}
\mathcal F_1(\rho_1)
\ge
\frac12
\left(
\underline\rho_1
\int_{\mathcal Z}
V_{1\mid2}(z)\,
\mu(\mathrm dz)
+
v_*M_{\mathrm{ex}}^{(1)}
\right)
+
\frac12\log\underline\rho_1
+
C(\rho_2)
=:
\mathcal F_*.
\end{equation}
Therefore,
\[
\inf_{\rho_1\in\mathcal D_{\mu,\underline\rho_1}(\mathcal Z)}
\mathcal F_1(\rho_1)
\ge
\mathcal F_*.
\]

\textbf{Step 2: Evaluate the coordinate map $\mathcal F_1$ at $\rho_1^{(\sigma)}$.}
For each $\sigma>0$, define the sublevel set
\[
\mathcal W_1^\sigma
:=
\left\{
z\in\mathcal Z:
V_{1\mid2}(z)
\le
v_*+\sigma
\right\}.
\]
By the definition of the essential infimum, $\mu(\mathcal W_1^\sigma)>0$. Since $\mu$ is a non-atomic Riemannian volume measure, we may choose measurable subsets $\mathcal S_1^\sigma\subseteq\mathcal W_1^\sigma$ such that $0<\mu(\mathcal S_1^\sigma)<\infty$ and $\mu(\mathcal S_1^\sigma)\to0$ as $\sigma\to0$. Define
\[
\rho_1^{(\sigma)}(z)
=
\underline\rho_1
+
\mathbbm{1}_{\mathcal S_1^\sigma}(z)
\frac{M_{\mathrm{ex}}^{(1)}}{\mu(\mathcal S_1^\sigma)}.
\]
Since $\rho_1^{(\sigma)}\ge\underline\rho_1$ and
\[
\int_{\mathcal Z}
\rho_1^{(\sigma)}(z)\,
\mu(\mathrm dz)
=
\underline\rho_1\mu(\mathcal Z)
+
\frac{M_{\mathrm{ex}}^{(1)}}{\mu(\mathcal S_1^\sigma)}
\mu(\mathcal S_1^\sigma)
=
\underline\rho_1\mu(\mathcal Z)
+
M_{\mathrm{ex}}^{(1)}
=
1,
\]
we have $\rho_1^{(\sigma)}\in\mathcal D_{\mu,\underline\rho_1}(\mathcal Z)$.

We now evaluate the coordinate map $\mathcal F_1$ at $\rho_1^{(\sigma)}$.

\emph{(i) Linear term.}
Since $\mathcal S_1^\sigma\subseteq\mathcal W_1^\sigma$, we have $V_{1\mid2}(z)\le v_*+\sigma$ on $\mathcal S_1^\sigma$, and hence
\begin{align*}
\int_{\mathcal Z}
\rho_1^{(\sigma)}(z)V_{1\mid2}(z)\,
\mu(\mathrm dz)
&=
\underline\rho_1
\int_{\mathcal Z}
V_{1\mid2}(z)\,
\mu(\mathrm dz)
+
\frac{M_{\mathrm{ex}}^{(1)}}{\mu(\mathcal S_1^\sigma)}
\int_{\mathcal S_1^\sigma}
V_{1\mid2}(z)\,
\mu(\mathrm dz)
\\
&\le
\underline\rho_1
\int_{\mathcal Z}
V_{1\mid2}(z)\,
\mu(\mathrm dz)
+
M_{\mathrm{ex}}^{(1)}
(v_*+\sigma).
\end{align*}

\emph{(ii) Log term.}
Split the domain as $\mathcal Z=(\mathcal Z\setminus\mathcal S_1^\sigma)\cup\mathcal S_1^\sigma$. Then
\begin{align*}
\int_{\mathcal Z}
\rho_2(z)\log\rho_1^{(\sigma)}(z)\,
\mu(\mathrm dz)
\;=\;
\int_{\mathcal Z\setminus\mathcal S_1^\sigma}
\rho_2(z)\log\underline\rho_1\,
\mu(\mathrm dz)
+
\int_{\mathcal S_1^\sigma}
\rho_2(z)
\log\!\left(
\underline\rho_1
+
\frac{M_{\mathrm{ex}}^{(1)}}{\mu(\mathcal S_1^\sigma)}
\right)
\mu(\mathrm dz).
\end{align*}
Define
\[
m_\sigma
:=
\int_{\mathcal S_1^\sigma}
\rho_2(z)\,
\mu(\mathrm dz).
\]
Since $\int_{\mathcal Z}\rho_2(z)\,\mu(\mathrm dz)=1$, the first term equals $(1-m_\sigma)\log\underline\rho_1$. Therefore,
\[
\int_{\mathcal Z}
\rho_2(z)\log\rho_1^{(\sigma)}(z)\,
\mu(\mathrm dz)
=
\log\underline\rho_1
+
m_\sigma L_\sigma,
\]
where
\[
L_\sigma
:=
\log\!\left(
1+
\frac{M_{\mathrm{ex}}^{(1)}}
{\underline\rho_1\mu(\mathcal S_1^\sigma)}
\right).
\]

Combining the two terms gives
\begin{equation}
\label{eq:f1_evaluate}
\mathcal F_1(\rho_1^{(\sigma)})
\le
\frac12
\left(
\underline\rho_1
\int_{\mathcal Z}
V_{1\mid2}(z)\,
\mu(\mathrm dz)
+
M_{\mathrm{ex}}^{(1)}(v_*+\sigma)
\right)
+
\frac12
\left(
\log\underline\rho_1
+
m_\sigma L_\sigma
\right)
+
C(\rho_2).
\end{equation}

\textbf{Step 3: Conclude convergence.}
We first show that the additional log-penalty $m_\sigma L_\sigma$ vanishes as $\sigma\to0$. Since $\rho_2\in\mathcal D_{\mu,\underline\rho_2}(\mathcal Z)\subset L^\infty(\mu)$,
\[
m_\sigma
=
\int_{\mathcal S_1^\sigma}
\rho_2(z)\,
\mu(\mathrm dz)
\le
\|\rho_2\|_\infty
\mu(\mathcal S_1^\sigma).
\]
Hence
\[
0
\le
m_\sigma L_\sigma
\le
\|\rho_2\|_\infty
\mu(\mathcal S_1^\sigma)
\log\!\left(
1+
\frac{M_{\mathrm{ex}}^{(1)}}
{\underline\rho_1\mu(\mathcal S_1^\sigma)}
\right).
\]
Since $\mu(\mathcal S_1^\sigma)\to0$ and $t\log(1+c/t)\to0$ as $t\downarrow0$ for every fixed $c>0$, it follows that
\[
m_\sigma L_\sigma
\longrightarrow0
\qquad
\text{as }\sigma\to0.
\]
Now subtract $\mathcal F_*$ from \Cref{eq:f1_evaluate} to obtain
\[
\mathcal F_1(\rho_1^{(\sigma)})
-
\mathcal F_*
\le
\frac12
\left(
M_{\mathrm{ex}}^{(1)}\sigma
+
m_\sigma L_\sigma
\right).
\]
Letting $\sigma\to0$ gives
\[
\limsup_{\sigma\to0}
\mathcal F_1(\rho_1^{(\sigma)})
\le
\mathcal F_*.
\]
Meanwhile, Step~1 showed
\[
\inf_{\rho_1\in\mathcal D_{\mu,\underline\rho_1}(\mathcal Z)}
\mathcal F_1(\rho_1)
\ge
\mathcal F_*.
\]
Since every $\rho_1^{(\sigma)}$ is feasible, we also have
\[
\inf_{\rho_1\in\mathcal D_{\mu,\underline\rho_1}(\mathcal Z)}
\mathcal F_1(\rho_1)
\le
\mathcal F_1(\rho_1^{(\sigma)})
\]
for every $\sigma>0$. Therefore,
\[
\inf_{\rho_1\in\mathcal D_{\mu,\underline\rho_1}(\mathcal Z)}
\mathcal F_1(\rho_1)
=
\mathcal F_*,
\qquad
\lim_{\sigma\to0}
\mathcal F_1(\rho_1^{(\sigma)})
=
\inf_{\rho_1\in\mathcal D_{\mu,\underline\rho_1}(\mathcal Z)}
\mathcal F_1(\rho_1).
\]
Thus, the coordinate infimum is approached by boundary-concentrating infimizing sequences whose excess mass lies in shrinking low-effective-potential regions of
\[
V_{1\mid2}
=
\frac{1}{\tau}U_{1\shortto2}
+
\log\rho_2.
\]
The $\rho_2$-coordinate statement follows identically after swapping indices and using
\[
V_{2\mid1}
=
\frac{1}{\tau}U_{2\shortto1}
+
\log\rho_1.
\]
\end{proof}

\subsection{Proof of \texorpdfstring{\Cref{thm:intrinsic_multimodal}}{Thm. 4.2}}
\label{app:intrinsic_multimodal}

\thmintrinsicmultimodal*

\begin{proofintuition}
This proof establishes that, in the low-temperature limit, the cross-modal parametric energy approaches the intrinsic dual-density free energy for each fixed pair $(\theta,\phi)$. The argument relies on an algebraic cancellation and the stability of kernel smoothing: \textbf{(i)} We first decompose both the parametric and intrinsic energies into their potential, entropy, and cross-modal divergence terms. Their potential and entropy terms cancel in the difference, leaving only discrepancies between KL divergences involving the true marginal densities and their kernel-smoothed counterparts. \textbf{(ii)} Because the true densities are bounded away from zero and the smoothed densities converge uniformly to them, all relevant logarithms remain in a Lipschitz regime for sufficiently small $\tau$. This allows the remaining log-ratio perturbations to be bounded directly by the KDE sup-errors of the two modalities. \textbf{(iii)} Since both KDE mismatches vanish as $\tau\to0^+$, the discrepancy between the parametric and intrinsic energy values also vanishes for each fixed $(\theta,\phi)$.
\end{proofintuition}

\begin{proof}
Fix any $(\theta,\phi)\in\Theta\times\Phi$. Under \Cref{assump:mm_regularity}, $\mathrm{sim}$ is bounded and Borel measurable, hence both cross-modal potentials $U_{\theta\shortto\phi}$ and $U_{\phi\shortto\theta}$ from \Cref{def:directional_potential} are Borel measurable and bounded. Moreover, since $\mathcal Z$ is compact and $\rho_\theta,\rho_\phi$ are continuous with $\rho_\theta\ge\underline\rho_\theta>0$ and $\rho_\phi\ge\underline\rho_\phi>0$, both densities are bounded above and below. Hence $\log\rho_\theta$ and $\log\rho_\phi$ are bounded, and therefore $D_{\mathrm{KL}}(\rho_\theta\|\rho_\phi)$ and $D_{\mathrm{KL}}(\rho_\phi\|\rho_\theta)$ are finite. Consequently, the multimodal functional $\mathcal F^{\mathrm{Sym}}_{\tau,\mathbf U_{\theta,\phi}}(\rho_\theta,\rho_\phi)$ is well-defined.

\textbf{Step 1: Functional discrepancy reduces to log-ratio perturbations.}
Recall the directional multimodal parametric energies from \Cref{def:sym_gef}:
\[
\mathcal J_\tau^{x\shortto y}(\theta,\phi)
=
\frac{1}{\tau}
\int_{\mathcal Z}U_{\theta\shortto\phi}(z)\,q_\theta(dz),
-
H_\times\!\left(q_\theta,\tilde\rho_{\phi,\tau}\right),
\qquad
\mathcal J_\tau^{y\shortto x}(\phi,\theta)
=
\frac{1}{\tau}
\int_{\mathcal Z}U_{\phi\shortto\theta}(z)\,q_\phi(dz),
-
H_\times\!\left(q_\phi,\tilde\rho_{\theta,\tau}\right),
\]
and
\[
\mathcal J_\tau^{\mathrm{Sym}}(\theta,\phi)
=
\frac12
\left(
\mathcal J_\tau^{x\shortto y}(\theta,\phi)
+
\mathcal J_\tau^{y\shortto x}(\phi,\theta)
\right).
\]
Using
\[
H_\times(q_1,\rho_2)
=
H(\rho_1)
+
D_{\mathrm{KL}}(\rho_1\|\rho_2),
\qquad
\rho_1:=\frac{\mathrm dq_1}{\mathrm d\mu},
\]
we can rewrite
\begin{equation}
\label{eq:mm_J_decomp}
\mathcal J_\tau^{\mathrm{Sym}}(\theta,\phi)
=
\frac12
\left(
\mathcal F_{\tau,U_{\theta\shortto\phi}}(\rho_\theta)
+
\mathcal F_{\tau,U_{\phi\shortto\theta}}(\rho_\phi)
\right)
-
\frac12
\left(
D_{\mathrm{KL}}(\rho_\theta\|\tilde\rho_{\phi,\tau})
+
D_{\mathrm{KL}}(\rho_\phi\|\tilde\rho_{\theta,\tau})
\right).
\end{equation}
On the other hand, by \Cref{def:mm_free_energy}, with
\[
D_{\mathrm{KL}}^{\mathrm{Sym}}(\rho_\theta,\rho_\phi)
:=
\frac12
\left(
D_{\mathrm{KL}}(\rho_\theta\|\rho_\phi)
+
D_{\mathrm{KL}}(\rho_\phi\|\rho_\theta)
\right),
\]
we have
\begin{equation}
\label{eq:mm_E_decomp}
\mathcal F^{\mathrm{Sym}}_{\tau,\mathbf U_{\theta,\phi}}(\rho_\theta,\rho_\phi)
=
\frac12
\left(
\mathcal F_{\tau,U_{\theta\shortto\phi}}(\rho_\theta)
+
\mathcal F_{\tau,U_{\phi\shortto\theta}}(\rho_\phi)
\right)
-
D_{\mathrm{KL}}^{\mathrm{Sym}}(\rho_\theta,\rho_\phi).
\end{equation}
Subtracting \Cref{eq:mm_J_decomp} from \Cref{eq:mm_E_decomp}, the potential and entropy terms cancel, giving the exact identity
\begin{equation}
\label{eq:mm_exact_gap}
\begin{aligned}
\mathcal F^{\mathrm{Sym}}_{\tau,\mathbf U_{\theta,\phi}}(\rho_\theta,\rho_\phi)
-
\mathcal J_\tau^{\mathrm{Sym}}(\theta,\phi)
=
\frac12
\Big(
D_{\mathrm{KL}}(\rho_\theta\|\tilde\rho_{\phi,\tau})
-
D_{\mathrm{KL}}(\rho_\theta\|\rho_\phi)
+
D_{\mathrm{KL}}(\rho_\phi\|\tilde\rho_{\theta,\tau})
-
D_{\mathrm{KL}}(\rho_\phi\|\rho_\theta)
\Big).
\end{aligned}
\end{equation}
For any $\mu$-densities $p,r_1,r_2$ with $r_1,r_2>0$ $\mu$-a.e., the elementary difference identity
\[
D_{\mathrm{KL}}(p\|r_1)
-
D_{\mathrm{KL}}(p\|r_2)
=
\int_{\mathcal Z}
p(z)
\log\frac{r_2(z)}{r_1(z)}
\,\mu(\mathrm dz)
\]
holds. Since $\kappa_\tau>0$, the smoothed densities $\tilde\rho_{\theta,\tau}$ and $\tilde\rho_{\phi,\tau}$ are strictly positive. Applying the identity with $(p,r_1,r_2)=(\rho_\theta,\tilde\rho_{\phi,\tau},\rho_\phi)$ and $(p,r_1,r_2)=(\rho_\phi,\tilde\rho_{\theta,\tau},\rho_\theta)$ gives
\begin{equation}
\label{eq:mm_exact_logratio}
\begin{aligned}
\mathcal F^{\mathrm{Sym}}_{\tau,\mathbf U_{\theta,\phi}}(\rho_\theta,\rho_\phi)
-
\mathcal J_\tau^{\mathrm{Sym}}(\theta,\phi)
=
\frac12
\Bigg(
\int_{\mathcal Z}
\rho_\theta(z)
\log
\frac{\rho_\phi(z)}
{\tilde\rho_{\phi,\tau}(z)}
\,\mu(\mathrm dz)
+
\int_{\mathcal Z}
\rho_\phi(z)
\log
\frac{\rho_\theta(z)}
{\tilde\rho_{\theta,\tau}(z)}
\,\mu(\mathrm dz)
\Bigg).
\end{aligned}
\end{equation}

\textbf{Step 2: Nonasymptotic bound by KDE sup-errors.}
By the lower bounds $\rho_\theta\ge\underline\rho_\theta$ and $\rho_\phi\ge\underline\rho_\phi$ and \Cref{lem:sharp_peak_smoothing} applied separately to $\rho_\theta$ and $\rho_\phi$, there exists $\tau_0(\theta,\phi)>0$ such that, for all $0<\tau\le\tau_0(\theta,\phi)$,
\[
\varepsilon_{\mathrm{kde}}^{(\theta)}(\tau)
:=
\|\tilde\rho_{\theta,\tau}-\rho_\theta\|_\infty
\le
\frac{\underline\rho_\theta}{2},
\qquad
\varepsilon_{\mathrm{kde}}^{(\phi)}(\tau)
:=
\|\tilde\rho_{\phi,\tau}-\rho_\phi\|_\infty
\le
\frac{\underline\rho_\phi}{2}.
\]
Hence, for every $z\in\mathcal Z$ and all such $\tau$,
\begin{equation}
\label{eq:mm_tilde_lower}
\tilde\rho_{\theta,\tau}(z)
\ge
\frac{\underline\rho_\theta}{2},
\qquad
\tilde\rho_{\phi,\tau}(z)
\ge
\frac{\underline\rho_\phi}{2}.
\end{equation}
On $[\underline\rho_\phi/2,\infty)$, the map $\log(\cdot)$ is Lipschitz with constant $2/\underline\rho_\phi$. Therefore, for every $z\in\mathcal Z$,
\[
\left|
\log\rho_\phi(z)
-
\log\tilde\rho_{\phi,\tau}(z)
\right|
\le
\frac{2}{\underline\rho_\phi}
\left|
\rho_\phi(z)
-
\tilde\rho_{\phi,\tau}(z)
\right|
\le
\frac{
2\,\varepsilon_{\mathrm{kde}}^{(\phi)}(\tau)
}{
\underline\rho_\phi
}.
\]
Equivalently,
\begin{equation}
\label{eq:mm_logratio_phi}
\left|
\log
\frac{\rho_\phi(z)}
{\tilde\rho_{\phi,\tau}(z)}
\right|
\le
\frac{
2\,\varepsilon_{\mathrm{kde}}^{(\phi)}(\tau)
}{
\underline\rho_\phi
}.
\end{equation}
Analogously,
\begin{equation}
\label{eq:mm_logratio_theta}
\left|
\log
\frac{\rho_\theta(z)}
{\tilde\rho_{\theta,\tau}(z)}
\right|
\le
\frac{
2\,\varepsilon_{\mathrm{kde}}^{(\theta)}(\tau)
}{
\underline\rho_\theta
}.
\end{equation}
Using \Cref{eq:mm_exact_logratio}, together with
\[
\int_{\mathcal Z}\rho_\theta(z)\,\mu(\mathrm dz)
=
\int_{\mathcal Z}\rho_\phi(z)\,\mu(\mathrm dz)
=
1,
\]
we obtain
\begin{align*}
\left|
\mathcal F^{\mathrm{Sym}}_{\tau,\mathbf U_{\theta,\phi}}(\rho_\theta,\rho_\phi)
-
\mathcal J_\tau^{\mathrm{Sym}}(\theta,\phi)
\right|
&\le
\frac12
\left(
\int_{\mathcal Z}
\rho_\theta(z)
\left|
\log
\frac{\rho_\phi(z)}
{\tilde\rho_{\phi,\tau}(z)}
\right|
\mu(\mathrm dz)
+
\int_{\mathcal Z}
\rho_\phi(z)
\left|
\log
\frac{\rho_\theta(z)}
{\tilde\rho_{\theta,\tau}(z)}
\right|
\mu(\mathrm dz)
\right)
\\[0.1em]
&\le
\frac12
\left(
\left\|
\log
\frac{\rho_\phi}
{\tilde\rho_{\phi,\tau}}
\right\|_\infty
+
\left\|
\log
\frac{\rho_\theta}
{\tilde\rho_{\theta,\tau}}
\right\|_\infty
\right)
\\[0.1em]
&\le
\frac12
\left(
\frac{
2\,\varepsilon_{\mathrm{kde}}^{(\phi)}(\tau)
}{
\underline\rho_\phi
}
+
\frac{
2\,\varepsilon_{\mathrm{kde}}^{(\theta)}(\tau)
}{
\underline\rho_\theta
}
\right)
\\[0.1em]
&\le
\frac{
\varepsilon_{\mathrm{kde}}^{(\theta)}(\tau)
+
\varepsilon_{\mathrm{kde}}^{(\phi)}(\tau)
}{
\min\{\underline\rho_\theta,\underline\rho_\phi\}
}.
\end{align*}
This proves the claimed nonasymptotic bound for all $0<\tau\le\tau_0(\theta,\phi)$.

\textbf{Step 3: Vanishing discrepancy as $\tau\to0^+$.}
By \Cref{lem:sharp_peak_smoothing} applied separately to $\rho_\theta$ and $\rho_\phi$ under \Cref{ass:sharp_peak},
\[
\varepsilon_{\mathrm{kde}}^{(\theta)}(\tau)
\longrightarrow0,
\qquad
\varepsilon_{\mathrm{kde}}^{(\phi)}(\tau)
\longrightarrow0
\qquad
\text{as }\tau\to0^+.
\]
Therefore, the bound in Step~2 implies
\[
\left|
\mathcal J_\tau^{\mathrm{Sym}}(\theta,\phi)
-
\mathcal F^{\mathrm{Sym}}_{\tau,\mathbf U_{\theta,\phi}}(\rho_\theta,\rho_\phi)
\right|
\longrightarrow0
\qquad
\text{as }\tau\to0^+,
\]
for every fixed $(\theta,\phi)\in\Theta\times\Phi$.
\end{proof}

\section{Extended Formal Statements}

This section provides extended formal statements by connecting the intrinsic geometric insights to parameterized training objectives, when the encoder families are sufficiently expressive.

\subsection{Unimodal Parametric Inheritance and Emergence of Uniformity}
\label{app:unimodal_inheritance}

This subsection formalizes the intuition used in \Cref{sec:unimodal_intrinsic}: unimodal InfoNCE inherits the intrinsic Gibbs geometry when (i) the kernel smoothing bias is small and (ii) the encoder family is expressive enough to (approximately) realize the intrinsic Gibbs equilibrium at the best attainable potential level.

\paragraph{Recap of objects.}
Fix $\tau>0$. For each $\theta\in\Theta$, recall the induced representation law $q_\theta\in\mathcal P(\mathcal Z)$ and encoded positive-pair law $\pi_{\theta\theta}\in\mathcal P(\mathcal Z\times\mathcal Z)$. Under the same setting of \Cref{thm:intrinsic_unimodal}, assume $q_\theta\ll\mu$ and write $\rho_\theta(z):=q_\theta(\mathrm dz)/\mu(\mathrm dz)$. Let $U_\theta$ be the alignment potential field induced by $\pi_{\theta\theta}$ (\Cref{def:align_potential}). The intrinsic functional is $\mathcal F_{\tau,U_\theta}(\rho)$ (\Cref{eq:free_energy_def}), and the parametric energy is $\mathcal J_\tau(\theta)$ (\Cref{def:uni_gef}).

\paragraph{Free-energy gap equals KL divergence.}
From \Cref{prop:gibbs_equilibrium}, for any fixed potential $U$ and temperature $\tau$, define the partition function
\begin{equation}
\label{eq:Gibbs_constant}
Z_\tau
:=
\int_{\mathcal Z}
\exp\!\big(-U(z)/\tau\big)\,\mu(\mathrm dz),
\end{equation}
and the associated Gibbs density
\begin{equation}
\label{eq:Gibbs_density}
\rho^*(z)
:=
\frac{\exp\!\big(-U(z)/\tau\big)}{Z_\tau}.
\end{equation}
Then, for any $\rho\in\mathcal D_\mu(\mathcal Z)$ with
$\rho\log(\rho/\rho^*)\in L^1(\mu)$,
\begin{equation}
\label{eq:free_energy_kl_identity}
\mathcal F_{\tau,U}(\rho)+\log Z_\tau
=
D_{\mathrm{KL}}\!\big(\rho\|\rho^*\big)
\ge 0.
\end{equation}
In particular, $\rho^*$ is the unique minimizer of $\mathcal F_{\tau,U}$, and the free-energy suboptimality controls proximity to the Gibbs equilibrium via KL divergence.

\begin{remark}[Proof of \Cref{eq:free_energy_kl_identity}]
Expanding
\[
D_{\mathrm{KL}}(\rho\|\rho^*)
=
\int_{\mathcal Z}
\rho(z)\log\frac{\rho(z)}{\rho^*(z)}\,\mu(\mathrm dz)
\]
and substituting $\log\rho^*=-(U/\tau)-\log Z_\tau$ yields \Cref{eq:free_energy_kl_identity}.
\end{remark}

\paragraph{Expressivity at the best attainable potential level.}
The intrinsic minimum of $\mathcal F_{\tau,U_\theta}$ equals $-\log Z_{\theta,\tau}$ by \Cref{eq:free_energy_kl_identity}, where
\[
Z_{\theta,\tau}
:=
\int_{\mathcal Z}
\exp\!\big(-U_\theta(z)/\tau\big)\,\mu(\mathrm dz)
\]
is the normalization constant of the Gibbs density. Whenever the minimum is attained, define the \emph{best attainable potential level} at temperature $\tau$ by choosing any
\begin{equation}
\label{eq:best_potential_level_def}
\hat\theta_\tau
\in
\arg\min_{\theta\in\Theta}
\bigl(-\log Z_{\theta,\tau}\bigr).
\end{equation}

\begin{assumption}[Expressivity at the best level]
\label{ass:unimodal_realizability}
There exists a function $\varepsilon_{\mathrm{real}}(\tau)\to0$ as $\tau\to0^+$ such that, for $\hat\theta_\tau$ in \Cref{eq:best_potential_level_def},
\begin{equation}
\label{eq:realizability_gap}
\mathcal F_{\tau,U_{\hat\theta_\tau}}
(\rho_{\hat\theta_\tau})
+
\log Z_{\hat\theta_\tau,\tau}
\le
\varepsilon_{\mathrm{real}}(\tau).
\end{equation}
Equivalently, at the best attainable potential level, the encoder-induced density $\rho_{\hat\theta_\tau}$ attains the intrinsic minimum of $\mathcal F_{\tau,U_{\hat\theta_\tau}}$ up to error $\varepsilon_{\mathrm{real}}(\tau)$.
\end{assumption}

\paragraph{Ground-state inheritance under encoder expressivity.}
Under \Cref{ass:sharp_peak} and mild regularity conditions, \Cref{thm:intrinsic_unimodal} bounds the gap between $\mathcal J_\tau(\theta)$ and $\mathcal F_{\tau,U_\theta}(\rho_\theta)$ by the KDE mismatch $\varepsilon_{\mathrm{kde}}^{(\theta)}(\tau)$. By \Cref{lem:sharp_peak_smoothing}, this mismatch vanishes for each fixed $\theta$. Since the relevant optimizers may depend on $\tau$, we state the following inheritance result along optimizing sequences for which the corresponding energy approximation errors vanish at the evaluated temperatures.

\begin{corollary}[Unimodal parametric inheritance along optimizing sequences]
\label{cor:unimodal_parametric_inheritance}
For each $\tau>0$, assume the following minimizers exist, and let
\[
\theta_\tau^*
\in
\arg\min_{\theta\in\Theta}\mathcal J_\tau(\theta),
\qquad
\hat\theta_\tau
\in
\arg\min_{\theta\in\Theta}
\bigl(-\log Z_{\theta,\tau}\bigr).
\]
Assume the conditions of \Cref{thm:intrinsic_unimodal} hold for $\theta_\tau^*$ and $\hat\theta_\tau$ for all sufficiently small $\tau>0$. Assume additionally that the expressivity condition in \Cref{ass:unimodal_realizability} holds along the sequence $\hat\theta_\tau$, and that the energy approximation errors satisfy
\[
\eta_\tau(\theta_\tau^*)
+
\eta_\tau(\hat\theta_\tau)
\to0
\qquad
\text{as }\tau\to0^+,\qquad\text{where }\ 
\eta_\tau(\theta)
:=
\big|
\mathcal J_\tau(\theta)
-
\mathcal F_{\tau,U_\theta}(\rho_\theta)
\big|.
\]
Further, let $\rho_{\theta_\tau^*}^{*,\tau}$ denote the Gibbs density associated with $U_{\theta_\tau^*}$ at temperature $\tau$:
\[
\rho_{\theta_\tau^*}^{*,\tau}(z)
:=
\frac{
\exp\!\big(-U_{\theta_\tau^*}(z)/\tau\big)
}{
Z_{\theta_\tau^*,\tau}
},
\qquad
Z_{\theta_\tau^*,\tau}
:=
\int_{\mathcal Z}
\exp\!\big(-U_{\theta_\tau^*}(z')/\tau\big)\,
\mu(\mathrm dz').
\]
Then,
\[
D_{\mathrm{KL}}\!\big(
\rho_{\theta_\tau^*}
\|
\rho_{\theta_\tau^*}^{*,\tau}
\big)
\to0\qquad\text{and}\qquad
\big\|
\rho_{\theta_\tau^*}
-
\rho_{\theta_\tau^*}^{*,\tau}
\big\|_{\mathrm{TV}}
\to0.
\]
Consequently, the learned representation measures inherit any low-temperature concentration property of the associated Gibbs sequence. In particular, if for some fixed $\sigma>0$ and some family of open sets $\mathcal O_\tau\subset\mathcal Z$ satisfying
\[
\mathcal O_\tau
\supset
\mathcal W^\sigma_{\theta_\tau^*}
:=
\Big\{
z\in\mathcal Z:
U_{\theta_\tau^*}(z)
\le
\operatorname*{ess\,inf}_{\mu}
U_{\theta_\tau^*}
+
\sigma
\Big\},
\]
the Gibbs sequence satisfies $(\rho_{\theta_\tau^*}^{*,\tau}\mu)(\mathcal Z\setminus\mathcal O_\tau)\to0$, 
then
\[
q_{\theta_\tau^*}
(\mathcal Z\setminus\mathcal O_\tau)
\to0.
\]
\end{corollary}

\begin{proofintuition}
The goal is to show that a sequence of parametric minimizers $\theta_\tau^*$ tracks the Gibbs equilibrium induced by its own alignment potential as $\tau\to0^+$. The proof has three steps. \textbf{(i)} For any fixed induced potential field $U_\theta$, the intrinsic free-energy gap equals the KL divergence from the corresponding Gibbs density. \textbf{(ii)} By the expressivity condition, the comparator $\hat\theta_\tau$ can approach the intrinsic optimum at the best attainable potential level. Since $\theta_\tau^*$ minimizes the parametric energy $\mathcal J_\tau$, and since the sequence-level energy approximation errors vanish for both $\theta_\tau^*$ and $\hat\theta_\tau$, the intrinsic free-energy gap of $\rho_{\theta_\tau^*}$ also vanishes. Hence $D_{\mathrm{KL}}(\rho_{\theta_\tau^*}\|\rho_{\theta_\tau^*}^{*,\tau})\to0$. \textbf{(iii)} Pinsker's inequality gives convergence in total variation, so any low-temperature concentration property of the Gibbs sequence transfers to the learned representation measures.
\end{proofintuition}

\begin{proof}
\textbf{Step 1: Well-defined ``best achievable'' equilibrium.}
We first isolate the intrinsic quantity that governs the ``best achievable'' equilibrium value at temperature $\tau$. Fix any $\theta\in\Theta$ and $\tau>0$. By the standing setup, $\mathcal Z$ is compact with $\mu(\mathcal Z)<\infty$; and by \Cref{assump:regularity}, $\mathrm{sim}:\mathcal Z\times\mathcal Z\to\mathbb R$ is continuous. Hence, by the extreme value theorem, $\mathrm{sim}$ is bounded on
$\mathcal Z\times\mathcal Z$:
\[
\mathrm{Sim}_{\max}
:=
\max_{z,w\in\mathcal Z}
\mathrm{sim}(z,w)
<
\infty.
\]
The induced alignment potential field (\Cref{def:align_potential}) is
\[
U_\theta(z)
=
-
\int_{\mathcal Z}
\mathrm{sim}(z,w)\,
\nu_{\theta,z}(\mathrm dw),
\]
hence
\[
U_\theta(z)
\ge
-\mathrm{Sim}_{\max}
\qquad
\text{for all }z\in\mathcal Z.
\]
Therefore, for any fixed $\tau>0$,
\[
0
<
\exp\!\left(
-\frac{U_\theta(z)}{\tau}
\right)
\le
\exp\!\left(
\frac{\mathrm{Sim}_{\max}}{\tau}
\right),
\qquad
z\in\mathcal Z,
\]
and thus
\[
0
<
Z_{\theta,\tau}
:=
\int_{\mathcal Z}
\exp\!\left(
-\frac{U_\theta(z)}{\tau}
\right)
\,\mu(\mathrm dz)
\le
\mu(\mathcal Z)
\exp\!\left(
\frac{\mathrm{Sim}_{\max}}{\tau}
\right)
<
\infty.
\]
Thus, the representation energy
\[
\mathcal F_{\tau,U_\theta}(\rho)
=
\frac{1}{\tau}
\int_{\mathcal Z}
U_\theta(z)\rho(z)\,\mu(\mathrm dz)
-
H(\rho)
\]
admits the unique Gibbs minimizer
\(
\rho_\theta^{*,\tau}(z)
=
\frac{
\exp\!\big(-U_\theta(z)/\tau\big)
}{
Z_{\theta,\tau}
}
\)
by \Cref{prop:gibbs_equilibrium}. The corresponding minimum is
\begin{equation}
\label{eq:lowest_level}
\inf_{\rho\in\mathcal D_\mu(\mathcal Z)}
\mathcal F_{\tau,U_\theta}(\rho)
=
\mathcal F_{\tau,U_\theta}
(\rho_\theta^{*,\tau})
=
-\log Z_{\theta,\tau}.
\end{equation}
We interpret $-\log Z_{\theta,\tau}$ as the lowest representation energy level induced by the potential field $U_\theta$, which is the best intrinsic value attainable by any density in $\mathcal D_\mu(\mathcal Z)$ once $U_\theta$ is fixed.

\textbf{Step 2: Representation energy gap equals a KL divergence.}
Fix $\theta$ and $\tau$. Recall that, by \Cref{prop:gibbs_equilibrium}, the unique Gibbs equilibrium is
\[
\rho_\theta^{*,\tau}(z)
=
\frac{
\exp\!\big(-U_\theta(z)/\tau\big)
}{
Z_{\theta,\tau}
},
\]
so
\[
\log\rho_\theta^{*,\tau}(z)
=
-\frac{U_\theta(z)}{\tau}
-
\log Z_{\theta,\tau}.
\]
Applying \Cref{eq:free_energy_kl_identity} with $\rho=\rho_\theta$ and $\rho^*=\rho_\theta^{*,\tau}$ yields
\[
D_{\mathrm{KL}}\!\big(
\rho_\theta\|
\rho_\theta^{*,\tau}
\big)
=
\mathcal F_{\tau,U_\theta}(\rho_\theta)
+
\log Z_{\theta,\tau}.
\]
Finally, by \Cref{eq:lowest_level},
\[
\inf_{\rho\in\mathcal D_\mu(\mathcal Z)}
\mathcal F_{\tau,U_\theta}(\rho)
=
\mathcal F_{\tau,U_\theta}
(\rho_\theta^{*,\tau})
=
-\log Z_{\theta,\tau},
\]
hence
\begin{equation}
\label{eq:FE_KL_identity_cor}
\mathcal F_{\tau,U_\theta}(\rho_\theta)
-
\inf_{\rho\in\mathcal D_\mu(\mathcal Z)}
\mathcal F_{\tau,U_\theta}(\rho)
=
\mathcal F_{\tau,U_\theta}(\rho_\theta)
+
\log Z_{\theta,\tau}
=
D_{\mathrm{KL}}\!\big(
\rho_\theta\|
\rho_\theta^{*,\tau}
\big).
\end{equation}

\textbf{Step 3: Transfer intrinsic optimality to the optimizing sequence.}
For each $\tau>0$, let
\[
\theta_\tau^*
\in
\arg\min_{\theta\in\Theta}
\mathcal J_\tau(\theta),
\qquad
\hat\theta_\tau
\in
\arg\min_{\theta\in\Theta}
\bigl(-\log Z_{\theta,\tau}\bigr).
\]
Define
\[
\eta_\tau(\theta)
:=
\big|
\mathcal J_\tau(\theta)
-
\mathcal F_{\tau,U_\theta}(\rho_\theta)
\big|.
\]
By the sequence condition in the statement,
\[
\eta_\tau(\theta_\tau^*)
+
\eta_\tau(\hat\theta_\tau)
\to0.
\]
By optimality,
\[
\mathcal J_\tau(\theta_\tau^*)
\le
\mathcal J_\tau(\hat\theta_\tau),
\]
hence
\begin{align*}
\mathcal F_{\tau,U_{\theta_\tau^*}}
(\rho_{\theta_\tau^*})
&\le
\mathcal J_\tau(\theta_\tau^*)
+
\eta_\tau(\theta_\tau^*)
\\
&\le
\mathcal J_\tau(\hat\theta_\tau)
+
\eta_\tau(\theta_\tau^*)
\\
&\le
\mathcal F_{\tau,U_{\hat\theta_\tau}}
(\rho_{\hat\theta_\tau})
+
\eta_\tau(\theta_\tau^*)
+
\eta_\tau(\hat\theta_\tau).
\end{align*}
Adding $\log Z_{\theta_\tau^*,\tau}$ and using
\[
\log Z_{\theta_\tau^*,\tau}
\le
\log Z_{\hat\theta_\tau,\tau}
\]
by the definition of $\hat\theta_\tau$, we obtain
\begin{align*}
\mathcal F_{\tau,U_{\theta_\tau^*}}
(\rho_{\theta_\tau^*})
+
\log Z_{\theta_\tau^*,\tau}
\;\le\;
\mathcal F_{\tau,U_{\hat\theta_\tau}}
(\rho_{\hat\theta_\tau})
+
\log Z_{\hat\theta_\tau,\tau}
+
\eta_\tau(\theta_\tau^*)
+
\eta_\tau(\hat\theta_\tau).
\end{align*}
Using \Cref{eq:realizability_gap} along the sequence $\hat\theta_\tau$ gives
\[
\mathcal F_{\tau,U_{\theta_\tau^*}}
(\rho_{\theta_\tau^*})
+
\log Z_{\theta_\tau^*,\tau}
\le
\varepsilon_{\mathrm{real}}(\tau)
+
\eta_\tau(\theta_\tau^*)
+
\eta_\tau(\hat\theta_\tau).
\]
By \Cref{eq:FE_KL_identity_cor}, applied at
$\theta=\theta_\tau^*$,
\[
D_{\mathrm{KL}}\!\big(
\rho_{\theta_\tau^*}
\|
\rho_{\theta_\tau^*}^{*,\tau}
\big)
\le
\varepsilon_{\mathrm{real}}(\tau)
+
\eta_\tau(\theta_\tau^*)
+
\eta_\tau(\hat\theta_\tau)
\xrightarrow{\tau\to0^+}
0.
\]
Therefore, by Pinsker's inequality,
\begin{equation}
\label{eq:pinsker_cor}
\big\|
\rho_{\theta_\tau^*}
-
\rho_{\theta_\tau^*}^{*,\tau}
\big\|_{\mathrm{TV}}
\le
\sqrt{
\tfrac12
D_{\mathrm{KL}}\!\big(
\rho_{\theta_\tau^*}
\|
\rho_{\theta_\tau^*}^{*,\tau}
\big)
}
\xrightarrow{\tau\to0^+}
0.
\end{equation}

Now let $\mathcal O_\tau$ be any family of open sets satisfying
\[
\mathcal O_\tau
\supset
\mathcal W^\sigma_{\theta_\tau^*}
\qquad\text{and}\qquad
(\rho_{\theta_\tau^*}^{*,\tau}\mu)
(\mathcal Z\setminus\mathcal O_\tau)
\to0.
\]
By the variational characterization of total variation, for any measurable $\mathcal A\subseteq\mathcal Z$,
\[
\big|
q_{\theta_\tau^*}(\mathcal A)
-
(\rho_{\theta_\tau^*}^{*,\tau}\mu)(\mathcal A)
\big|
\le
\big\|
\rho_{\theta_\tau^*}
-
\rho_{\theta_\tau^*}^{*,\tau}
\big\|_{\mathrm{TV}}.
\]
Taking $\mathcal A=\mathcal Z\setminus\mathcal O_\tau$ gives
\[
q_{\theta_\tau^*}
(\mathcal Z\setminus\mathcal O_\tau)
\le
(\rho_{\theta_\tau^*}^{*,\tau}\mu)
(\mathcal Z\setminus\mathcal O_\tau)
+
\big\|
\rho_{\theta_\tau^*}
-
\rho_{\theta_\tau^*}^{*,\tau}
\big\|_{\mathrm{TV}}
\xrightarrow{\tau\to0^+}
0.
\]
This proves the claimed inherited concentration.
\end{proof}

\begin{remark}[Why this implies uniformity is a tie-breaker]
Inside the aligned basin
\[
\big\{
z:
U_\theta(z)
\approx
\operatorname*{ess\,inf}\nolimits_{\mu}U_\theta
\big\},
\]
the potential term in $\mathcal F_{\tau,U_\theta}$ is nearly constant. In that regime, minimizing $\mathcal F_{\tau,U_\theta}$ reduces approximately to maximizing entropy subject to the remaining constraints imposed by the encoder family, so the selected configuration is the most internally dispersed distribution compatible with being well-aligned. This is the precise sense in which ``uniformity'' in unimodal contrastive learning is not a competing global force but an entropic tie-breaker within the alignment basin.
\end{remark}

\subsection{Multimodal Parametric Inheritance}
\label{app:multimodal_inheritance}

This subsection formalizes the sense in which the practical multimodal energy $\mathcal J^{\mathrm{Sym}}_\tau(\theta,\phi)$ inherits the intrinsic barrier-driven coordinate geometry of $\mathcal F^{\mathrm{Sym}}_{\tau,\mathbf U}$ when the encoder family is sufficiently expressive. Concretely, fixing one modality (say $\phi$), we show that if the image encoder family can realize (up to a vanishing gap) the intrinsic coordinate infimum against the peer density, then any parametric minimizer concentrates its \emph{excess} mass near minimizers of the effective barrier potential $V_{\theta\mid\phi}=\tfrac{1}{\tau}U_{\theta\shortto\phi}+\log\rho_\phi$.

\paragraph{Background-plus-excess decomposition.}
Given a density floor $\underline\rho_i>0$ with $\underline\rho_i\,\mu(\mathcal Z)<1$, any feasible density can be decomposed as
\[
\rho_i(z)=\underline\rho_i+M_{\mathrm{ex}}^{(i)}\bar\rho_i(z),
\quad\text{with }\ 
M_{\mathrm{ex}}^{(i)}:=1-\underline\rho_i\,\mu(\mathcal Z),
\quad\text{satisfying }\ 
\bar\rho_i\ge0,\quad \int_{\mathcal Z}\bar\rho_i(z)\,\mu(\mathrm dz)=1.
\]
Here $\underline\rho_i$ plays the role of a fixed background mass and $\bar\rho_i=(\rho_i-\underline\rho_i)/M_{\mathrm{ex}}^{(i)}$ is the normalized distribution of the remaining excess budget $M_{\mathrm{ex}}^{(i)}$.

\begin{corollary}[Multimodal parametric inheritance under model expressivity]
\label{cor:multimodal_inheritance}
Assume the setting of \Cref{thm:intrinsic_multimodal}. Fix a peer encoder $\phi\in\Phi$ and, for each $\tau>0$, assume a minimizer exists and let $\theta_\tau^*\in\arg\min_{\theta\in\Theta}\mathcal J_\tau^{\mathrm{Sym}}(\theta,\phi)$. Write the density floor parameterized by $\theta_\tau^*$ as $\underline\rho_{\theta_\tau^*}$ with $\underline\rho_{\theta_\tau^*}\mu(\mathcal Z)<1$, and define the set of strictly positive $\mu$-densities bounded below by $\underline\rho_{\theta_\tau^*}$ as
\[
\mathcal D_{\mu,\underline\rho_{\theta_\tau^*}}(\mathcal Z)
:=
\big\{
\rho\in\mathcal D_\mu(\mathcal Z)\cap L^\infty(\mu):
\rho\ge\underline\rho_{\theta_\tau^*}\ \mu\text{-a.e.}
\big\},
\]
and assume that the corresponding excess budget remains uniformly nondegenerate for all sufficiently small $\tau$, 
\[
M_{\mathrm{ex}}^{(\theta_\tau^*)}
:=
1-\underline\rho_{\theta_\tau^*}\mu(\mathcal Z)
\ge m_{\mathrm{ex}}, \qquad\text{for some }\ m_{\mathrm{ex}}>0.
\]
Define the normalized excess density as
\(
\bar\rho_{\theta_\tau^*}(z)
:=
{(\rho_{\theta_\tau^*}(z)-\underline\rho_{\theta_\tau^*})}
/{M_{\mathrm{ex}}^{(\theta_\tau^*)}}.
\)
Further, assume the encoder family $\{f_\theta\}_\Theta$ is sufficiently expressive: there exists $\varepsilon_{\mathrm{real}}(\tau)\to0$ as $\tau\downarrow0^+$ such that
\begin{equation}
\label{eq:mm_coord_gap}
\mathcal F^{\mathrm{Sym}}_{\tau,\mathbf U_{\theta_\tau^*,\phi}}
(\rho_{\theta_\tau^*},\rho_\phi)
\le
\inf_{\rho\in\mathcal D_{\mu,\underline\rho_{\theta_\tau^*}}(\mathcal Z)}
\mathcal F^{\mathrm{Sym}}_{\tau,\mathbf U_{\theta_\tau^*,\phi}}
(\rho,\rho_\phi)
+
\varepsilon_{\mathrm{real}}(\tau).
\end{equation}
Let
\(
V_{\theta_\tau^*\mid\phi}(z)
:=
\frac{1}{\tau}U_{\theta_\tau^*\shortto\phi}(z)
+
\log\rho_\phi(z)
\)
be the effective potential field parameterized by $(\theta_\tau^*,\phi)$. For $\sigma>0$, define
\[
\mathcal W_{\theta_\tau^*}^{\sigma}
:=
\big\{
z\in\mathcal Z:
V_{\theta_\tau^*\mid\phi}(z)
\le
\operatorname*{ess\,inf}\nolimits_{\mu}V_{\theta_\tau^*\mid\phi}
+
\sigma
\big\}.
\]
Then, for every $\sigma>0$,
\begin{equation}
\label{eq:mm_excess_leakage}
\bar\rho_{\theta_\tau^*}\mu
\big(
\mathcal Z\setminus\mathcal W_{\theta_\tau^*}^{\sigma}
\big)
\le
\frac{
2\,\varepsilon_{\mathrm{real}}(\tau)
}{
\sigma M_{\mathrm{ex}}^{(\theta_\tau^*)}
}.
\end{equation}
Moreover, if $\sigma=\sigma(\tau)\downarrow0$ satisfies $\varepsilon_\mathrm{real}(\tau)/\sigma\to0$, then
\[
\bar\rho_{\theta_\tau^*}\mu
\big(
\mathcal Z\setminus
\mathcal W_{\theta_\tau^*}^{\sigma(\tau)}
\big)
\longrightarrow0.
\]
A symmetric statement holds for optimizing in $\phi$ with $\theta$ fixed.
\end{corollary}

\begin{proofintuition}
The proof proceeds by decoupling the induced density $\rho_{\theta_\tau^*}$ into a fixed background floor $\underline\rho_{\theta_\tau^*}$ and a free excess mass $M_{\mathrm{ex}}^{(\theta_\tau^*)}$. By the expressivity assumption, the parameterized model can closely approach the coordinate infimum of the intrinsic functional. We construct a sequence of infimizing competitors that concentrate all their excess mass inside shrinking sublevel sets of the effective barrier potential $V_{\theta_\tau^*\mid\phi}$, driving the cross-entropy term toward its lower bound $\log\underline\rho_{\theta_\tau^*}$. By comparing the energy of the parameterized density against these competitors, we show that any excess mass placed outside the sublevel set $\mathcal W_{\theta_\tau^*}^\sigma$ incurs a linear energy penalty. The expressivity gap $\varepsilon_{\mathrm{real}}(\tau)$ bounds this penalty, directly restricting the fraction of excess mass that can leak outside the effective-potential sublevel set.
\end{proofintuition}

\begin{proof}
We provide the derivation for modality $\mathcal X$ with parameter $\theta\in\Theta$. The corresponding result for $\phi\in\Phi$ follows by interchanging modalities and notation.

Fix $\phi\in\Phi$ and $\tau>0$, and let $\theta_\tau^*\in\arg\min_{\theta\in\Theta}\mathcal J_\tau^{\mathrm{Sym}}(\theta,\phi)$. Since $\rho_{\theta_\tau^*}\ge\underline\rho_{\theta_\tau^*}$ $\mu$-a.e., $M_{\mathrm{ex}}^{(\theta_\tau^*)}=1-\underline\rho_{\theta_\tau^*}\mu(\mathcal Z)>0$, and $\int_{\mathcal Z}\rho_{\theta_\tau^*}(z)\,\mu(\mathrm dz)=1$, we have $\bar\rho_{\theta_\tau^*}\ge0$ $\mu$-a.e.\ and
\[
\int_{\mathcal Z}
\bar\rho_{\theta_\tau^*}(z)\,\mu(\mathrm dz)
=
\frac{
1-\underline\rho_{\theta_\tau^*}\mu(\mathcal Z)
}{
M_{\mathrm{ex}}^{(\theta_\tau^*)}
}
=
1,
\]
hence $\bar\rho_{\theta_\tau^*}\mu$ is a valid probability measure.

Define the effective potential field and its essential infimum
\[
V_{\theta_\tau^*\mid\phi}(z)
=
\frac1\tau U_{\theta_\tau^*\shortto\phi}(z)
+
\log\rho_\phi(z),
\qquad
v_*
:=
\operatorname*{ess\,inf}_{\mu}
V_{\theta_\tau^*\mid\phi},
\]
and, for $\sigma>0$, define the sublevel set
\[
\mathcal W^\sigma_{\theta_\tau^*}
=
\left\{
z\in\mathcal Z:
V_{\theta_\tau^*\mid\phi}(z)
\le
v_*+\sigma
\right\}.
\]

\textbf{Step 1: A quantitative implication of near-optimality.}
Consider the coordinate functional with the peer encoder $\phi$ fixed:
\[
\mathcal F_\tau(\rho)
:=
\mathcal F^{\mathrm{Sym}}_{\tau,\mathbf U_{\theta_\tau^*,\phi}}
(\rho,\rho_\phi),
\qquad
\rho\in
\mathcal D_{\mu,\underline\rho_{\theta_\tau^*}}(\mathcal Z).
\]
As made explicit in the proof of \Cref{prop:mm_coordinate_extremal}, by expanding $\mathcal F^{\mathrm{Sym}}_{\tau,\mathbf U_{\theta_\tau^*,\phi}}$ and collecting $\rho$-dependent terms, $\mathcal F_\tau$ can be written in the form
\begin{equation}
\label{eq:Ft_decomp}
\mathcal F_\tau(\rho)
=
\frac12
\int_{\mathcal Z}
V_{\theta_\tau^*\mid\phi}(z)\rho(z)\,\mu(\mathrm dz)
+
\frac12
\int_{\mathcal Z}
\rho_\phi(z)\log\rho(z)\,\mu(\mathrm dz)
+
\mathrm{const}_\tau(\phi),
\end{equation}
where $\mathrm{const}_\tau(\phi)$ is independent of $\rho$. By the floor constraint $\rho\ge\underline\rho_{\theta_\tau^*}$ $\mu$-a.e., the monotonicity of the logarithm, and the fact that $\rho_\phi$ is a probability density, we have
\[
\int_{\mathcal Z}
\rho_\phi(z)\log\rho(z)\,\mu(\mathrm dz)
\ge
\int_{\mathcal Z}
\rho_\phi(z)\log\underline\rho_{\theta_\tau^*}\,\mu(\mathrm dz)
=
\log\underline\rho_{\theta_\tau^*}.
\]
Hence, for any $\rho\in\mathcal D_{\mu,\underline\rho_{\theta_\tau^*}}(\mathcal Z)$,
\begin{equation}
\label{eq:Ft_lower}
\mathcal F_\tau(\rho)
\ge
\frac12
\int_{\mathcal Z}
V_{\theta_\tau^*\mid\phi}(z)\rho(z)\,\mu(\mathrm dz)
+
\frac12\log\underline\rho_{\theta_\tau^*}
+
\mathrm{const}_\tau(\phi).
\end{equation}
Let
\[
\mathcal F_\tau^*
:=
\inf_{\rho\in\mathcal D_{\mu,\underline\rho_{\theta_\tau^*}}(\mathcal Z)}
\mathcal F_\tau(\rho).
\]
Applying \Cref{prop:mm_coordinate_extremal} with density floor $\underline\rho=\underline\rho_{\theta_\tau^*}$ to $\mathcal F_\tau$ yields, for each $\delta>0$, an infimizing competitor $\rho^{(\delta)}\in\mathcal D_{\mu,\underline\rho_{\theta_\tau^*}}(\mathcal Z)$ whose excess mass is supported on a shrinking set $\mathcal S^\delta\subseteq\mathcal W_{\theta_\tau^*}^\delta$ with $\mu(\mathcal S^\delta)\to0$ as $\delta\to0$, such that $\mathcal F_\tau(\rho^{(\delta)})\to\mathcal F_\tau^*$ as $\delta\to0$.

Moreover, since $\rho^{(\delta)}=\underline\rho_{\theta_\tau^*}$ on $\mathcal Z\setminus\mathcal S^\delta$ and $\rho_\phi$ is continuous on compact $\mathcal Z$, we have
\begin{align*}
\int_{\mathcal Z}
\rho_\phi(z)\log\rho^{(\delta)}(z)\,\mu(\mathrm dz)
&=
\log\underline\rho_{\theta_\tau^*}
+
\int_{\mathcal S^\delta}
\rho_\phi(z)
\log\!\left(
1+
\frac{
M_{\mathrm{ex}}^{(\theta_\tau^*)}
}{
\underline\rho_{\theta_\tau^*}\mu(\mathcal S^\delta)
}
\right)
\,\mu(\mathrm dz)
\\
&\le
\log\underline\rho_{\theta_\tau^*}
+
\|\rho_\phi\|_\infty
\mu(\mathcal S^\delta)
\log\!\left(
1+
\frac{c}{\mu(\mathcal S^\delta)}
\right)
\\
&=
\log\underline\rho_{\theta_\tau^*}
+
o_\delta(1),
\end{align*}
where $o_\delta(1)\to0$ as $\delta\to0$ because $\mu(\mathcal S^\delta)\log(1/\mu(\mathcal S^\delta))\to0$. In addition, since $\rho^{(\delta)}-\underline\rho_{\theta_\tau^*}$ places all excess mass inside $\{z\in\mathcal Z:V_{\theta_\tau^*\mid\phi}(z)\le v_*+\delta\}$,
\begin{align*}
\int_{\mathcal Z}
V_{\theta_\tau^*\mid\phi}(z)
\rho^{(\delta)}(z)\,\mu(\mathrm dz)
&=
\underline\rho_{\theta_\tau^*}
\int_{\mathcal Z}
V_{\theta_\tau^*\mid\phi}(z)\,\mu(\mathrm dz)
+
M_{\mathrm{ex}}^{(\theta_\tau^*)}
\int_{\mathcal Z}
V_{\theta_\tau^*\mid\phi}(z)
\bar\rho^{(\delta)}(z)\,\mu(\mathrm dz)
\\
&\le
\underline\rho_{\theta_\tau^*}
\int_{\mathcal Z}
V_{\theta_\tau^*\mid\phi}(z)\,\mu(\mathrm dz)
+
M_{\mathrm{ex}}^{(\theta_\tau^*)}(v_*+\delta),
\end{align*}
where
\(
\bar\rho^{(\delta)}
:=
{
(\rho^{(\delta)}-\underline\rho_{\theta_\tau^*})
}/{
M_{\mathrm{ex}}^{(\theta_\tau^*)}
}.
\)
Plugging these bounds into \Cref{eq:Ft_decomp} and letting $\delta\to0$ yields
\begin{equation}
\label{eq:Fstar_upper}
\mathcal F_\tau^*
\le
\frac12
\left(
\underline\rho_{\theta_\tau^*}
\int_{\mathcal Z}
V_{\theta_\tau^*\mid\phi}(z)\,\mu(\mathrm dz)
+
M_{\mathrm{ex}}^{(\theta_\tau^*)}v_*
\right)
+
\frac12\log\underline\rho_{\theta_\tau^*}
+
\mathrm{const}_\tau(\phi).
\end{equation}
Now apply \Cref{eq:Ft_lower} to $\rho=\rho_{\theta_\tau^*}$ and subtract \Cref{eq:Fstar_upper}. Using
\[
\int_{\mathcal Z}
V_{\theta_\tau^*\mid\phi}(z)
\rho_{\theta_\tau^*}(z)\,\mu(\mathrm dz)
=
\underline\rho_{\theta_\tau^*}
\int_{\mathcal Z}
V_{\theta_\tau^*\mid\phi}(z)\,\mu(\mathrm dz)
+
M_{\mathrm{ex}}^{(\theta_\tau^*)}
\int_{\mathcal Z}
V_{\theta_\tau^*\mid\phi}(z)
\bar\rho_{\theta_\tau^*}(z)\,\mu(\mathrm dz),
\]
we obtain
\begin{equation}
\label{eq:gap_controls_EV}
\mathcal F_\tau(\rho_{\theta_\tau^*})
-
\mathcal F_\tau^*
\ge
\frac{M_{\mathrm{ex}}^{(\theta_\tau^*)}}{2}
\int_{\mathcal Z}
\big(
V_{\theta_\tau^*\mid\phi}(z)-v_*
\big)
\bar\rho_{\theta_\tau^*}(z)\,\mu(\mathrm dz).
\end{equation}

\textbf{Step 2: Use expressivity and convert to leakage.}
By the expressivity assumption in \Cref{eq:mm_coord_gap},
\[
\mathcal F_\tau(\rho_{\theta_\tau^*})
\le
\mathcal F_\tau^*
+
\varepsilon_{\mathrm{real}}(\tau).
\]
Combining this with \Cref{eq:gap_controls_EV} gives
\begin{equation}
\label{eq:EV_bound}
\int_{\mathcal Z}
\big(
V_{\theta_\tau^*\mid\phi}(z)-v_*
\big)
\bar\rho_{\theta_\tau^*}(z)\,\mu(\mathrm dz)
\le
\frac{
2\,\varepsilon_{\mathrm{real}}(\tau)
}{
M_{\mathrm{ex}}^{(\theta_\tau^*)}
}.
\end{equation}
Finally, since $V_{\theta_\tau^*\mid\phi}(z)-v_*\ge\sigma$ $\mu$-a.e.\ on $\mathcal Z\setminus\mathcal W^\sigma_{\theta_\tau^*}$, we have
\[
\begin{aligned}
\int_{\mathcal Z}
\big(
V_{\theta_\tau^*\mid\phi}(z)-v_*
\big)
\bar\rho_{\theta_\tau^*}(z)\,\mu(\mathrm dz)
&\ge
\int_{\mathcal Z\setminus\mathcal W^\sigma_{\theta_\tau^*}}
\big(
V_{\theta_\tau^*\mid\phi}(z)-v_*
\big)
\bar\rho_{\theta_\tau^*}(z)\,\mu(\mathrm dz)
\\
&\ge
\sigma
\int_{\mathcal Z\setminus\mathcal W^\sigma_{\theta_\tau^*}}
\bar\rho_{\theta_\tau^*}(z)\,\mu(\mathrm dz)
\\
&=
\sigma\,
\bar\rho_{\theta_\tau^*}\mu
\big(
\mathcal Z\setminus\mathcal W^\sigma_{\theta_\tau^*}
\big),
\end{aligned}
\]
and therefore, by \Cref{eq:EV_bound},
\[
\bar\rho_{\theta_\tau^*}\mu
\big(
\mathcal Z\setminus\mathcal W^\sigma_{\theta_\tau^*}
\big)
\le
\frac{
2\,\varepsilon_{\mathrm{real}}(\tau)
}{
\sigma M_{\mathrm{ex}}^{(\theta_\tau^*)}
},
\]
which is exactly \Cref{eq:mm_excess_leakage}. The concluding asymptotic statement follows immediately: by the uniform lower bound $M_{\mathrm{ex}}^{(\theta_\tau^*)}\ge m_{\mathrm{ex}}>0$, if $\sigma=\sigma(\tau)\downarrow0$ and $\varepsilon_{\mathrm{real}}(\tau)/\sigma(\tau)\to0$, then $\bar\rho_{\theta_\tau^*}\mu\bigl(\mathcal Z\setminus\mathcal W_{\theta_\tau^*}^{\sigma(\tau)}\bigr)\to0$ as $\tau\to0^+$.
\end{proof}

\section{Experimental Details}
\label{app:extend_exp}
This section details the experimental procedures used to empirically validate our theoretical framework. Specifically, we provide comprehensive setups for evaluating: (i) the convergence of finite-batch InfoNCE gradients to their deterministic counterparts; (ii) the Gibbs-shaped concentration and spread behavior of unimodal training under sharp kernels; (iii) the emergence of a persistent, population-level modality gap induced by conditional heterogeneity; and (iv) the real-world implications of cross-modal divergence via probing and fine-tuning CLIP models on MS-COCO.

\subsection{Large-Batch Gradient Consistency Across Similarity Functions}
\label{app:grad_consist_exp}

This experiment provides a numerical illustration of the large-batch consistency established in \Cref{thm:nce_consistency}. We verify that as the number of negatives $N$ increases, the finite-batch InfoNCE gradient aligns increasingly well with a high-fidelity large-$N$ reference gradient. We evaluate this behavior under two different similarity functions and embedding geometries. The corresponding large-batch consistency for the full symmetric multimodal objective is established separately in \Cref{thm:sym_consistency}.

\paragraph{Data and pairing.}
We construct a synthetic unimodal dataset in $\mathcal X\subset\mathbb R^m$ ($m=64$) from a mixture of $K=4$ Gaussians:
\[
\mathbf x
\sim
\sum_{k=1}^K
\pi_k
\mathcal N
\left(
\boldsymbol\mu_k,
\sigma^2\mathbf I
\right).
\]
The component means $\{\boldsymbol\mu_k\}$ are random directions scaled by a separation factor of $4$, with $\sigma=1$. Positive pairs are generated via additive noise $\tilde{\mathbf x}=\mathbf x+\boldsymbol\varepsilon$, where $\boldsymbol\varepsilon\sim\mathcal N(\mathbf 0,\sigma_{\mathrm{aug}}^2\mathbf I)$. Negative samples $\{\mathbf x_j^-\}_{j=1}^N$ are i.i.d.\ draws from the same data distribution.

\paragraph{Encoder and regimes.}
We employ a linear encoder $f_{\mathbf W}(x)$ with parameter $\mathbf W\in\mathbb R^{n\times m}$ ($n=128$) under two distinct geometric regimes:
\[
z
=
f_{\mathbf W}(x)
=
\begin{cases}
\mathrm{normalize}(\mathbf W x)\in\mathbb S^{n-1},
& \text{(Spherical regime)},\\
\tanh(\mathbf W x)\in[-1,1]^n,
& \text{(Compact Euclidean regime)}.
\end{cases}
\]
The spherical case corresponds to the standard feature-normalized setting, while the compact Euclidean case provides a compact non-normalized embedding domain suitable for evaluating distance-based similarities. The latter is intended to test the broader partition-field mechanism numerically rather than to instantiate the constant kernel-volume assumption exactly.

We evaluate the exponential kernel defined in \Cref{sec:setup} with two canonical similarity functions: the cosine similarity
\[
\mathrm{sim}_{\mathrm{cos}}(z,w)
=
\langle z,w\rangle
\]
and the RBF similarity
\[
\mathrm{sim}_{\mathrm{rbf}}(z,w)
=
-\frac{1}{n}\|z-w\|^2,
\]
with temperatures $\tau_{\mathrm{cos}}=0.1$ and $\tau_{\mathrm{rbf}}=1.0$, respectively. The RBF similarity includes the factor $1/n$ to maintain $O(1)$ logits. We set augmentation noise $\sigma_{\mathrm{aug}}=0.05$ for the cosine regime and $\sigma_{\mathrm{aug}}=0.2$ for the RBF regime.

\paragraph{Loss and gradients.}
For each seed, we sample a fixed batch of $B=64$ anchors/positives and a large ``reference'' pool of $N_{\mathrm{ref}}=4096$ negatives. For a given $N$, we compute the loss $\hat\ell_{B,N}(\mathbf W)$ using the first $N$ negatives from this pool:
\[
\hat\ell_{B,N}(\mathbf W)
=
\frac{1}{B}
\sum_{i=1}^B
\left[
-\frac{1}{\tau}
\mathrm{sim}(z_i,\tilde z_i)
+
\log
\left(
\exp\!\left(
\frac{\mathrm{sim}(z_i,\tilde z_i)}{\tau}
\right)
+
\sum_{j=1}^N
\exp\!\left(
\frac{\mathrm{sim}(z_i,w_j^-)}{\tau}
\right)
\right)
\right].
\]
Let
\[
\mathsf g_N
:=
\nabla_{\mathbf W}
\hat\ell_{B,N}(\mathbf W)
\]
denote the gradient estimate using $N$ negatives, and let
\[
\mathsf g_{\mathrm{ref}}
:=
\mathsf g_{N_{\mathrm{ref}}}
\]
denote the large-$N$ reference gradient. Fixing the negative pool ensures that variations in $\mathsf g_N$ across $N$ arise from the number of included negatives rather than from resampling the negative pool.

\paragraph{Metrics and results.}
We sweep $N\in\{4,8,16,32,64,128,256,512,1024\}$ and report two metrics: (i) \emph{gradient alignment} $\cos(\mathsf g_N,\mathsf g_{\mathrm{ref}})$, measuring directional agreement; and (ii) \emph{relative error} $\|\mathsf g_N-\mathsf g_{\mathrm{ref}}\|/\|\mathsf g_{\mathrm{ref}}\|$, measuring magnitude deviation. Results are averaged over 20 random seeds; error bars indicate $\pm1$ standard deviation across seeds. As in \Cref{fig:grad_alignment,fig:grad_relerr}, across both the spherical and RBF regimes, the gradient alignment increases with $N$ while the relative error decreases. These results are consistent with the large-batch gradient-consistency theory: as the negative partition becomes better approximated by a large sample, the finite-$N$ gradient approaches its large-$N$ reference. The rapid convergence (alignment $>0.95$ at $N=256$) further suggests that the large-batch gradient approximation can already be accurate at moderate numbers of negatives in these controlled settings.

\subsection{Unimodal Gibbs Equilibrium and Low-Temperature Concentration}
\label{app:uni_s2_exp}

This experiment provides a numerical illustration of the intrinsic unimodal geometry characterized in \Cref{sec:unimodal}. Specifically, we examine two predictions of the variational perspective: (i) for fixed $\tau>0$, the functional $\mathcal F_{\tau,U}$ admits the unique Gibbs equilibrium characterized in \Cref{prop:gibbs_equilibrium}; and (ii) as $\tau\to0^+$, this equilibrium concentrates around the low-potential regions of $U$, as predicted by the ground-state concentration result in \Cref{prop:ground_state}.

\begin{figure*}[t]
    \centering
    \includegraphics[width=\textwidth]{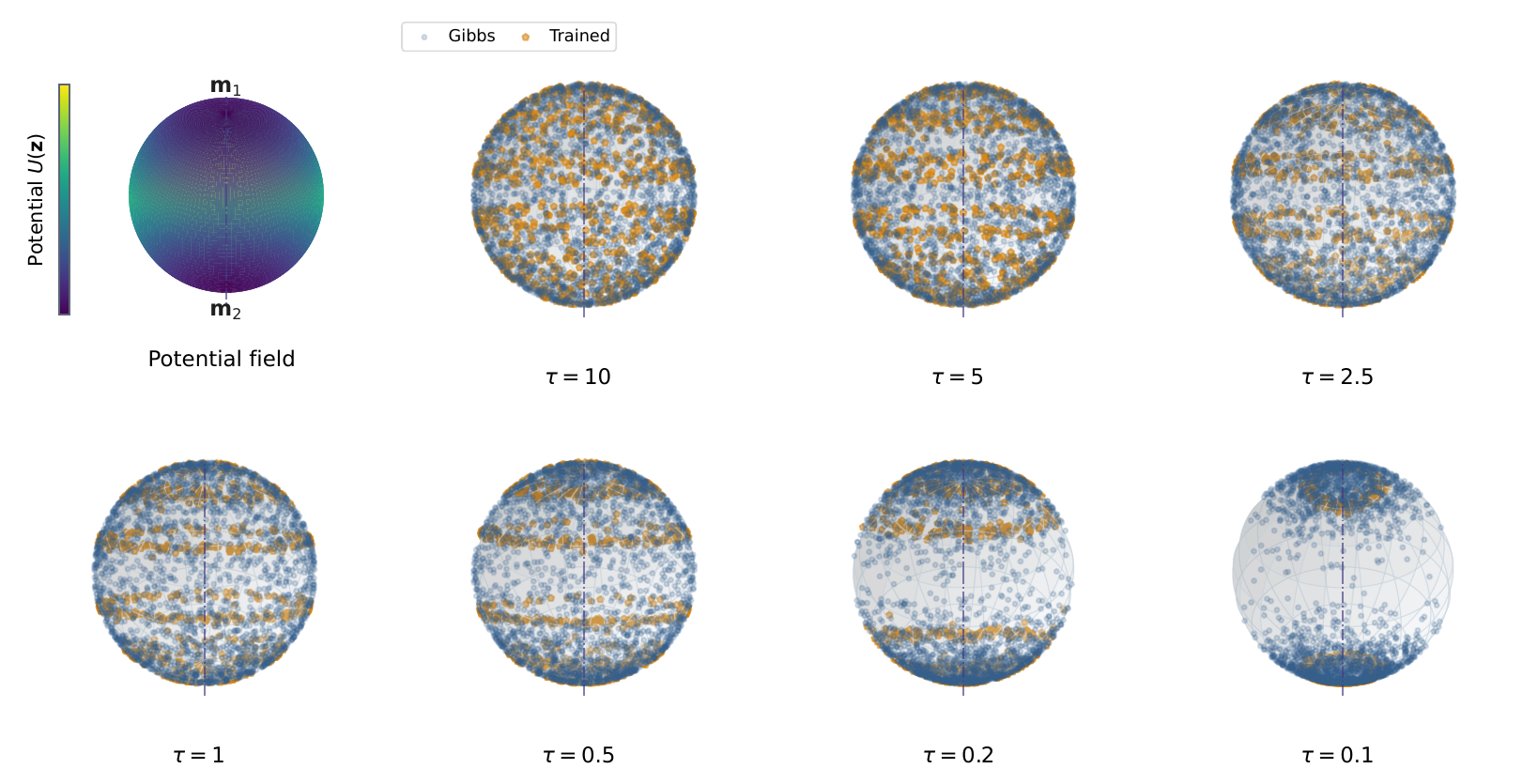}
    \caption{Unimodal potential landscape on $\mathbb S^2$ and distributions across temperature $\tau$. Left top: the two-well potential $U$ in \Cref{eq:uni_s2_potential} (colored by value), with basin centers $\mathbf m_1,\mathbf m_2$. Others: Gibbs samples (blue; importance-resampled) and trained particles (orange; minimizing \Cref{eq:uni_s2_particle_obj}) across temperatures. As $\tau$ decreases, both distributions concentrate around the low-energy basins.}
    \label{fig:uni_s2_overlays}
\end{figure*}

\paragraph{Representation space and potential field.}
We define the domain as the unit sphere $\mathcal Z=\mathbb S^2\subset\mathbb R^3$ equipped with surface measure $\mu$ and geodesic distance $d_{\mathbb S^2}$. We construct a smooth, non-isotropic potential $U(z)$ using a log-sum-exp mixture of two von-Mises--Fisher-like modes:
\begin{equation}
\label{eq:uni_s2_potential}
U(z)
:=
-\frac{1}{\gamma}
\log\!\left(
w\,\exp\!\big(\kappa\langle z,\mathbf m_1\rangle\big)
+
(1-w)\,\exp\!\big(\kappa\langle z,\mathbf m_2\rangle\big)
\right),
\qquad
z\in\mathbb S^2,
\end{equation}
with $\gamma=12$, concentration $\kappa=12$, and mixture weight $w=0.5$. The modes are centered at $\mathbf m_1=(0,0,1)$ and the separated direction $\mathbf m_2={(0.85,0.15,-0.50)}/{\|(0.85,0.15,-0.50)\|}$, creating two distinct low-energy basins (\Cref{fig:uni_s2_overlays}, left top).

\paragraph{Particle training objective.}
To emulate minimization of
\[
\mathcal F_{\tau,U}(\rho)
=
\frac{1}{\tau}
\int_{\mathcal Z}
U(z)\rho(z)\,
\mu(\mathrm dz)
-
H(\rho)
\]
(\Cref{def:uni_free_energy}) without committing to a parametric encoder, we represent $\rho$ by $M$ particles $\{\mathbf z_i\}_{i=1}^M\subset\mathbb S^2$ and optimize a standard particle/KDE surrogate. We parameterize unconstrained variables $\{\mathbf v_i\}_{i=1}^M\subset\mathbb R^3$ and project them to the sphere via $\mathbf z_i=\mathbf v_i/\|\mathbf v_i\|$ at each iteration. We approximate the density at particle locations using a geodesic Gaussian kernel on the sphere,
\begin{equation}
\label{eq:uni_s2_kde}
\hat\rho_h(\mathbf z_i)
:=
\frac{1}{M}
\sum_{j=1}^M
\exp\!\left(
-\frac{
d_{\mathbb S^2}(\mathbf z_i,\mathbf z_j)^2
}{
2h^2
}
\right),
\end{equation}
with bandwidth $h=0.35$. The optimized objective is the particle approximation
\begin{equation}
\label{eq:uni_s2_particle_obj}
\widehat{\mathcal F}_{\tau,U}(\{\mathbf z_i\})
:=
\frac{1}{\tau}\cdot\frac{1}{M}
\sum_{i=1}^M
U(\mathbf z_i)
+
\frac{1}{M}
\sum_{i=1}^M
\log\hat\rho_h(\mathbf z_i),
\end{equation}
where the second term is a KDE-based surrogate for $-H(\rho)$ up to an additive constant. We minimize \Cref{eq:uni_s2_particle_obj} using Adam \citep{kingma2014adam} for $5000$ steps with learning rate $5\times10^{-2}$, while injecting Gaussian noise from $\mathcal N(\mathbf 0,\sigma^2\mathbf I)$ with $\sigma=0.06$ at each iteration. This Langevin-inspired perturbation encourages continued exploration of the landscape and reduces sensitivity to shallow local traps, while the KDE entropy term promotes dispersion among the particles.

We sweep $\tau\in\{10,5.0,2.5,1.0,0.5,0.2,0.1\}$ and run $20$ seeds. As a reference, we estimate the Gibbs equilibrium
\[
\rho_\tau^\star(z)
\propto
\exp\!\left(
-\frac{U(z)}{\tau}
\right)
\]
(\Cref{prop:gibbs_equilibrium}) by importance sampling: we draw $n_{\mathrm{mc}}=120{,}000$ uniform points on $\mathbb S^2$, weight them by $\exp(-U/\tau)$, and compute weighted expectations. For visualization, we use a separate pool of $24{,}000$ uniform points and perform weighted sampling without replacement to draw $2400$ Gibbs points per $\tau$.

\paragraph{Evaluation metric.}
To quantify low-$\tau$ concentration, we report the \emph{cap mass} within geodesic radius $\varepsilon=0.50$ around either potential well:
\[
\mathrm{CapMass}_\varepsilon(\rho)
:=
(\rho\mu)
\left(
\mathcal C_\varepsilon(\mathbf m_1)
\cup
\mathcal C_\varepsilon(\mathbf m_2)
\right),
\]
where
\[
\mathcal C_\varepsilon(\mathbf m)
:=
\left\{
z\in\mathbb S^2:
d_{\mathbb S^2}(z,\mathbf m)
\le
\varepsilon
\right\}.
\]
For the trained particles, this is simply the fraction of particles whose distance to $\{\mathbf m_1,\mathbf m_2\}$ is at most $\varepsilon$. For the Gibbs baseline, we compute the same quantity as a weighted Monte Carlo estimate. We report the mean $\pm$ standard error over the $20$ seeds.

\paragraph{Results.}
\Cref{fig:uni_s2_overlays} visualizes the potential landscape and the particle distributions across $\tau$ together with the corresponding Gibbs reference. At high temperature (e.g., $\tau=10$), both the Gibbs samples and the trained particles spread broadly over the sphere: the entropy term in $\mathcal F_{\tau,U}$ is comparatively strong, producing a diffuse distribution consistent with entropic dispersion. As $\tau$ decreases (e.g., $\tau=0.2,\tau=0.1$), the potential term $\tfrac{1}{\tau}\int_{\mathcal Z}U(z)\rho(z)\,\mu(\mathrm dz)$ becomes increasingly influential, and both distributions concentrate near the two low-energy basins. Notably, the trained particles exhibit semi-regular lattice-like bands, in contrast to the random scatter of the Gibbs samples. This micro-structure is consistent with the repulsion induced by the KDE entropy surrogate, which encourages the finite particle set to spread within the occupied regions. Such finite-particle organization does not alter the macroscopic prediction: at $\tau=0.1$, the global mass is visibly localized around the low-energy basins, in qualitative agreement with ground-state concentration (\Cref{prop:ground_state}).

This trend is quantified in \Cref{fig:ground_state}: the cap mass $\mathrm{CapMass}_\varepsilon(\rho)$ increases as $\tau$ decreases for both the Gibbs baseline and the trained particles. The trained curve closely tracks the same trend, with small deviations attributable to the finite particle count $M$, finite optimization time, and fixed KDE bandwidth $h$, which imposes a smoothing scale on the entropy surrogate. Overall, the experiment supports the unimodal geometric picture: the Gibbs reference exhibits systematic low-temperature concentration near low-potential regions of $U$, while optimization of the particle free-energy surrogate qualitatively tracks this behavior across temperatures.

\subsection{Multimodal Structural Gap under Controlled Misalignment}
\label{app:mm_gap_exp}

This experiment provides a controlled empirical illustration of the \emph{multimodal} mechanism developed in \Cref{sec:multimodal}, motivated by the well-documented \emph{modality gap} phenomenon observed in CLIP-style training \citep{liang2022mind}. Our theory shows that although symmetric multimodal InfoNCE admits a stable large-batch deterministic limit, its intrinsic landscape is \emph{cross-coupled}, with each coordinate governed by an effective potential that depends on the peer marginal. When the two modalities exhibit heterogeneous conditional laws---arising from noise, distinct semantic density, or mismatched semantic structure---these directional forces need not be simultaneously compatible with coincident marginals and can sustain a persistent separation. We illustrate this mechanism in a controlled setting by progressively introducing structured cross-modal misalignment and measuring the resulting marginal gap.

\paragraph{Latent variables and paired observations.}
We generate paired observations $(\mathbf{x}_1,\mathbf{x}_2)$ representing two modalities that stem from a shared latent angle $\alpha\in(-\pi,\pi]$. To rule out the trivial solution where marginals are uniform (and thus trivially match regardless of noise), we sample the shared latent $\alpha$ from a \emph{non-uniform}, multi-peaked mixture of von Mises distributions:
\[
\alpha
\sim
w\,\mathrm{VM}(\mu_1,\kappa)
+
(1-w)\,\mathrm{VM}(\mu_2,\kappa),
\qquad
\text{with}\;\;
w=0.7,\quad
\mu_1=0,\quad
\mu_2=\pi,\quad
\kappa=6.
\]
Both modalities observe this angle as a projection onto the 2D unit circle, corrupted by independent Gaussian observation noise $\boldsymbol{\xi}_1,\boldsymbol{\xi}_2\sim\mathcal N(\mathbf 0,\sigma_{\mathrm{obs}}^2\mathbf I)$, where $\sigma_{\mathrm{obs}}=0.02$. Modality 1 is generated by mapping the shared latent directly to the observation space:
\[
\mathbf{x}_1
:=
[\cos\alpha,\ \sin\alpha]^\top
+
\boldsymbol{\xi}_1.
\]
Modality 2 observes a \emph{misaligned} version of the latent. We introduce an additive angular shift $\eta\sim\mathcal N(0,\sigma_{\mathrm{mis}}^2)$ before mapping it to the observation space:
\[
\mathbf{x}_2
:=
[\cos(\alpha+\eta),\ \sin(\alpha+\eta)]^\top
+
\boldsymbol{\xi}_2.
\]
Thus, increasing the misalignment scale $\sigma_{\mathrm{mis}}$ increases the \emph{cross-modal conditional mismatch} between $\mathbf{x}_1$ and $\mathbf{x}_2$, while keeping the marginal complexity of both modalities comparable.

\begin{figure}[tbp]
    \centering
    \includegraphics[width=0.24\linewidth]{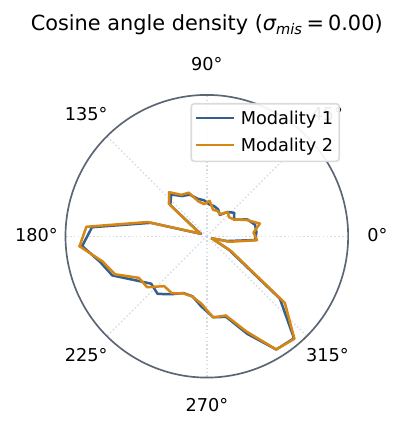}
    \hfill
    \includegraphics[width=0.24\linewidth]{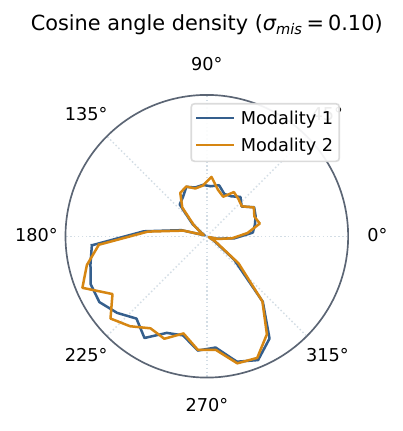}
    \hfill
    \includegraphics[width=0.24\linewidth]{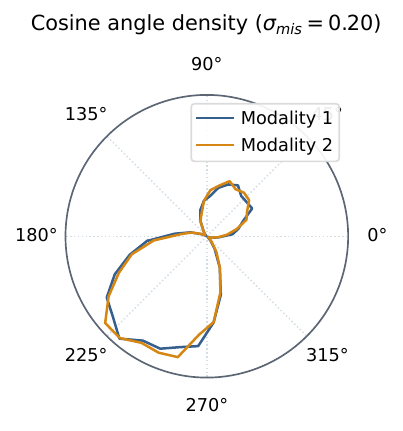}
    \hfill
    \includegraphics[width=0.24\linewidth]{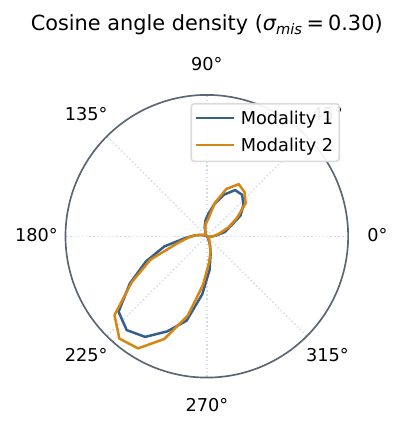}
    \vspace{2mm}
    \includegraphics[width=0.24\linewidth]{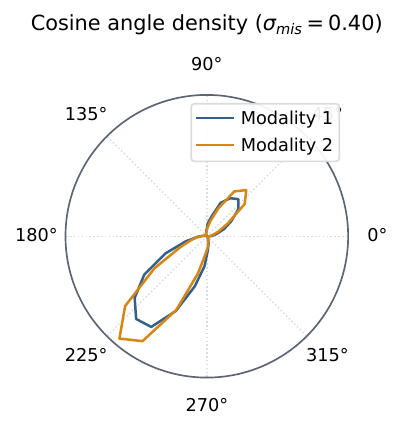}
    \hfill
    \includegraphics[width=0.24\linewidth]{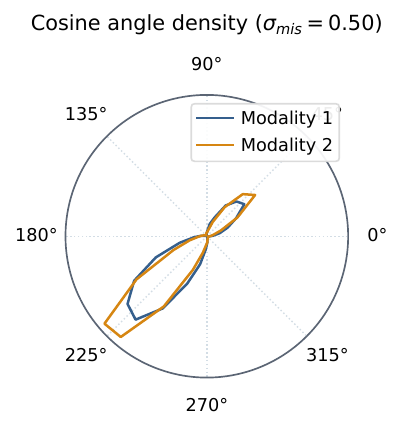}
    \hfill
    \includegraphics[width=0.24\linewidth]{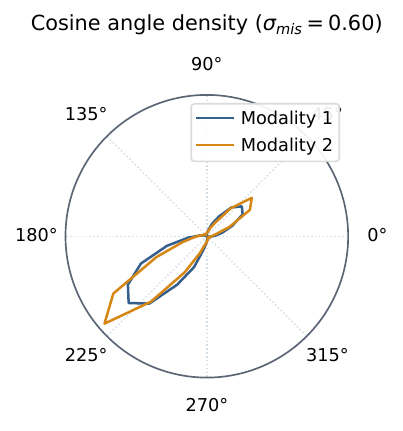}
    \hfill
    \includegraphics[width=0.24\linewidth]{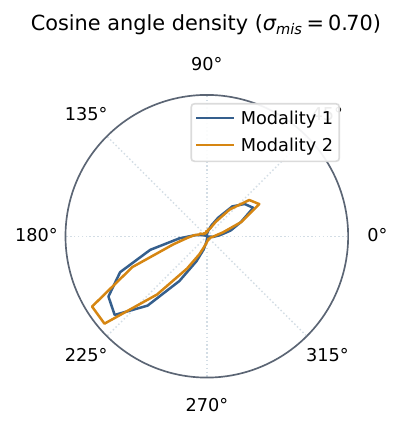}
    \caption{Polar marginals across misalignment. Estimated angle marginals of the two modalities on $\mathbb S^1$. The mismatch becomes increasingly apparent as $\sigma_{\mathrm{mis}}$ grows.}
    \label{fig:mm_polar_grid}
\end{figure}

\paragraph{Encoders, learning objective, and protocol.}
To isolate the directional alignment geometry and prevent magnitude from confounding the representation geometry, we apply bias-free linear encoders $f,g:\mathbb R^2\to\mathbb R^2$ to the observed modalities. The unnormalized embeddings $\tilde{\mathbf z}_1=f(\mathbf x_1)$ and $\tilde{\mathbf z}_2=g(\mathbf x_2)$ are then projected onto the unit circle:
\[
\mathbf z_1
=
\frac{\tilde{\mathbf z}_1}{\|\tilde{\mathbf z}_1\|},
\qquad
\mathbf z_2
=
\frac{\tilde{\mathbf z}_2}{\|\tilde{\mathbf z}_2\|}
\in
\mathbb S^1.
\]
We adopt the cosine similarity $\mathrm{sim}(z,w)=\langle z,w\rangle$ and the corresponding exponential kernel $\kappa_\tau(z,w)=\exp(\mathrm{sim}(z,w)/\tau)$ with temperature $\tau=0.07$. This kernel directly instantiates the InfoNCE objective central to our multimodal analysis. Specifically, for a batch of $N+1$ paired observations, training minimizes the symmetric CLIP loss:
\[
\mathcal L_{\mathrm{CLIP}}(f,g)
=
\frac12
\left(
\mathcal L_{\mathrm{CE}}
\big(
\mathbf S/\tau,
\mathbf y_{\mathrm{diag}}
\big)
+
\mathcal L_{\mathrm{CE}}
\big(
\mathbf S^\top/\tau,
\mathbf y_{\mathrm{diag}}
\big)
\right),
\]
where \(S\in\mathbb R^{(N+1)\times (N+1)}\) is the pairwise similarity matrix with entries $\mathbf S_{ij}:=\langle\mathbf z_{1,i},\mathbf z_{2,j}\rangle$, $\mathcal L_{\mathrm{CE}}$ denotes the standard cross-entropy loss, and the target $\mathbf y_{\mathrm{diag}}$ indicates that the matched positive pairs lie on the diagonal ($i=j$).

We use Adam with learning rate $5\times10^{-3}$ for $2000$ steps and batch size $256$. We sweep eight misalignment scales $\sigma_{\mathrm{mis}}\in\{0,0.1,0.2,0.3,0.4,0.5,0.6,0.7\}$; larger scales induce progressively more severe cross-modal misalignment, yet all settings retain meaningful statistical dependence between positive pairs. For each $\sigma_{\mathrm{mis}}$, we repeat training over $20$ random seeds (seeds $0$--$19$). The large batch size makes the finite-batch objective and gradients better approximate the large-batch regime characterized in \Cref{thm:sym_consistency}.

\paragraph{Evaluation metrics.}
After training, we evaluate the learned representations on $n_{\mathrm{eval}}=8000$ fresh paired samples. For the unit-normalized embeddings $\mathbf z_1,\mathbf z_2\in\mathbb S^1$, we compute the embedding angles $a_1=\mathrm{atan2}(\mathbf z_{1,y},\mathbf z_{1,x})$ and $a_2=\mathrm{atan2}(\mathbf z_{2,y},\mathbf z_{2,x})$, where the subscripts $x$ and $y$ denote the respective coordinate axes. The following visual diagnostics are considered:

\emph{(i) Marginal gap.}
We discretize $a_1$ and $a_2$ into $60$ bins on $(-\pi,\pi]$ and estimate the two empirical angle densities $\hat\rho_\theta$ and $\hat\rho_\phi$. As a scalar summary of marginal mismatch, we report the symmetric KL divergence $D_{\mathrm{KL}}^{\mathrm{Sym}}(\hat\rho_\theta,\hat\rho_\phi)$ averaged over $20$ random seeds (seeds $0$--$19$), with error bars or shaded areas showing $\pm$ standard error.

\emph{(ii) Polar marginals.}
To visualize \emph{where} the mismatch occurs on the circle, we plot the two estimated angle marginals as polar density curves for each misalignment scale $\sigma_{\mathrm{mis}}$.

\emph{(iii) Joint-angle heatmaps.}
To visualize the learned \emph{cross-modal coupling} beyond marginals, we plot a log-scaled 2D histogram, i.e., a $\log(1+\text{count})$ heatmap, of the joint distribution of $(a_1,a_2)$. Diagonal concentration indicates near-deterministic alignment $a_2\approx a_1$; off-diagonal spread indicates increasing cross-modal mismatch.

\emph{(iv) Angle-shift density.}
Finally, we compute the wrapped angular difference $\Delta a:=\mathrm{wrap}(a_2-a_1)\in(-\pi,\pi]$ and plot its empirical density. This directly measures the distribution of residual misalignment in embedding space.

For the scalar marginal-gap curve, we average over $20$ random seeds and report the mean with a $\pm$SEM band. For the qualitative diagnostics (polar marginals, joint-angle heatmaps, and angle-shift densities), we display one fixed representative run (seed 0) and reuse the same held-out latent sample across all misalignment scales.

\begin{figure}[tbp]
    \centering
    \includegraphics[width=0.24\linewidth]{plots/multimodal_num/mm_joint_cosine_sig0.00.pdf}
    \hfill
    \includegraphics[width=0.24\linewidth]{plots/multimodal_num/mm_joint_cosine_sig0.10.pdf}
    \hfill
    \includegraphics[width=0.24\linewidth]{plots/multimodal_num/mm_joint_cosine_sig0.20.pdf}
    \hfill
    \includegraphics[width=0.24\linewidth]{plots/multimodal_num/mm_joint_cosine_sig0.30.pdf}
    \vspace{2mm}
    \includegraphics[width=0.24\linewidth]{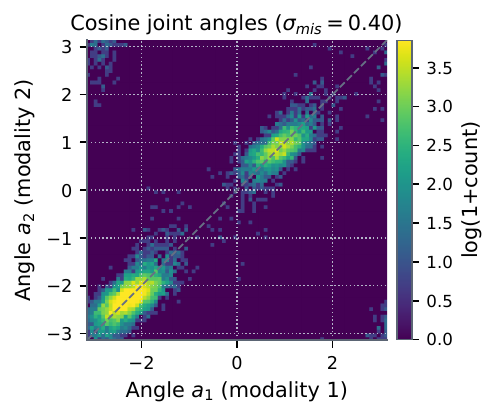}
    \hfill
    \includegraphics[width=0.24\linewidth]{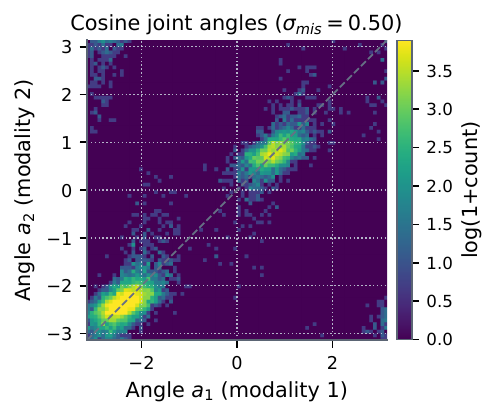}
    \hfill
    \includegraphics[width=0.24\linewidth]{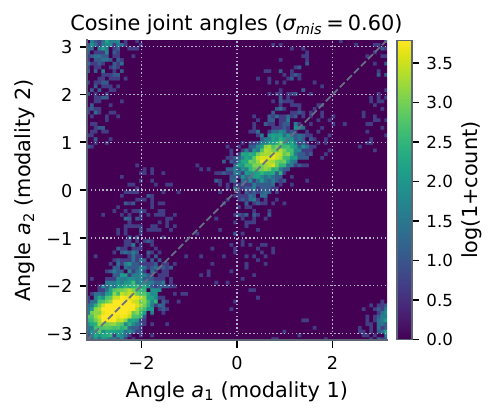}
    \hfill
    \includegraphics[width=0.24\linewidth]{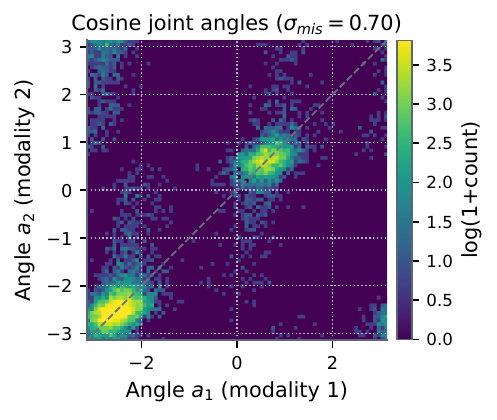}
    \caption{Joint-angle coupling across misalignment. 2D histograms of $(a_1,a_2)$ from $\sigma_{\mathrm{mis}}=0.0$ to $\sigma_{\mathrm{mis}}=0.7$. Diagonal concentration at small $\sigma_{\mathrm{mis}}$ indicates near-deterministic alignment, while increasing off-diagonal spread at larger $\sigma_{\mathrm{mis}}$ indicates increasingly noisy cross-modal coupling.}
    \label{fig:mm_joint_grid}
\end{figure}

\paragraph{Results.}
\Cref{fig:divergence_persist_main} shows that the marginal discrepancy $D_{\mathrm{KL}}^{\mathrm{Sym}}(\hat\rho_\theta,\hat\rho_\phi)$ increases as the latent misalignment scale $\sigma_{\mathrm{mis}}$ grows. Since the encoder architecture, similarity function, and optimization protocol remain fixed across all runs, the controlled perturbation $\sigma_{\mathrm{mis}}$ isolates the role of conditional mismatch in this setting. The result is consistent with the theoretical mechanism: as the two modalities become less conditionally compatible, maintaining both pairwise alignment and coincident marginals becomes increasingly difficult under the symmetric objective, and a persistent marginal separation emerges.

The polar marginals in \Cref{fig:mm_polar_grid} provide a structural view of \emph{how} this separation manifests geometrically. First, across all settings ($\sigma_{\mathrm{mis}}\ge0$), the learned angular marginals are clearly \emph{multi-peaked} rather than uniform. The density concentrates in specific angular regions, reflecting the underlying non-uniform latent law and showing that the learned representations retain its multi-modal structure. Second, the evolution of these peaks reveals the gap mechanism. At $\sigma_{\mathrm{mis}}=0$, the polar curves for the two modalities nearly overlap, consistent with a compatible pairing law where symmetric InfoNCE can simultaneously achieve alignment and marginal coincidence. As $\sigma_{\mathrm{mis}}$ increases from zero, the curves begin to diverge within the same modal regions: the peaks shift relative to each other, and their probability masses differ. This mismatch becomes visually more pronounced as $\sigma_{\mathrm{mis}}$ increases, tracking the quantitative rise in the gap curve.

The joint-angle heatmaps in \Cref{fig:mm_joint_grid} elucidate the mechanism behind this separation. At $\sigma_{\mathrm{mis}}=0$, probability mass concentrates sharply along the diagonal $a_2\approx a_1$, indicating a near-deterministic coupling. Because the latent distribution is bimodal, this diagonal concentration is not uniform but forms distinct bright spots (high-density clusters) corresponding to the latent modes. As $\sigma_{\mathrm{mis}}$ increases, these clusters persist---indicating that the model still captures the coarse latent structure---but they deform into thickened, elongated regions. The diagonal band broadens significantly, reflecting increasing uncertainty in the conditional coupling between $a_1$ and $a_2$. The accompanying changes in marginal occupancy observed in \Cref{fig:mm_polar_grid} show that, in this controlled setting, broader conditional mismatch is accompanied by increasing marginal separation.

\begin{figure}[tbp]
    \centering
    \includegraphics[width=0.24\linewidth]{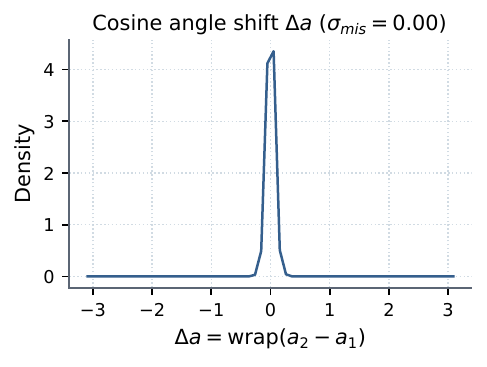}
    \hfill
    \includegraphics[width=0.24\linewidth]{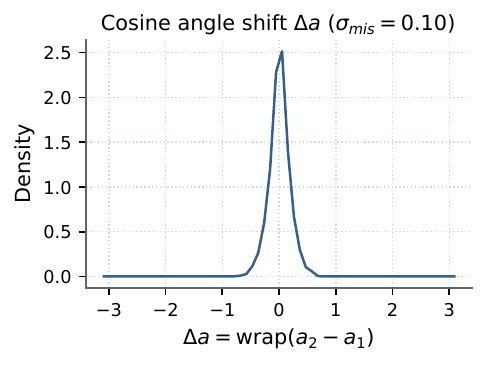}
    \hfill
    \includegraphics[width=0.24\linewidth]{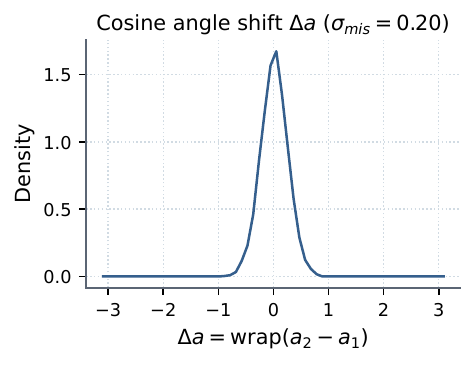}
    \hfill
    \includegraphics[width=0.24\linewidth]{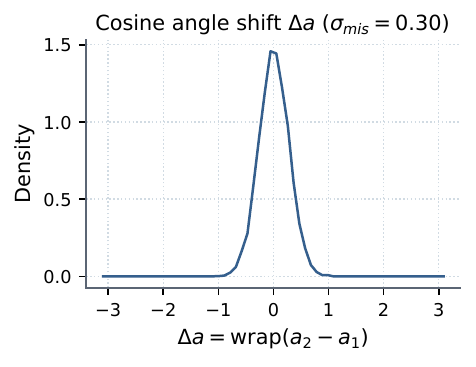}
    \vspace{2mm}
    \includegraphics[width=0.24\linewidth]{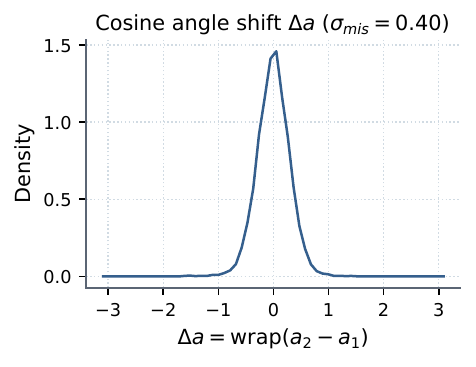}
    \hfill
    \includegraphics[width=0.24\linewidth]{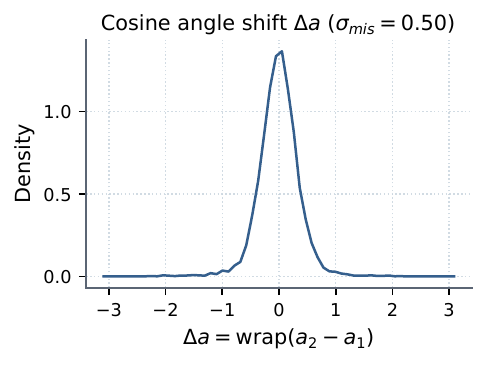}
    \hfill
    \includegraphics[width=0.24\linewidth]{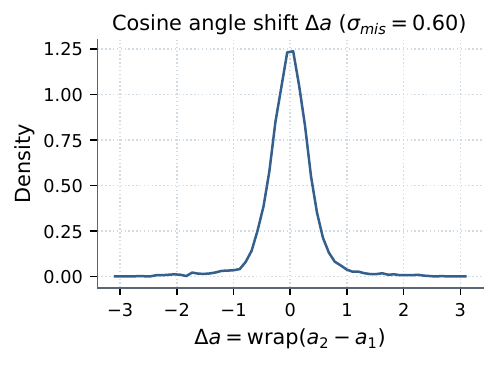}
    \hfill
    \includegraphics[width=0.24\linewidth]{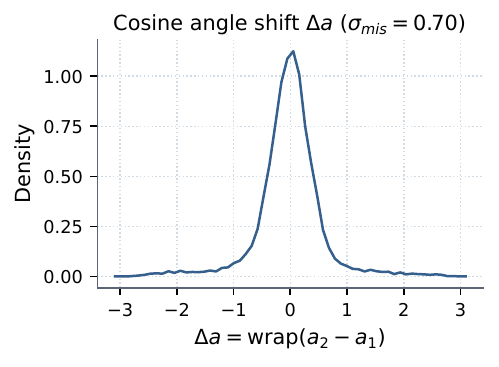}
    \caption{Residual embedding-space misalignment. Densities of $\Delta a=\mathrm{wrap}(a_2-a_1)$ from $\sigma_{\mathrm{mis}}=0.0$ to $\sigma_{\mathrm{mis}}=0.7$. The distribution broadens as $\sigma_{\mathrm{mis}}$ increases, indicating larger and more variable residual mismatch in embedding space.}
    \label{fig:mm_delta_grid}
\end{figure}

Finally, the angle-shift densities in \Cref{fig:mm_delta_grid} quantify the residual embedding misalignment $\Delta a=\mathrm{wrap}(a_2-a_1)$. As $\sigma_{\mathrm{mis}}$ increases, the distribution of $\Delta a$ transitions from a sharp peak near zero to a progressively wider distribution. The residual shift becomes both larger in magnitude and more variable, matching the qualitative signature expected from the injection of additive latent noise $\eta\sim\mathcal N(0,\sigma_{\mathrm{mis}}^2)$.

In summary, these empirical patterns provide a controlled illustration of the multimodal mechanism characterized in \Cref{sec:multimodal}. Our theoretical analysis shows that the symmetric objective induces two directional alignment fields coupled through their marginals, so heterogeneous conditional structure can make marginal coincidence unstable and sustain a separation-inducing contribution. In this experiment, the perturbation $\alpha+\eta$ progressively increases the mismatch between the two directional conditional laws. Empirically, this produces two related effects: (i) the broadening diagonal band in \Cref{fig:mm_joint_grid} reflects increasing uncertainty in the cross-modal coupling; and (ii) the growing marginal discrepancy in \Cref{fig:divergence_persist_main} and \Cref{fig:mm_polar_grid} shows that this conditional mismatch is accompanied by increasing separation between the learned modality marginals. Together, these observations are consistent with the intrinsic coordinate geometry developed in \Cref{sec:multimodal}: once strict cross-modal compatibility is weakened, the symmetric objective can sustain a persistent modality gap rather than enforcing identical marginal representations.

\subsection{MS-COCO Experiments: Pretrained Gap and Controlled Pair Corruption}
\label{app:coco_exp}

We complement the synthetic experiments with two studies on MS-COCO. The first is an observational study of pretrained CLIP-like models, designed to assess whether these models exhibit a modality gap and whether improved retrieval performance necessarily corresponds to a smaller cross-modal distributional discrepancy. The second is a controlled intervention study designed to test how weakening image--text compatibility during training affects the modality gap. Together, these studies relate the intrinsic geometric analysis to realistic image--text representations.

\subsubsection{Probing Pretrained Models on MS-COCO}

\begin{table}[t]
    \caption{Retrieval performance and cross-modal discrepancy for pretrained OpenCLIP checkpoints evaluated on the first 5{,}000 images of MS-COCO \texttt{val2017}. Retrieval is reported as image-to-text and text-to-image Recall@1. Cross-modal discrepancy is quantified by the energy distance, RBF-MMD$^2$ at the median bandwidth, the centroid gap $\|\mu_I-\mu_T\|$, and centroid cosine similarity. Lower energy distance, MMD$^2$, and centroid gap indicate smaller cross-modal discrepancy.}
    \label{tab:coco_pretrained_clip_gap}
    \centering
    \resizebox{0.7\linewidth}{!}{%
    \begin{tabular}{lcccccc}
    \toprule
    Model & I$\to$T R@1 & T$\to$I R@1 & $\mathcal{E}(p_I,p_T)$ & MMD$^2_{\sigma=\mathrm{med}}$ & $\|\mu_I-\mu_T\|$ & $\cos(\mu_I,\mu_T)$ \\
    \midrule
    PE-Core-L-14-336 & 0.741 & 0.559 & 0.599 & 0.255 & 0.854 & 0.097 \\
    PE-Core-B-16 & 0.703 & 0.499 & 0.545 & 0.237 & 0.808 & 0.189 \\
    ViT-bigG-14 & 0.670 & 0.506 & 0.288 & 0.129 & 0.598 & 0.384 \\
    ViT-H-14 & 0.662 & 0.485 & 0.375 & 0.159 & 0.697 & 0.048 \\
    ViT-L-14 & 0.569 & 0.357 & 0.697 & 0.303 & 0.898 & 0.184 \\
    ResNet-50x16 & 0.559 & 0.354 & 0.502 & 0.230 & 0.761 & 0.349 \\
    RN101 & 0.498 & 0.299 & 0.443 & 0.228 & 0.670 & 0.604 \\
    ViT-B-16 & 0.492 & 0.302 & 0.688 & 0.313 & 0.871 & 0.309 \\
    ResNet-50 & 0.480 & 0.289 & 0.627 & 0.276 & 0.849 & 0.237 \\
    \bottomrule
    \end{tabular}%
    }
\end{table}

\paragraph{Setup and evaluations.}
We evaluate a collection of pretrained OpenCLIP checkpoints \cite{ilharco_gabriel_2021_5143773} on MS-COCO \texttt{val2017}. For each checkpoint, we use the first 5000 validation images and up to 5 captions per image, yielding one normalized image embedding ($\mathbf z_I$) per image and one normalized text embedding ($\mathbf z_T$) per caption. Retrieval is evaluated in both directions under the many-captions-to-one-image matching protocol. Image-to-text Recall@$K$ assesses whether any of the top-$K$ retrieved captions is associated with the query image, whereas text-to-image Recall@$K$ assesses whether the correct image appears among the top-$K$ retrieved images for a given caption. We report retrieval performance using Recall@1 (R@1), and define AvgR@1 as the mean of the R@1 scores in the two retrieval directions.

To quantify cross-modal discrepancy, we compute the centroid gap $\|\mu_I-\mu_T\|$ with $\mu_I=\mathbb E[\mathbf z_I]$ and $\mu_T=\mathbb E[\mathbf z_T]$, and the energy distance
\[
\mathcal E(p_I,p_T)
=
2\mathbb E[d(\mathbf z_I,\mathbf z_T)]
-
\mathbb E[d(\mathbf z_I,\mathbf z_I')]
-
\mathbb E[d(\mathbf z_T,\mathbf z_T')],
\]
where $p_I$ and $p_T$ denote the image and text embedding marginals, $\mathbf z_I,\mathbf z_I'\sim p_I$ and $\mathbf z_T,\mathbf z_T'\sim p_T$ are independent draws, and $d(\cdot,\cdot)$ is the Euclidean distance between normalized embeddings. Whereas the centroid gap captures only the separation between the modality means, the energy distance compares cross-modal distances with within-modal distances and thus provides a broader measure of distributional mismatch. Unlike retrieval, which evaluates alignment between matched image--caption pairs, the energy distance operates at the distribution level: it is small when the image and text embedding marginals are distributionally similar and large when cross-modal distances remain systematically greater than within-modal distances. We additionally report an RBF-MMD diagnostic using a bandwidth selected by the median-distance heuristic.

\begin{figure}[t]
    \centering
    \includegraphics[width=0.45\linewidth]{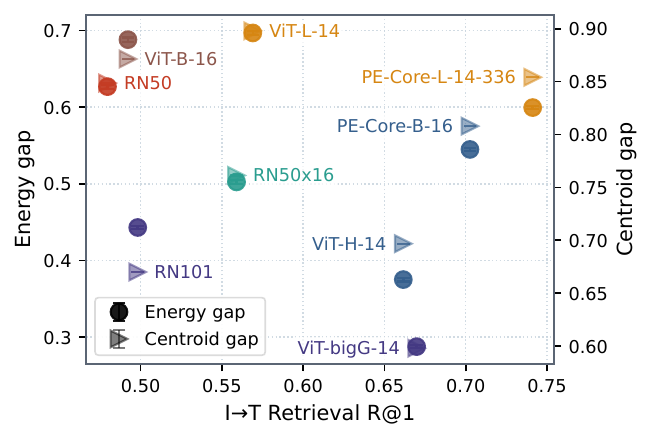}
    \hspace{1.5em}
    \includegraphics[width=0.45\linewidth]{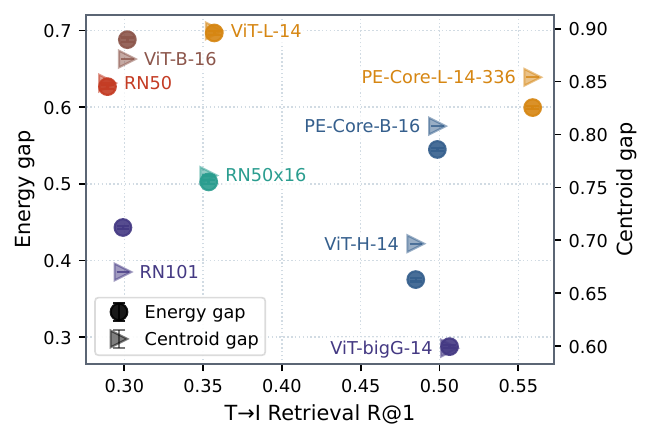}
    \caption{Directional retrieval versus cross-modal gap on MS-COCO for pretrained OpenCLIP checkpoints. Left: image-to-text retrieval (I$\to$T R@1) versus gap. Right: text-to-image retrieval (T$\to$I R@1) versus gap. In both panels, the left y-axis shows energy distance, and the right y-axis shows centroid gap. The relation is clearly non-monotone: models with similar directional retrieval can exhibit substantially different cross-modal discrepancy. This illustrates that retrieval performance and modality gap are related but not equivalent, and that strong pretrained models may still maintain a substantial structural separation between image and text marginals.}
    \label{fig:coco_retrieval}
\end{figure}

\paragraph{Results.}
As shown in \Cref{tab:coco_pretrained_clip_gap}, the results indicate that retrieval and cross-modal discrepancy are only partially aligned. As already seen in the AvgR@1 plot in \Cref{fig:coco} (left), better retrieval often corresponds to a smaller gap, but the trend is clearly non-monotone. For instance, PE-Core-L-14-336 achieves the best retrieval yet still has a large energy distance and centroid gap, whereas ViT-bigG-14 attains slightly lower retrieval but the smallest gap values in the table. The separate I$\to$T and T$\to$I versus gap plots (\Cref{fig:coco_retrieval}) show that this is not caused by averaging: models with nearly identical directional retrieval can still differ substantially in energy and centroid discrepancy, e.g., ViT-bigG-14 versus PE-Core-B-16 in T$\to$I retrieval, or ViT-L-14 versus ResNet-50x16 in I$\to$T retrieval. The RBF-MMD values show the same broad pattern, suggesting that the observation is not specific to a single discrepancy metric. Overall, these results show that a modality gap persists across the evaluated pretrained CLIP variants and is not reducible to retrieval performance alone; instead, it reflects an additional geometric property of the learned image and text marginals.

\subsubsection{Controlled Same-Category Pair Corruption on MS-COCO}
\label{app:coco_samecat_corruption}

\paragraph{Setup and evaluations.}
To test more directly whether weakening image--text compatibility enlarges the modality gap, we perform a controlled corruption experiment on MS-COCO \texttt{train2017}. We first build an index that records, for each image, its caption set, its COCO category labels, and the pool of other images sharing each category. During training, the image is always kept fixed. With probability $1-p$, we pair it with one of its own captions; with probability $p$, we instead replace the caption by a randomly chosen caption from a \emph{different} image that shares at least one category with the current image. Thus, the corruption preserves coarse semantic relatedness while breaking exact instance-level correspondence. We sweep the corruption probability over $p\in\{0,0.25,0.50,0.75,1.00\}$.

We fine-tune pretrained OpenCLIP backbones (ResNet-50 and ViT-B-16) using the standard symmetric CLIP loss. To isolate the effect of pair corruption rather than full-model adaptation, we freeze the backbone towers and update only the projection heads together with the logit scale. Training uses AdamW with learning rate $10^{-5}$, weight decay $0.05$, batch size $256$, and $5000$ optimization steps. Evaluation follows the same protocol as in the previous experiment: we encode the first $5000$ images of MS-COCO \texttt{val2017} together with up to $5$ captions per image, report image-to-text and text-to-image Recall@1, and define AvgR@1 as their mean. As the geometric diagnostic, we report the centroid gap $\|\mu_I-\mu_T\|$.

\begin{figure}[t]
    \centering
    \includegraphics[width=0.45\linewidth]{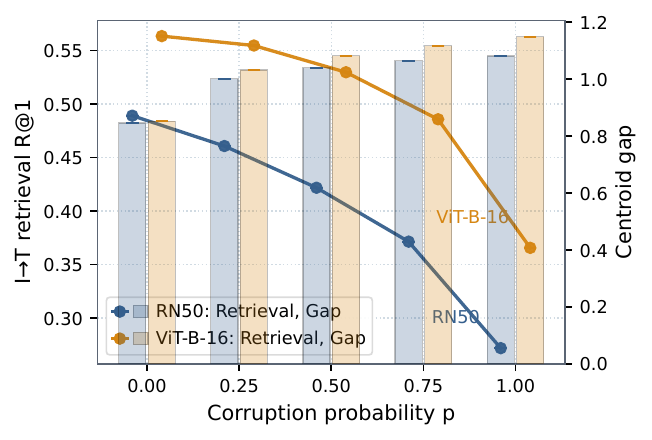}
    \hspace{1.5em}
    \includegraphics[width=0.45\linewidth]{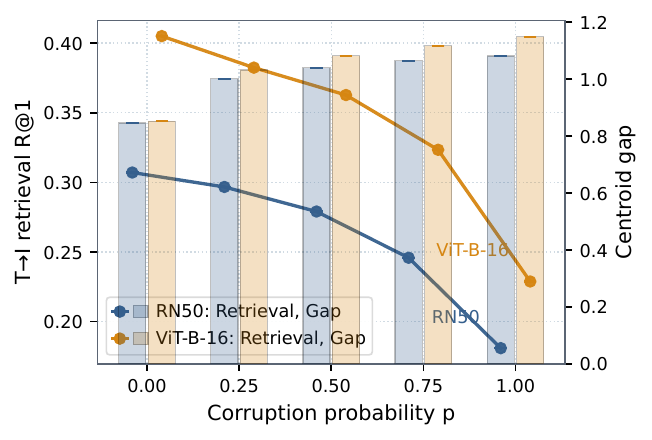}
    \caption{Directional retrieval under same-category caption corruption on MS-COCO. Left: image-to-text retrieval (I$\to$T R@1) versus corruption probability $p$. Right: text-to-image retrieval (T$\to$I R@1) versus $p$. Bars show the centroid gap on the right axis. For both RN50 and ViT-B-16, increasing same-category mispairing degrades retrieval in both directions while systematically enlarging the modality gap.}
    \label{fig:coco_exp2_retrieval}
\end{figure}

\paragraph{Results.}
As shown in \Cref{fig:coco} (right), the effect of corruption is clean for both backbones. As the caption corruption probability $p$ increases, retrieval performance degrades while the centroid gap grows. For ResNet-50, AvgR@1 drops from $0.398$ at $p=0$ to $0.226$ at $p=1$, while the centroid gap increases from $0.845$ to $1.081$. For ViT-B-16, AvgR@1 drops from $0.484$ to $0.297$, while the centroid gap rises from $0.851$ to $1.148$. The same pattern holds in both retrieval directions, as shown in \Cref{fig:coco_exp2_retrieval}: for ResNet-50, I$\to$T R@1 decreases from $0.489$ to $0.272$ and T$\to$I R@1 from $0.307$ to $0.181$; for ViT-B-16, I$\to$T R@1 decreases from $0.564$ to $0.366$ and T$\to$I R@1 from $0.405$ to $0.229$.

Several aspects of this trend are important. First, the corruption is not arbitrary: the replacement caption is sampled from a same-category image, so the supervisory signal remains semantically plausible at a coarse level. Nevertheless, this weaker form of mispairing is already associated with a substantial increase in geometric separation between the image and text marginals. Second, the centroid gap reacts very early: even moving from $p=0$ to $p=0.25$ causes a sharp increase in gap for both ResNet-50 and ViT-B-16, while the retrieval drop is initially much milder. This suggests that cross-modal distributional separation is highly sensitive to degradation in pairwise compatibility, and may emerge before retrieval collapses. Third, the monotone behavior is consistent across both architectures, indicating that the phenomenon is not specific to either of the two backbone families evaluated here.

Overall, this experiment provides a controlled real-data analogue of the mechanism developed in the main text. Exact image--text compatibility is progressively weakened, and the learned representations respond in two coupled ways: matched-pair retrieval worsens, and the image/text marginals drift farther apart. This behavior is consistent with the structural-gap view of symmetric contrastive learning: weakening pairwise compatibility can make cross-modal coincidence harder to sustain, and in this controlled setting the modality gap widens systematically.

\section{Discussion}
\label{app:implications}

\subsection{Limitations and Future Directions}

We analyze symmetric InfoNCE in a tractable large-batch, low-temperature geometric regime. These assumptions isolate the core mechanism, but they also mark the present boundary of the theory and point to several natural extensions.

\paragraph{Finite-temperature and finite-batch effects.}
Our analysis combines a large-batch regime ($N\to\infty$ at fixed $\tau$) with low-temperature asymptotics ($\tau\to0^+$) for the intrinsic geometry. Practical training instead uses finite batches and moderate temperatures, introducing stochasticity and entropic smoothing that may soften the barrier-like geometry identified in the limiting analysis. Characterizing finite-temperature and finite-batch corrections, especially how stochastic gradient noise interacts with such barriers, is a natural next step toward sharper practical predictions.

\paragraph{Singularities and boundary behavior.}
The multimodal analysis imposes a strict density floor to regularize the logarithmic barrier near the boundary of the density space. Removing this technical constraint would require analyzing the singular geometry of the KL divergence near this boundary, and is likely essential for understanding singular measures, including distributions that collapse onto lower-dimensional manifolds or discrete sets.

\paragraph{Geometric homogeneity.}
We assume a fixed compact embedding manifold with constant kernel volume. Extending the framework to heterogeneous geometries, where the local kernel volume varies across the space, would broaden the present measure-theoretic analysis toward heterogeneous embedding geometries, including settings relevant to hyperbolic contrastive learning and other objectives with spatially varying normalization.

\paragraph{Objective geometry versus optimization dynamics.}
Our theory describes the geometry of the objective rather than the trajectory of a specific optimization algorithm. Understanding how stochastic optimization explores the resulting barrier-shaped landscape, and under what compatibility conditions it can reach configurations with exact or near-exact marginal matching, remains an important open problem at the interface of geometric analysis and learning theory.

\subsection{Broader Implications}

Our analysis suggests that contrastive learning should be understood not only through pointwise alignment, but through the population geometry induced on the representation space. In the large-batch regime, InfoNCE gives rise to deterministic fields and intrinsic energies over $\mathcal Z$, clarifying why strong retrieval can coexist with persistent modality-level discrepancies: matching positive pairs does not by itself control the induced marginal geometry. In particular, the multimodal theory shows that improving pairwise alignment alone need not eliminate the modality gap, because pairwise alignment does not by itself determine the induced marginal geometry. This points to a practical lesson for both evaluation and objective design: beyond retrieval, one should monitor distributional diagnostics of the learned marginals, and interventions aimed at closing the gap may need to act on the induced population geometry, for example through explicit discrepancy regularization, shared reference fields, or other mechanisms that weaken the barrier-like coupling between modalities.

More broadly, this perspective suggests that intrinsic population-level analyses can be useful for isolating principled properties of representation-learning objectives that are not obvious from parameter-space dynamics alone. In this view, objective geometry, optimization dynamics, parametrization-specific inductive biases, and identifiability guarantees address distinct but complementary questions: the intrinsic geometry clarifies which population-level configurations the loss favors in the analyzed regime, optimization theory explains how algorithms explore that landscape, and identifiability or generative analyses address when the resulting representations correspond to meaningful latent structure. Our results, therefore, point to a broader methodological lesson: understanding modern representation learning may require combining intrinsic objective-level analysis with complementary theories of optimization, parametrization, and recoverability.

\section{Revision Note}
\label{app:changelog}

This is a revised presentation of the ICML 2026 version of this paper. The scientific claims, main theoretical results, and empirical conclusions are unchanged. In particular, this revision does not introduce a new theorem-level claim or alter the substantive conclusions of any theorem, proposition, corollary, or experiment. The changes are primarily intended to improve notation, rigor of exposition, scope clarity, and readability.

\paragraph{Changes relative to the ICML 2026 version.}
\begin{itemize}
    \item \textbf{Notation and formal setup.}
    We standardized notation throughout the paper, including distinguishing probability measures from densities, writing $\mathcal D_\mu(\mathcal Z)$ for $\mu$-densities, using $\mathrm{sim}$ consistently for the similarity function, making the finite-$N$ dependence of InfoNCE objectives explicit, and harmonizing measure, Jacobian, and random-variable notation. Several standing regularity conditions and density-domain conventions are also stated more explicitly.

    \item \textbf{Proof presentation and technical details.}
    We reorganized the proofs to make intermediate arguments more explicit, including the passage from finite-batch quantities to population limits, differentiation under expectations, kernel-smoothing arguments, Gibbs optimality, the multimodal extremal construction, and the parametric-inheritance results. These revisions clarify the existing arguments and their assumptions; they do not change the claims proved in the ICML version.

    \item \textbf{Scope and interpretation.}
    We refined the surrounding exposition to distinguish more clearly between pointwise large-batch objective/gradient consistency and optimization trajectories, and between multimodal coordinate geometry and existence of an attained ``best response''. Correspondingly, some informal language has been replaced by more precise terminology such as \emph{coordinate infimum}, \emph{infimizing sequence}, and \emph{barrier-like coupling}. The underlying mathematical conclusions are unchanged.

    \item \textbf{Organization and exposition.}
    The paper has been reformatted for the preprint version, figures and captions have been polished, and the supplementary material has been reorganized for easier navigation. In particular, lemmas, proofs of the unimodal and multimodal results, extended parametric-inheritance statements, experimental details, and discussion are now separated into dedicated appendices.

    \item \textbf{Experimental and discussion text.}
    The experimental setups and reported results are retained, while their descriptions have been revised to align more precisely with the scope of the theory. We also expanded the discussion of limitations and broader implications, especially the distinction between intrinsic objective geometry, optimization dynamics, parametrization, and identifiability.
\end{itemize}

For the archival version of record and citation, please refer to the ICML 2026 paper.

\end{document}